\documentclass[10pt,twocolumn,letterpaper]{article}

\usepackage[pagenumbers]{cvpr} 

\usepackage{graphicx}
\usepackage{amsmath}
\usepackage{amssymb}
\usepackage{booktabs}
\usepackage{CJKutf8}  
\usepackage{array}
\usepackage{tabularx}
\usepackage{multirow}
\usepackage{subcaption}
\usepackage[dvipsnames]{xcolor}
\usepackage{breakcites} 
\usepackage{comment}
\usepackage{url}
\usepackage{colortbl}
\usepackage{nccmath}
\usepackage{makecell}
\usepackage{diagbox}
\usepackage{hhline}
\usepackage{makecell} 
\usepackage[accsupp]{axessibility}

\usepackage{xcolor}
\usepackage{wrapfig}
\usepackage{soul}

\definecolor{yellow}{rgb}{1,1, 0.6}
\definecolor{lightyellow}{rgb}{1,1, 0.8}
\definecolor{orange}{rgb}{1, 0.8, 0.6}
\definecolor{coral}{RGB}{246,131,65}
\definecolor{pinkred}{rgb}{1, 0.6, 0.6}
\definecolor{hotpink}{RGB}{238,64,195}
\definecolor{lavender}{RGB}{207,226,243}
\definecolor{gainsboro}{RGB}{208,224,227}
\definecolor{gainsboro2}{RGB}{217,234,211}
\definecolor{blanchedalmond}{RGB}{252,229,205}

\usepackage{datenumber}
\usepackage{calc}
\usepackage[mmddyyyy]{datetime}
\newcounter{datetoday}
\newcounter{diffyears}
\newcounter{diffmonths}
\newcounter{diffdays}
\newcommand{\difftoday}[3]{%
      \setmydatenumber{datetoday}{\the\year}{\the\month}{\the\day}%
      \setmydatenumber{diffdays}{#1}{#2}{#3}%
      \addtocounter{diffdays}{-\thedatetoday}%
      \ifnum\value{diffdays}>0
        \def\diffbefore{}%
        \def\diffafter{left}%
      \else
        \def\diffbefore{}%
        \def\diffafter{ago}%
        \setcounter{diffdays}{-\value{diffdays}}%
      \fi
      \setcounter{diffyears}{\value{diffdays}/365}%
      \setcounter{diffdays}{\value{diffdays}-365*\value{diffyears}}%
      \setcounter{diffmonths}{\value{diffdays}/30}%
      \setcounter{diffdays}{\value{diffdays}-30*\value{diffmonths}}%
      \diffbefore
      \ifnum\value{diffyears}=0
      \else
        \ifnum\value{diffyears}>1
            \thediffyears\space years,
        \else
            \thediffyears\space year,
        \fi
      \fi
      \ifnum\value{diffmonths}=0
      \else
        \ifnum\value{diffmonths}>1
            \thediffmonths\space months
        \else
            \thediffmonths\space month
        \fi
      \fi
      \ifnum\value{diffdays}=0
      \else
        \ifnum\value{diffdays}>1
            \thediffdays\space days
        \else
            \thediffdays\space day
        \fi
      \fi
      \diffafter
}

\definecolor{cvprblue}{rgb}{0.21,0.49,0.74}
\usepackage[pagebackref,breaklinks,colorlinks,citecolor=cvprblue]{hyperref}

\usepackage[capitalize]{cleveref}
\crefname{section}{Sec.}{Secs.}
\Crefname{section}{Section}{Sections}
\Crefname{table}{Table}{Tables}

\newcolumntype{Y}{>{\centering\arraybackslash}X}

\def\confName{CVPR}
\def\confYear{2025}

\begin{document}
\begin{CJK}{UTF8}{}
\CJKfamily{mj}

\title{
Pixel-aligned RGB-NIR Stereo Imaging and Dataset for Robot Vision 
}

\author{
Jinnyeong Kim ~ ~ ~
Seung-Hwan Baek \\
POSTECH\\
}

\twocolumn[{%
\renewcommand\twocolumn[1][]{#1}%
\maketitle
}]

\begin{abstract}
Integrating RGB and NIR stereo imaging provides complementary spectral information, potentially enhancing robotic 3D vision in challenging lighting conditions. However, existing datasets and imaging systems lack pixel-level alignment between RGB and NIR images, posing challenges for downstream vision tasks.
In this paper, we introduce a robotic vision system equipped with  pixel-aligned RGB-NIR stereo cameras and a LiDAR sensor mounted on a mobile robot. The system simultaneously captures pixel-aligned pairs of RGB stereo images, NIR stereo images, and temporally synchronized LiDAR points. Utilizing the mobility of the robot, we present a dataset containing continuous video frames under diverse lighting conditions.
We then introduce two methods that utilize the pixel-aligned RGB-NIR images: an RGB-NIR image fusion method and a feature fusion method. The first approach enables existing RGB-pretrained vision models to directly utilize RGB-NIR information without fine-tuning. The second approach fine-tunes existing vision models to more effectively utilize RGB-NIR information.
Experimental results demonstrate the effectiveness of using pixel-aligned RGB-NIR images across diverse lighting conditions.
\end{abstract}

\section{Introduction}
\label{sec:intro}
RGB imaging captures visible light in wavelengths from 400\,nm to 700\,nm and serves as the primary data format for visual computing. 
Near-infrared (NIR) imaging with active illumination captures light in the range of 750\,nm to 1000\,nm, offering effective information in conditions where where RGB imaging is inadequate, such as low-light environments during night or poorly lit indoor conditions. 
NIR light is imperceptible to human vision, allowing for using active NIR illumination without disturbing people, which facilitates robust 3D imaging

~\cite{li2007nirilumface,Walz_2023_gatedstereo,brucker2024cross,liu2016multispectral_pedestrian} and eye tracking~\cite{broadbent2023cognitive_driver_eyetracking}. Additionally, NIR light offers supplementary scene information due to different spectral reflectance between RGB and NIR wavelengths~\cite{kim2012_3dhyperspectral,li2021near_polar_3d,hansen2010_rgbnirfacephotometric}.

Combining these benefits, RGB-NIR imaging emerges as a promising modality for challenging environments requiring robust and efficient 2D and 3D vision capabilities. RGB-NIR imaging has been studied for applications including object detection~\cite{takumi2017multispectral_detection, zhao2023metafusion} and 3D reconstruction~\cite{brucker2024cross, poggi2022cross, hong2022reflection,huang2023polarization_nir_3d}. However, existing RGB-NIR imaging systems and datasets face significant challenges compared to standard RGB vision systems. A major issue is the use of separate RGB and NIR cameras at different poses, leading to pixel misalignment between the NIR and RGB images. Previous methods address such misalignments using image registration, pose estimation, and other processing techniques, which can accumulate errors and limit effectiveness~\cite{Zhi_2018_material_cross_spectral,
brucker2024cross,Shin_Park_Kweon_2023}.

In this work, we develop a robotic vision system capable of capturing pixel-aligned RGB-NIR stereo images and synchronized LiDAR point clouds. Mounted on a mobile platform, the system enables exploration in both indoor and outdoor environments, facilitating data collection under diverse lighting conditions. Using this system, we collect and share a dataset of pixel-aligned RGB-NIR stereo images with synchronized LiDAR point clouds as ground-truth depth for 43 real-world scenes including 80,000 frames. Additionally, we provide a synthetic RGB-NIR dataset with ground-truth dense depth labels for 2,238 synthetic scenes. To demonstrate the effectiveness of pixel-aligned RGB-NIR imaging and datasets, we propose an RGB-NIR image fusion method and an RGB-NIR feature fusion method.
The first method fuses the pixel-aligned RGB-NIR images into a three-channel image that can be input into pretrained RGB vision models without finetuning. The second method involves developing a stereo depth estimation technique by finetuning a pretrained stereo network~\cite{lipson2021raft} to utilize pixel-aligned RGB-NIR features. Our experiments show improvements in various downstream tasks over using only RGB or NIR data, as well as over methods relying on pixel-misaligned RGB-NIR datasets, particularly in challenging scenarios.

In summary, our contributions are as follows.
\begin{itemize}
    \item We develop a robotic vision system with pixel-aligned RGB-NIR stereo cameras and LiDAR, capturing synchronized RGB-NIR stereo images and LiDAR point clouds under diverse lighting conditions.
    \item We present a large-scale dataset of pixel-aligned RGB-NIR stereo images and LiDAR point clouds, collected in various indoor and outdoor environments.
    \item We propose an RGB-NIR image fusion method and an RGB-NIR stereo depth estimation network using RGB-NIR feature fusion.
    \item We evaluate the proposed methods on diverse downstream tasks and illumination conditions, on synthetic and real-world datasets, outperforming baselines.
\end{itemize}

\section{Related Work}
\label{sec:related}

\paragraph{RGB-NIR Imaging}
Conventional RGB-NIR imaging systems employ separate RGB and NIR cameras positioned at different viewpoints, such as the Kinect sensor, Intel RealSense D415, and various multi-view RGB-NIR systems~\cite{choe2018ranus, Zhi_2018_material_cross_spectral, Shin_Park_Kweon_2023, couprie2013nyu_dataset, dai2017scannet, xia2018gibson_mobilerobot, yadav2023habitat, Lee_Cho_Shin_Kim_Myung_2022_Vivid, Choi_Kim_Hwang_Park_Yoon_An_Kweon_2018}. However, the captured RGB and NIR images from these systems exhibit pixel misalignment, necessitating depth-dependent registration between the RGB and NIR images. This introduces a significant challenge for effectively utilizing RGB and NIR information, creating a chicken-and-egg problem between depth estimation using spectrally fused images and spectral pose alignment using estimated depth.

To address the pixel misalignment, sequential capture with interchangeable RGB and NIR filters has been used~\cite{brown2011_rgbnir_filter, Jin_2023_DarkVision}; however, this approach is limited to capturing static scenes due to the time delay between captures. Designing custom color filter arrays for image sensors to record four channels (RGB-NIR) has also been explored~\cite{monno2018single_sensor_rgb_nir}. However, using a single sensor with a fixed global exposure for all four channels makes it unsuitable for challenging environments where the dynamic ranges for RGB and NIR channels differ significantly, such as in dark indoor or outdoor conditions with active NIR illumination.
In the domain of RGB-thermal imaging, beam splitters have been used to combine an RGB camera and a thermal camera, enabling pixel-aligned RGB-thermal imaging~\cite{kim2018multispectral, hwang2015multispectral, Gao_2019_hybrid_rgb_thermal_cam}.

We capture pixel-aligned RGB-NIR stereo images alongside LiDAR point clouds using two prism-based RGB-NIR dual-sensor cameras. The system is mounted on a mobile robot, allowing scalable data acquisition of pixel-aligned RGB-NIR stereo images in both indoor and outdoor conditions.

\paragraph{RGB-NIR Fusion}
To utilize information from both RGB and NIR spectra, fusing RGB-NIR images has been extensively studied with applications in image enhancement for long-distance visibility and fog penetration~\cite{awad2019adaptive_rgb_nir,herrera2019color,Connah_Drew_Finlayson_2014_grad_poisson,zhao2020bayesian_fusion}, as well as in information visualization leveraging different RGB-NIR reflectance properties~\cite{li_2021_viz_nir_fusion_spectrum,hansen2010_rgbnirfacephotometric,shibata2016versatile_nir,herrera2021tophatnir}. Conventional methods apply simple arithmetic operations such as addition and subtraction in RGB, HSL, HSV, and YUV color spaces to create images that contain both RGB and NIR information~\cite{toet1996false_color_fusion, Liu_Huang_2010_EM, herrera2019color, herrera2021tophatnir,
fredembach2008colouring_hsv, park2016color_restore_rgbn}.
Edge-aware filtering techniques have been employed to refine noisy and low-contrast RGB images using guidance from clean and high-contrast NIR images, showing promising performance in various imaging scenarios~\cite{Li_2013_guidedfiltering_fusion, jang2017colour_fusion_dehazing, li_2021_viz_nir_fusion_spectrum, Connah_Drew_Finlayson_2014_grad_poisson}. More recently, learning-based approaches have attempted to enhance RGB images and their 3D vision applications using NIR guidance by fusing RGB-NIR features~\cite{su2021multidenoising, Jung_2020_FusionNet, Jin_2023_DarkVision, li2018infrared_vgg, Wang_Wang_Yang_Fang_Wan_2023} or RGB-thermal features~\cite{hwang2015multispectral, Zhang_2024_TFDet, deevi2024rgbx}.

Leveraging our pixel-aligned RGB-NIR images, we develop a RGB-NIR image fusion method that effectively encodes information from both RGB and NIR images into a single three-channel image with learned spatially-varying weights. The resulting fused image can be directly used as input to neural networks pretrained on RGB images. 

\paragraph{RGB-NIR Depth Estimation}
Cross-spectral correspondence matching between RGB and NIR images using an RGB-NIR color transform network has demonstrated promising depth estimation results when the RGB-NIR reflectances are similar~\cite{Zhi_2018_material_cross_spectral}. Liang et al. showed that GAN based architecture can enhance RGB-NIR color transform on cross-spectral stereo matching~\cite{liang2019rgbnir_UCSSM}. Active stereo systems using structured NIR illumination, such as the Kinect and Intel RealSense D415, employ multiple RGB and NIR cameras at different viewpoints. These systems often suffer from the aforementioned chicken-and-egg problem between depth estimation using spectrally fused images and spectral fusion using the estimated depth. Brucker et al.~\cite{brucker2024cross} fuse red, blue, and clean channels of stereo RCCB cameras with gating-based time-of-flight cameras positioned differently but still face a similar chicken-and-egg problem. In the domain of RGB-thermal imaging, Guo et al.~\cite{Guo_2023_cross_stereo_est_and_data} demonstrate effective depth estimation using pixel-aligned RGB-thermal stereo cameras with beam splitters. 

We develop a depth estimation method leveraging our pixel-aligned RGB-NIR images for robust depth estimation, thereby bypassing the chicken-and-egg problem between depth estimation and spectral fusion.

\begin{figure*}[t]
  \centering
  \includegraphics[width=\textwidth]{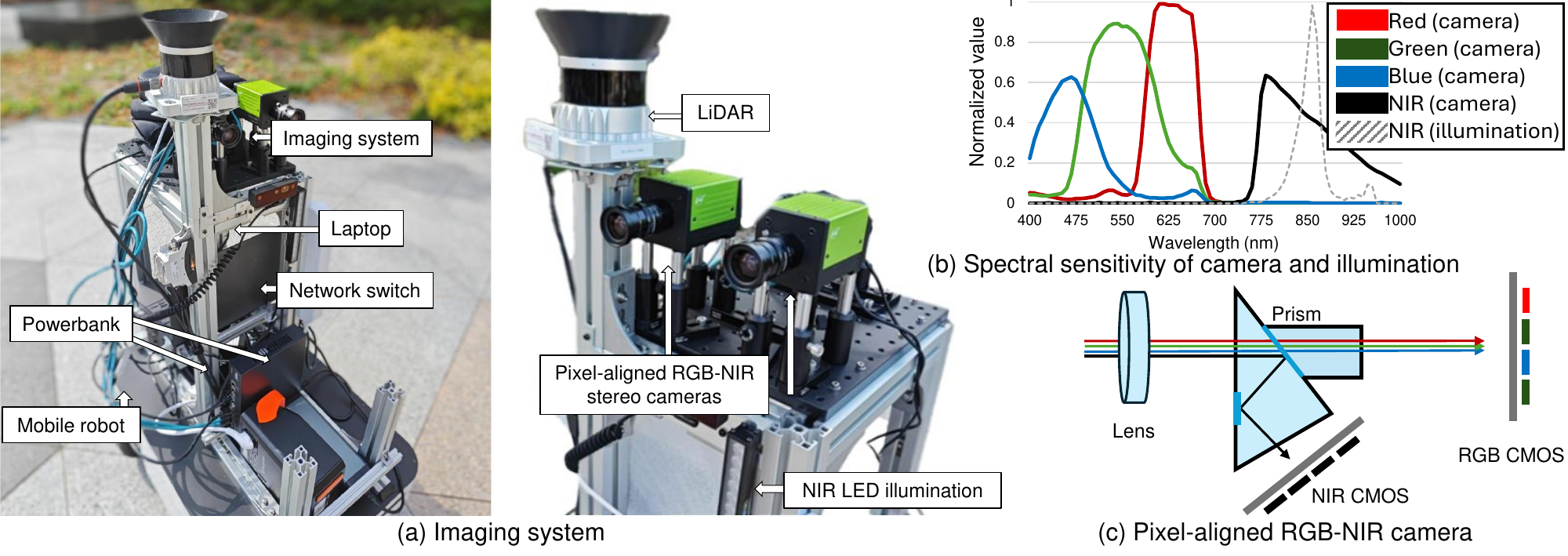}
  \caption{\textbf{RGB-NIR stereo imaging system}. (a) Stereo camera setup integrated with LiDAR and active NIR illumination. 
  (b) Spectral sensitivity profiles of the camera sensors and the irradiance profile of the NIR illumination. 
(c) Diagram of the pixel-aligned RGB-NIR camera featuring a dichroism prism, where RGB light penetrates the prism and NIR light is reflected, achieving spectral separation. 
} 
  \label{fig:jai_camera}
  \vspace{-2mm}
\end{figure*}

\section{Pixel-aligned RGB-NIR Stereo Imaging}

\paragraph{Robotic Imaging System}
Figure~\ref{fig:jai_camera}(a) illustrates our robotic imaging system, which integrates a stereo pair of pixel-aligned RGB-NIR cameras, NIR illumination, and a LiDAR mounted on a mobile robot (AgileX Ranger Mini 2.0). Each RGB-NIR camera (JAI FS-1600D-10GE), as shown in Figure~\ref{fig:jai_camera}(c), employs a dichroic prism to separate RGB and NIR light, independently captured by RGB and NIR CMOS sensors. To enhance NIR visibility under varying lighting conditions, we include active NIR illumination (Advanced Illumination
AL295-150850IC) without disturbing human vision. The spectral sensitivities of the camera and the illumination are shown in Figure~\ref{fig:jai_camera}(b). We installed a LiDAR (OUSTER OS1) to acquire ground-truth sparse depth maps, along with angular velocity and linear acceleration data from its inertial measurement unit.
We use RJ-45 interfaces for the data transmission of the cameras and LiDAR. 
We set up an AC powerbank to supply power to the devices that require an AC adapter. These devices are connected to a laptop, which triggers the capture pipeline and manages real-time data storage.
We manage separate threads on the laptop for the left camera, right camera, and LiDAR, acquiring synchronized data. 

\paragraph{Image Formation}
We model the image formation of the pixel-aligned RGB-NIR camera, which captures RGB and NIR images on separate CMOS sensors. This configuration results in the same exposure time and gain across the RGB channels, with separate exposure time and gain for the NIR channel: $t_R = t_G = t_B \neq t_{\text{NIR}}$ and $g_R = g_G = g_B \neq g_{\text{NIR}}$. 
The image captured by camera $c \in \{\text{left}, \text{right}\}$ for channel $i \in \{\text{R}, \text{G}, \text{B}, \text{NIR}\}$ is given by
\begin{align}
\label{eq:image_formation}
    I_{i}^{c}(p^c) = \eta_1 + g_i \left( \eta_2 + t_i \left( R_i^c(p^c) \left( E_i^c(p^c) + L_i^c(p^c) \right) \right) \right),
\end{align}
where $p^c$ is a pixel of camera $c$, $R_i^c(p^c)$ is the reflectance at pixel $p^c$ for channel $i$, and $\eta_1$, $\eta_2$ are Gaussian noise in the post-gamma and pre-gamma stages, respectively.
The intensities measured in each channel are influenced by environmental illumination, denoted as $E_i^c(p^c)$ for channel $i$. The active illumination $L_i^c(p^c)$ is present only in the NIR channel, satisfying $L_i^c(p^c) = 0$ for $i \in \{\text{R}, \text{G}, \text{B}\}$, since the active illumination is confined to the NIR spectrum (see Figure~\ref{fig:jai_camera}(c)).

\paragraph{Calibration}
One advantage of using pixel-aligned cameras is that calibration between RGB and NIR images is unnecessary. Thus, we only calibrate the {stereo cameras} using checkerboard calibration~\cite{bouguet2004camera_calibration}.
The cameras contain precision time protocol, which enables maintaining temporal difference between stereo cameras less than 1 microsecond. We also obtain the LiDAR extrinsics by using corresponding point pairs between left-camera images and LiDAR point clouds~\cite{quan1999linear_pnp}.

\begin{figure*}[t]
    \newcommand{\GCL}{\cellcolor[HTML]{C6EFCE}}
    \newcommand{\GCR}{\cellcolor[HTML]{FFC7CE}}
    \newcommand{\GCY}{\cellcolor[HTML]{FFF2CC}}
    
    \begin{minipage}{\textwidth}
        \begin{minipage}{0.48\linewidth}
            \centering
            \vspace{1mm} 
            \includegraphics[width=\textwidth]{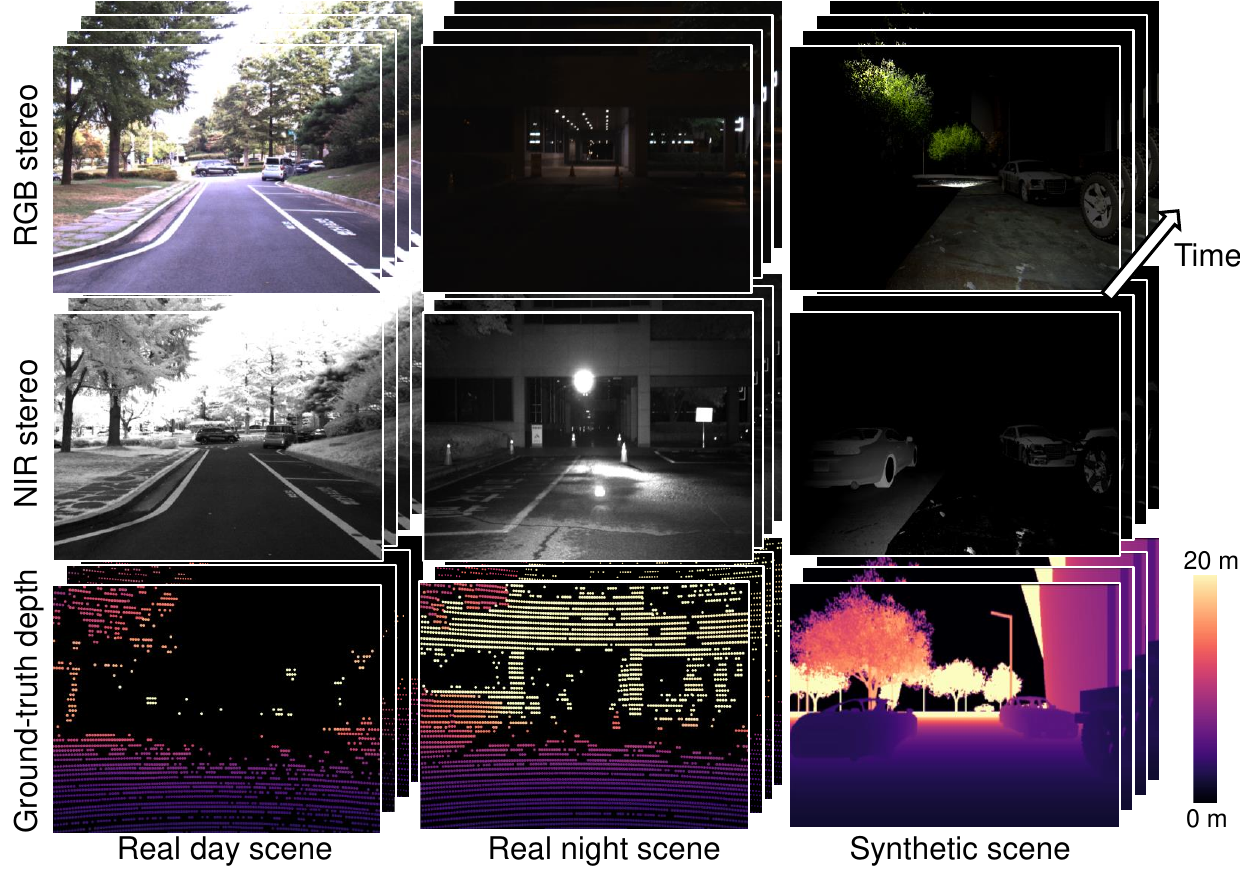}
             
            \footnotesize(a) Our dataset samples
        \end{minipage}
        \hfill
        \begin{minipage}{0.51\linewidth}
            \centering
            \setlength{\tabcolsep}{2pt}
            \footnotesize
            \newcolumntype{C}{>{\centering\arraybackslash}X}
\begin{tabularx}{\linewidth}{|p{1cm}|C|C|C|C|C|p{3cm}|}

\hline
\textbf{} & \multicolumn{4}{c|}{\textbf{RGB-NIR alignment}} & \multirow{2}{*}{\makecell{GT \\ depth}} & \multirow{2}{*}{Environment} \\ 
\cline{2-5}
 & Pixel-aligned & Multi-view & RGB stereo & NIR stereo & & \\  \Xhline{2\arrayrulewidth}\Xhline{0.2pt}
 
\cite{chebrolu2017agricultural} & \GCL O & \GCL O & \GCR X & \GCR X & \GCL O & \GCR \scriptsize Outdoor, day \\ \hline

\cite{takumi2017multispectral_detection} & \GCR X & \GCL O & \GCR X & \GCR X & \GCR X & \GCY \scriptsize Outdoor, day, night \\ \hline

\cite{valada2017rgb_nir_seg_robot} & \GCR X & \GCL O & \GCL O & \GCR X & \GCL O & \GCR \scriptsize Outdoor, day \\ \hline

\cite{toet2016dataset_nir_fusion_aligned} & \GCL \GCL O & \GCR X & \GCR X & \GCR X & \GCR X & \GCY \scriptsize Outdoor, day, night \\ \hline

\cite{Zhi_2018_material_cross_spectral} & \GCR X & \GCL O & \GCR X & \GCR X & \GCR X & \GCY \scriptsize Outdoor, sunny, dark \\ \hline

\cite{choe2018ranus} & \GCR X & \GCL O & \GCR X & \GCR X & \GCR X & \GCY \scriptsize Outdoor, day, low-light \\ \hline

\cite{Shin_Park_Kweon_2023} & \GCR X & \GCL O & \GCL O & \GCL O & \GCL \GCL O & \GCL \scriptsize Outdoor, day, night, rain \\ \hline

\cite{mortimer_2024_goose} & \GCL O & \GCL O & \GCL O & \GCR X & \GCL O & \GCY \scriptsize Outdoor, offload, day \\ \hline

\Xhline{2\arrayrulewidth}\Xhline{0.2pt}

Ours & \GCL O & \GCL O & \GCL O  &\GCL O& \GCL O & \GCL \scriptsize Outdoor, day, night, indoor \\ \hline
Ours (synthetic) & \GCL O & \GCL O & \GCL O & \GCL O & \GCL O & \GCL \scriptsize Outdoor, day, night, indoor \\ \hline

\end{tabularx}

 \vspace{2mm}
\footnotesize(b) RGB-NIR machine vison dataset comparison
        \end{minipage}
    \end{minipage}
    \vspace{3mm}
    \begin{minipage}{\textwidth}
        \centering
            \includegraphics[width=\textwidth]{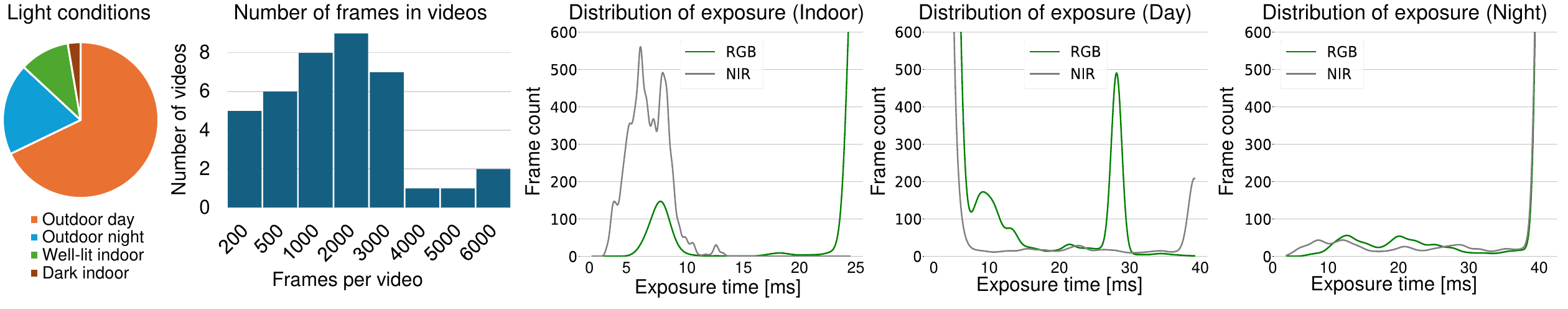}
             \footnotesize(c) Statistics of our dataset
    \end{minipage}

    \vspace{-3mm}
    \caption{\textbf{RGB-NIR stereo dataset}. 
    (a) Our dataset comprises stereo RGB image pairs, stereo NIR image pairs, and sparse depth point clouds, all captured from continuous video sequences. 
    (b) A comparison of our dataset with other RGB-NIR image datasets, specifically curated for 3D vision tasks. 
    (c) A quantitative analysis of our dataset is presented, illustrating the distribution of frames under varying lighting conditions, a histogram showing the number of frames per video sequence, {and the frame exposure time distributions for RGB and NIR sensors under three distinct lighting scenarios.}} 
    \vspace{-2mm}
    \label{fig:dataset}
\end{figure*}

\paragraph{Real-world Dataset}
Using the system, we acquire a dataset consisting of 39 videos totaling 73,000 frames under various lighting conditions for training and 4 videos of 7000 frames for testing. For each frame, we provide pixel-aligned RGB-NIR stereo images, a sparse LiDAR {point cloud}, sensor timestamps, and exposure values for the RGB and NIR CMOS sensors.
Each video was recorded at frame rates of 5–10~Hz, with durations ranging from five to over thirty minutes. Figure~\ref{fig:dataset}(a) shows samples from our real dataset and augmented synthetic dataset. We categorize our dataset based on lighting conditions: outdoor day, outdoor night, well-lit indoor, and dark indoor. Figure~\ref{fig:dataset}(c) shows the statistics of our dataset.
We use auto exposure on both RGB and NIR sensors, with fixed gain values, and recorded the exposure time for each image. The distribution of exposure times in our dataset varies depending on lighting conditions. For the indoor dataset, many frames have short exposure times for NIR images, as active illumination is sufficient to illuminate nearby indoor scenes. In the daytime dataset, strong sunlight causes the exposure times to shift towards very short durations under direct sunlight and longer durations in shaded areas. For the night dataset, although many frames exhibit long exposure times for both RGB and NIR, the average exposure time for NIR frames is shorter than that for RGB frames.

\paragraph{Synthetic Dataset}
We augmented synthetic RGB stereo-image dataset~\cite{mayer2016sceneflow} to obtain RGB-NIR stereo dataset with ground-truth dense depth. Figure~\ref{fig:dataset}(a) shows a sample of synthetic scenes.
We compute the RGB reflectance map $R_\text{RGB}$ and the NIR reflectance map and $R_\text{NIR}$ using the RGB-to-NIR color synthesis method~\cite{Gruber_2019_gated2depth}.
We then simulate diverse environmental lighting and obtain the RGB-NIR stereo images using Equation~\eqref{eq:image_formation}.
Our augmented dataset comprising 2,238 short sequences (10 frames each) and four long sequences with total 2,200 frames. Our augmentation method is detailed in the supplemental material.

\paragraph{Dataset Comparison}
Figure~\ref{fig:dataset}(b) compares our dataset with existing RGB-NIR stereo datasets in terms of alignment, environment diversity, and the availability of ground truth depth. 
Unlike datasets requiring color transformation or material segmentation~\cite{choe2018ranus,Zhi_2018_material_cross_spectral}, we capture stereo images directly in each spectral band, enhancing quality and reducing computational demands. 
While others need pose transformations due to differing camera positions~\cite{Lee_Cho_Shin_Kim_Myung_2022_Vivid,Shin_Park_Kweon_2023}, our system uses pixel-aligned RGB-NIR camera. 
By overcoming these challenges, our dataset offers high-quality RGB-NIR stereo images suitable for various applications, including autonomous vehicles and robotics.

\begin{figure*}[t]
  \centering
  \includegraphics[width=\textwidth]{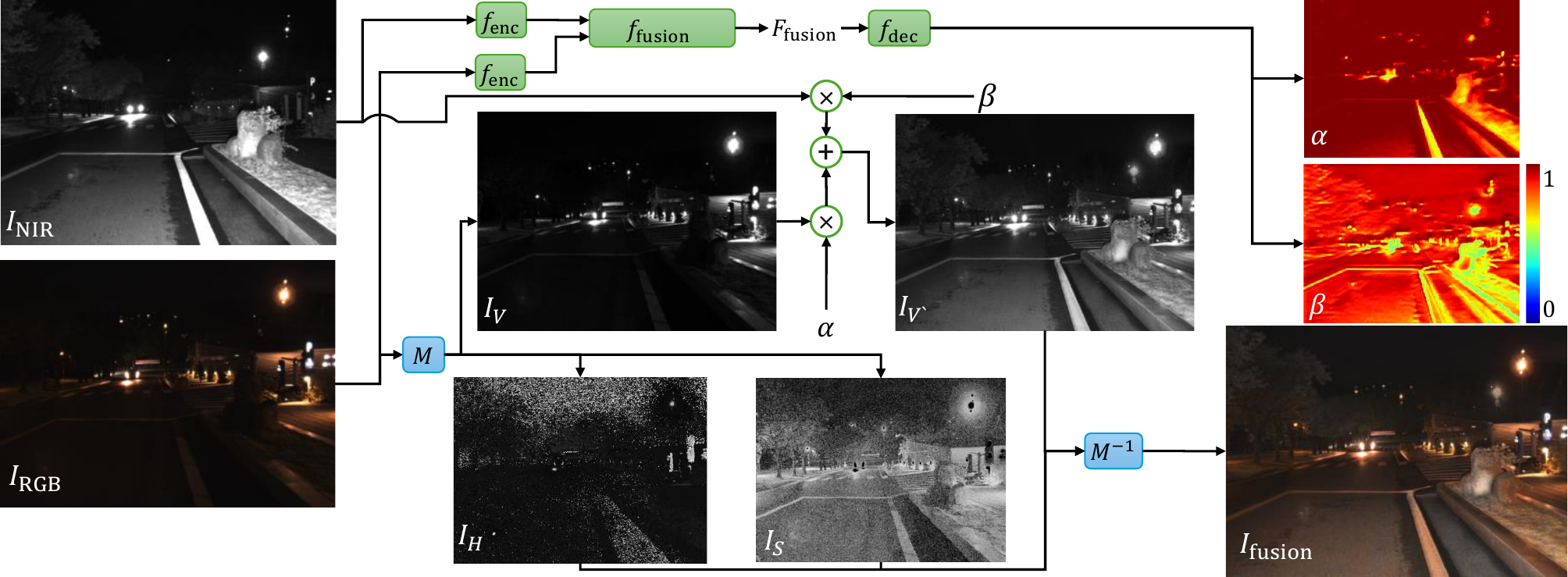}
  \caption{\textbf{RGB-NIR image fusion}.  We fuse the RGB image $I_{\text{RGB}}$ and the image $I_{\text{NIR}}$ as a weighted sum in the brightness domain $V$ after converting RGB image into HSV channel. The spatially-varying weights $\alpha,\beta$ are learned to effectively fuse RGB and NIR images. The fused image can be used as inputs to vision models such as object detection, stereo depth estimation, and structure from motion. } 
  \vspace{-1mm}
  \label{fig:color_fusion}
\end{figure*}
\section{Pixel-aligned RGB-NIR Fusion}
\label{sec:fusion}

We introduce two methods fusing pixel-aligned RGB-NIR images to improve performance of vision models. 

\subsection{Image Fusion with Learned Weights}
\label{sec:image_fusion}
First, we fuse pixel-aligned RGB-NIR images into a single three-channel image that can be fed to vision models pretrained for RGB images without finetuning. Figure~\ref{fig:color_fusion} shows the overview of our image-fusion method.

\paragraph{Baseline}
As a baseline, we use the HSV channel blending method~\cite{fredembach2008colouring_hsv} to enhance performance on vision application without losing photometric consistency of both RGB and NIR. Given RGB and NIR images for the camera $c$, the baseline method converts the RGB image into HSV representation:
\begin{equation}
    [I_H^c, I_S^c, I_V^c]^\intercal = M [I_R^c, I_G^c, I_B^c]^\intercal,
\end{equation}
where $M$ is the RGB-to-HSV conversion matrix. $I_H^c, I_S^c, I_V^c$ are the hue, saturation, and brightness images.
We then combine the brightness channel $I_V^c$ with the NIR image $I_\text{NIR}^c$ to include details from both while the hue and saturation channels remain unchanged. The fused HSV image is converted back to the RGB domain obtaining the fused image $I_\text{fusion}^c$:
\begin{equation}
\label{eq:fusion}
    I_\text{fusion}^c = M^{-1} [I_H^c, I_S^c, \alpha I_V^c + \beta I_\text{NIR}^c]^\intercal
\end{equation}
where $\alpha$ and $\beta$ are fusion weights.

The baseline method has a limitation because the weights $\alpha$ and $\beta$ are predefined. This makes the fusion process non-adaptable to various environments and lighting conditions which make image details in NIR and RGB images vary significantly.

\paragraph{Learned Spatially-varying Weights}
Instead of using predefined constants $\alpha$ and $\beta$, we propose to learn spatially-varying $\alpha$ and $\beta$ using an attention-based MLP.
We first encode each image into a 256-channel feature map as follows:
\begin{equation}
    f_\text{enc}(I_\text{RGB}^c) = F_\text{RGB}^c \quad \text{and} \quad f_\text{enc}(I_\text{NIR}^c) = F_\text{NIR}^c
\end{equation}
where $f_\text{enc}$ is a ResNet-based feature extractor. For the NIR feature encoding, we copy the NIR channel three times as an input to the pretrained three-channel feature extractor.

We then obtain fused feature map $F_\text{fusion}^c$ by combining $F_\text{RGB}^c$ and $F_\text{NIR}^c$ using attentional fusion:
\begin{equation}
    F_\text{fusion}^c = f_\text{fusion}(F_\text{RGB}^c, F_\text{NIR}^c), 
\end{equation}
where the feature fusion module $f_\text{fusion}$ computes a sum of RGB and NIR feature maps with feature attention~\cite{Dai_2021_aff}.
Finally, spatially varying weights $\alpha$ and $\beta$ are derived by the decoder  $f_\text{dec}$ consisting of several convolutional layers:
\begin{equation}
    f_\text{dec}(F_\text{fusion}^c) = \alpha, \beta.
\end{equation}

Once the spatially-varying scene-dependent values of $\alpha, \beta$ are obtained, we estimate the fused image using Equation~\eqref{eq:fusion}.  
Then, we filter $I_\text{fusion}^c$ using guided filtering with the reference NIR image $I_\text{NIR}^c$~\cite{Li_2013_guidedfiltering_fusion}.

We evaluate the proposed method on object detection, depth estimation, and structure from motion described in Section~\ref{sec:results}. 
We only train the image-fusion models using photometric loss and stereo depth reconstruction loss, without training the downstream vision models.
Refer to the Supplemental Document for details.

\begin{figure}[t]
  \centering
  \includegraphics[width=\linewidth]{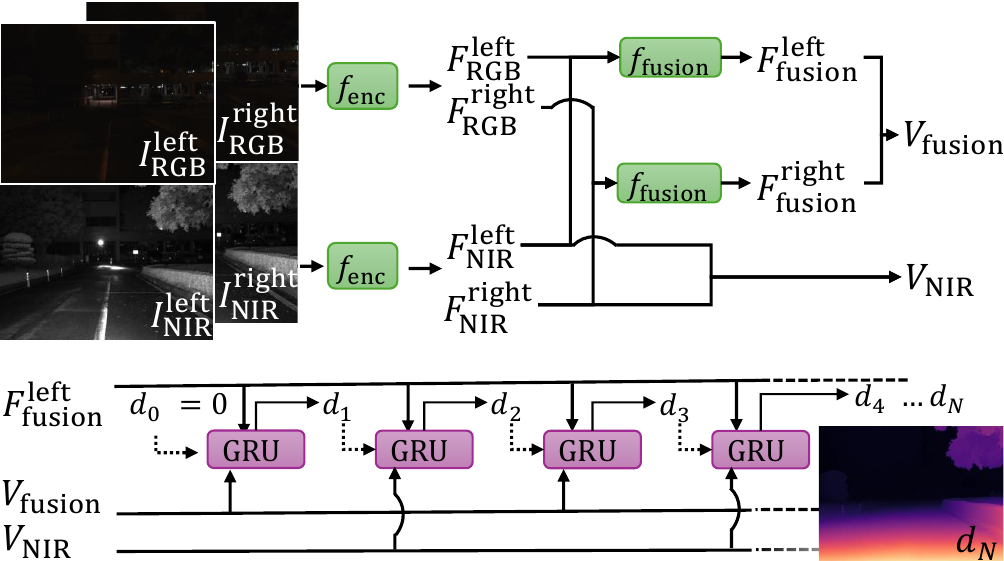}
  \caption{\textbf{RGB-NIR stereo depth estimation model}. We modified RAFT-Stereo~\cite{lipson2021raft} with attentional feature fusion and alternative correlation search for RGB-NIR depth estimation. We extract features from RGB and NIR images, fuse them, and build cost volumes. We estimate disparity by repeatedly feeding the cost volume of fused features and cost volume of NIR features to the GRU unit whose hidden state is initialized with $F_\text{fusion}^\text{left}$. } 
  \vspace{-2mm}
  \label{fig:depth_model}
\end{figure}

\begin{figure*}[ht]
  \centering
  \includegraphics[width=\textwidth]{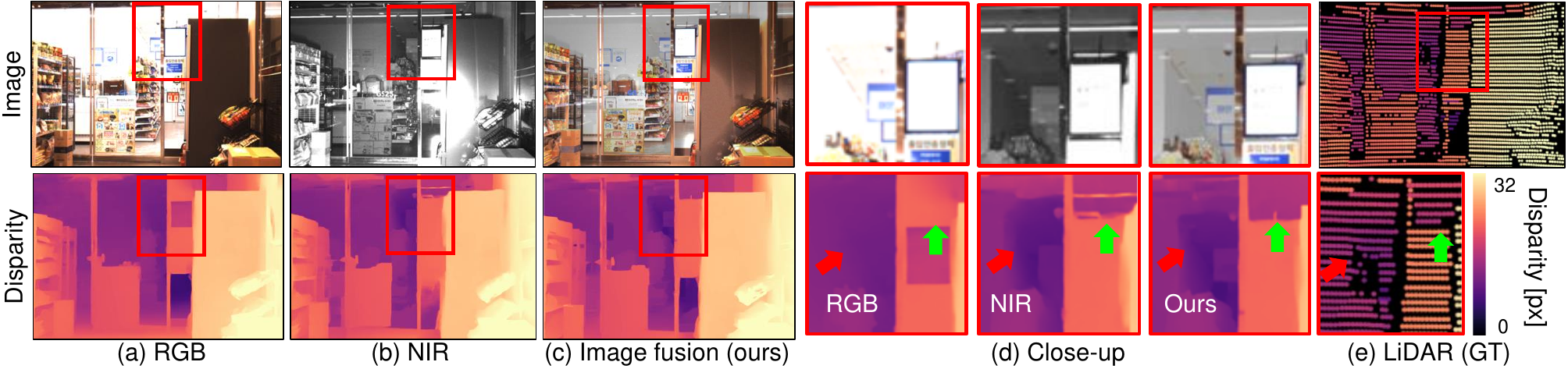}
  \caption{\textbf{Image fusion for pretrained RAFT-stereo~\cite{lipson2021raft}}.  (a)\&(b) Using a single modality either RGB or NIR images often results in sub-optimal disparity estimation in challenging lighting conditions. (c) Our RGB-NIR fused image enables robust disparity estimation without finetuning the pretrained model. (d) Closeups. (e) Ground-truth LiDAR sparse disparity. 
  }
  \label{fig:color_compare_depth}
\end{figure*}

\subsection{Feature Fusion for Stereo Depth Estimation}
\label{sec:feature_fusion}
While the image fusion method proposed in Section~\ref{sec:image_fusion} allows the use of existing vision models without finetuning, it reduces the image dimensions from four channels (RGB and NIR) to three output channels, which may result in information loss.

Here, we develop a method that fuses RGB and NIR features and finetunes a downstream model. We use the RAFT-Stereo network~\cite{lipson2021raft} as an example. Figure~\ref{fig:depth_model} shows a diagram of our proposed stereo depth estimation network.
We extract features for the RGB stereo images $I_\text{RGB}^{c}$ and NIR stereo images $I_\text{NIR}^{c}$, where $c \in \{\text{left}, \text{right}\}$, using the same ResNet-based feature extractor:
$F_\text{RGB}^c = f_\text{enc}(I_\text{RGB}^c), \quad F_\text{NIR}^c = f_\text{enc}(I_\text{NIR}^c)$.
We then apply the attention-based fusion method from Section~\ref{sec:image_fusion}:
    $F_{\text{fusion}}^c = f_{\text{fusion}}(F_{\text{RGB}}^c, F_{\text{NIR}}^c)$~\cite{Dai_2021_aff}.

We compute correlation volumes for not only the fused features and but also NIR features:
\begin{equation}
    V_s(x,y,k) = F_s^{\text{left}}(x,y)\cdot F_s^{\text{right}}(x+k,y), \, {s \in \{\text{fusion},\text{NIR}\}},
\end{equation}
where $(x, y)$ denotes the pixel location, $k$ is the disparity, and $\cdot$ represents the inner product.
We found that using the two correlation volumes for the fused features and the NIR features effectively utilize cross-spectral information, resulting in better depth accuracy as shown in Table~\ref{tab:ablation_costvolume}. 

We estimate a series of disparity maps $\{d_1, \dots, d_N\}$ using the GRU structure of the RAFT-Stereo network~\cite{lipson2021raft}. {By feeding $F_\text{fusion}^\text{left}$ as context feature in the RAFT-Stereo}, we alternate between the fused and NIR correlation volumes, $V_\text{fusion}$ and $V_\text{NIR}$, as input to the GRU at each iteration, inspired by {the cross-spectral time-of-flight imaging} method~\cite{brucker2024cross}.

Our method leverages spectral information from both RGB and NIR features, placing more emphasis on NIR images that are robust to environmental lighting. We fine-tune our model on the synthetic and real training datasets using the disparity reconstruction loss and the LiDAR loss, respectively. For more details on the loss functions and optimization, we refer to the Supplemental Document.

\begin{table}[t]
\small
\centering
\resizebox{\linewidth}{!}{%
\begin{tabularx}{\linewidth}{|c|>{\centering\arraybackslash}X|>{\centering\arraybackslash}X|}
\hline
\textbf{Methods} & \textbf{(a) Depth RMSE~[m]~$\downarrow$} & \textbf{(b) Detection mAP~$\uparrow$} \\ 
\hline
RGB & 8.943 & 0.756 \\ \hline
NIR & 9.646 & 0.703 \\ \hline
YCrCb~\cite{herrera2019color} & 8.528 & 0.571 \\ \hline
Bayesian\cite{zhao2020bayesian_fusion} & 9.516 & 0.745 \\ \hline
DarkVision\cite{Jin_2023_DarkVision} & 8.313 & 0.762  \\
\hline
Adaptive\cite{awad2019adaptive_rgb_nir} & 7.830 & 0.773 \\
\hline
VGG-NIR\cite{li2018infrared_vgg} &9.654 & 0.726  \\
\hline

HSV(our baseline)~\cite{fredembach2008colouring_hsv} & 7.692 &0.744  \\
\Xhline{2\arrayrulewidth}\Xhline{0.2pt}
\textbf{Ours} & \textbf{7.567} & \textbf{0.809} \\
\hline
\end{tabularx}
}
\caption{\textbf{Comparison of image fusion methods for pretrained RAFT-stereo~\cite{lipson2021raft} and YOLO~\cite{yolov8_ultralytics}.} Our RGB-NIR image fusion method outperforms other image-fusion methods for stereo depth estimation and object detection. }
\label{tab:fusion_comparison}
\end{table}

\section{Results}
\label{sec:results}

We present the evaluation of our proposed pixel-aligned RGB-NIR image fusion and feature fusion methods, along with comprehensive ablation studies. 

\subsection{Pixel-aligned RGB-NIR Image Fusion}
\label{sec:results_image_fusion}
To evaluate our pixel-aligned RGB-NIR image fusion method in Section~\ref{sec:image_fusion}, we conducted assessments across three downstream tasks: depth estimation, object detection, and structure from motion.

\paragraph{Depth Estimation}
Figure~\ref{fig:color_compare_depth} shows that using our fused image as an input to the pre-trained stereo depth estimation network~\cite{lipson2021raft} yields improved results over single-modality predictions using only RGB or NIR images.
Table~\ref{tab:fusion_comparison}(a) shows quantitative analysis compared to those single-modality predictions as well as using other RGB-NIR image fusion methods. Our proposed RGB-NIR image fusion outperforms all the baselines. Refer to the Supplemental Document for corresponding qualitative results.
Our method also outperforms the baselines on another stereo-depth network~\cite{Li_2022_CREStereo} and monocular depth-estimation network~\cite{yang2023depthanything}, which can be found in the Supplemental Document.

\begin{figure}[t]
  \centering
  \includegraphics[width=\linewidth]{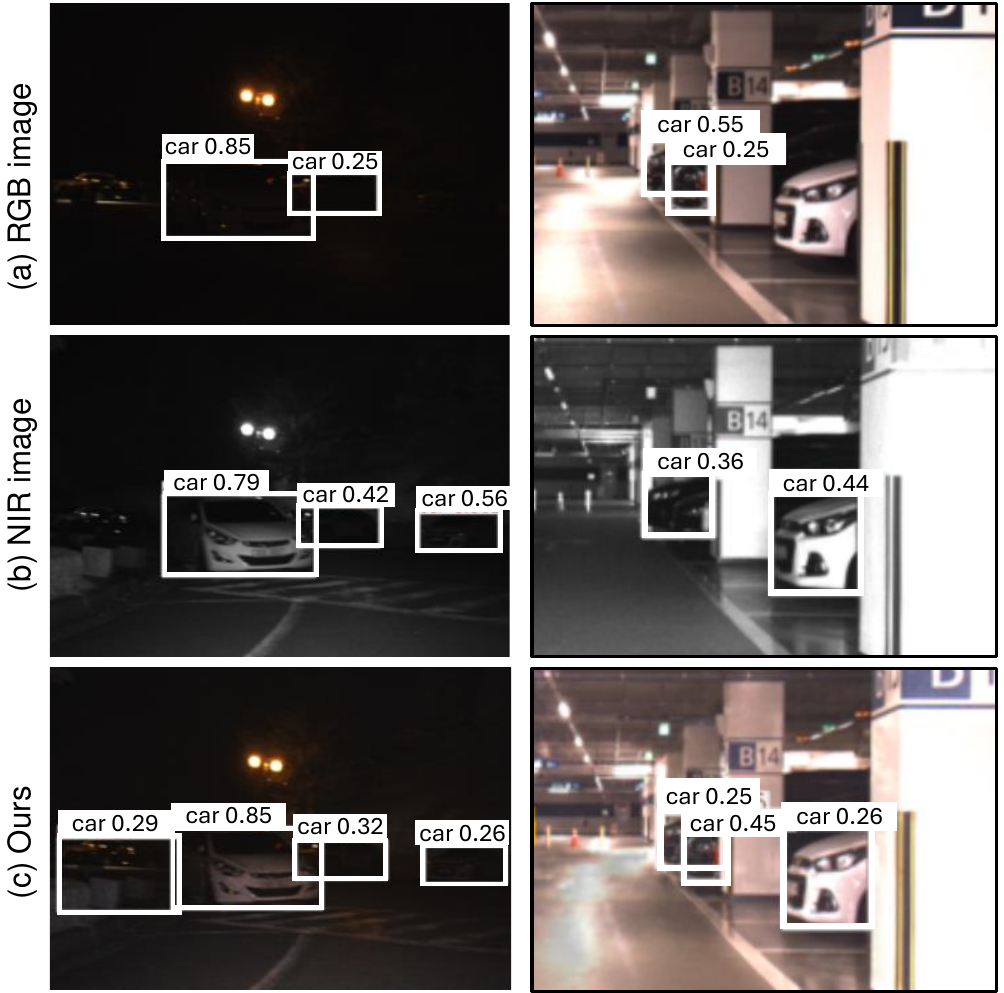}
  \caption{\textbf{Image fusion for pretrained YOLO~\cite{yolov8_ultralytics}}. {
(a) Using the RGB images struggle with object detection in scenes with low light conditions or high saturation due to the under- and over-exposed regions by the limited dynamic range of the cameras.
(b) NIR images suffer from challenges due to variations in material reflectance and the absence of detailed chromatic information for robust object detection. 
(c) Our RGB-NIR image fusion integrates the strengths of both RGB and NIR modalities, enabling robust object detection without retraining the model.}
}
\vspace{-2mm}
  \label{fig:image_fusion}
\end{figure}

\paragraph{Object Detection}
We use the YOLO object detector~\cite{yolov8_ultralytics} to test our RGB-NIR image fusion method. 
Figure~\ref{fig:image_fusion} and Table~\ref{tab:fusion_comparison}(b) shows that our image fusion method consistently outperforms single-modality predictions of either RGB or NIR, and other RGB-NIR image fusion methods.

\begin{figure}[t]
  \centering
  \includegraphics[width=\linewidth]{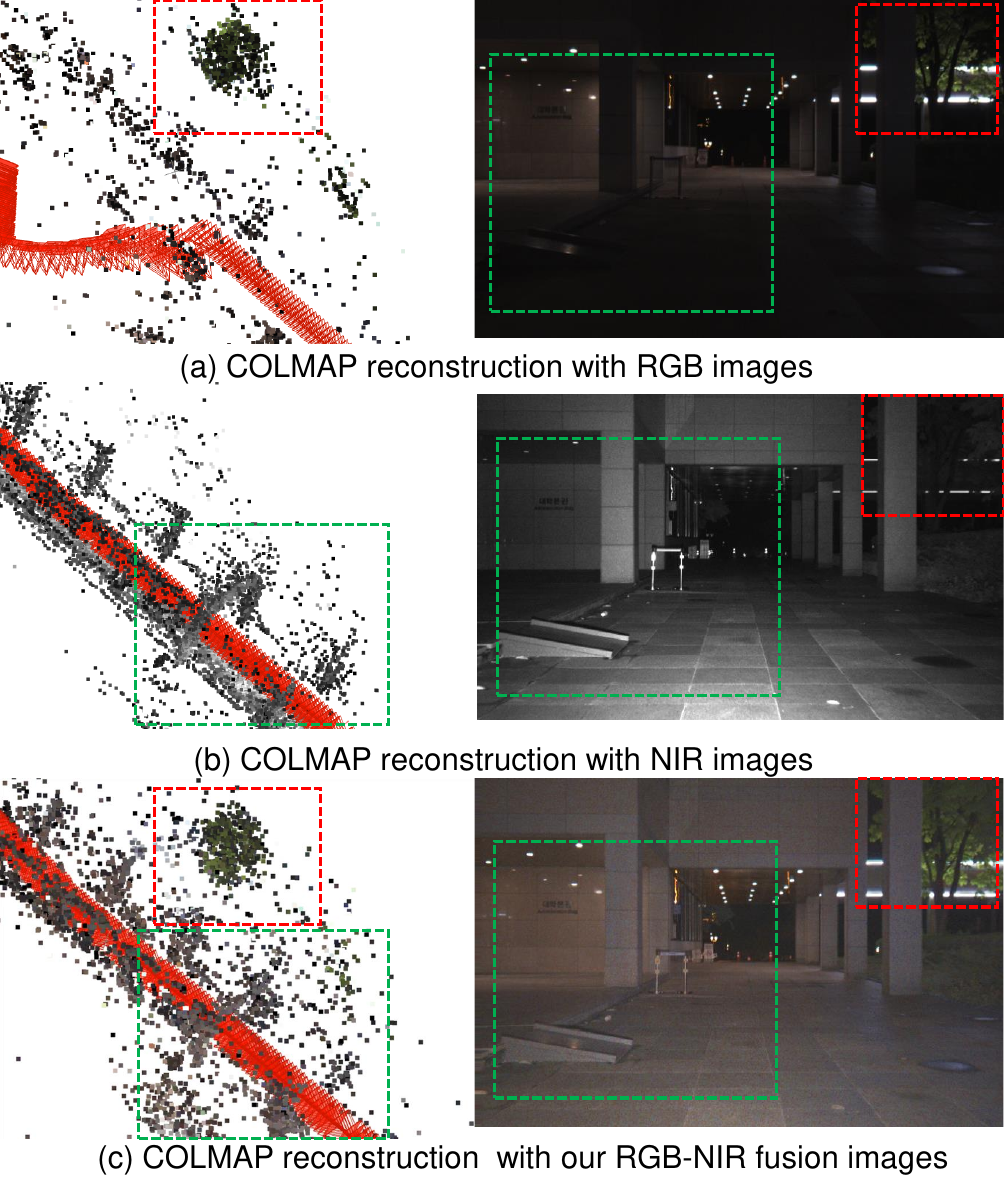}
  \caption{\textbf{Image fusion for structure from motion~\cite{schonberger2016sfm_colmap}.} 
  {
  Compared to using (a) RGB images and (b) NIR images only, our RGB-NIR fused images (c) enable accurate reconstruction of both regions indicated by red and green rectangles. 
  The green region is located closer to the imaging system, where active illumination enhances the visibility of distinct features in the NIR domain. The red region is situated farther from the imaging system, where features are well illuminated in the RGB domain due to ambient light.
  (c) Our RGB-NIR image fusion integrates these complementary features, enabling robust 3D reconstruction.}
  }
  \vspace{-3mm}
  \label{fig:fusion_colmap}
\end{figure}

\paragraph{Structure from Motion}
Figure~\ref{fig:fusion_colmap} shows that our RGB-NIR fused images enables robust reconstruction for a structure-from-motion method~\cite{schonberger2016sfm_colmap} on challenging lighting conditions, where RGB and NIR images provide different scene visibility.
This suggested the potential utility of our fused RGB-NIR images for downstream applications, including scene reconstruction and view synthesis.

\begin{figure*}[t]
  \centering
  \includegraphics[width=\textwidth]{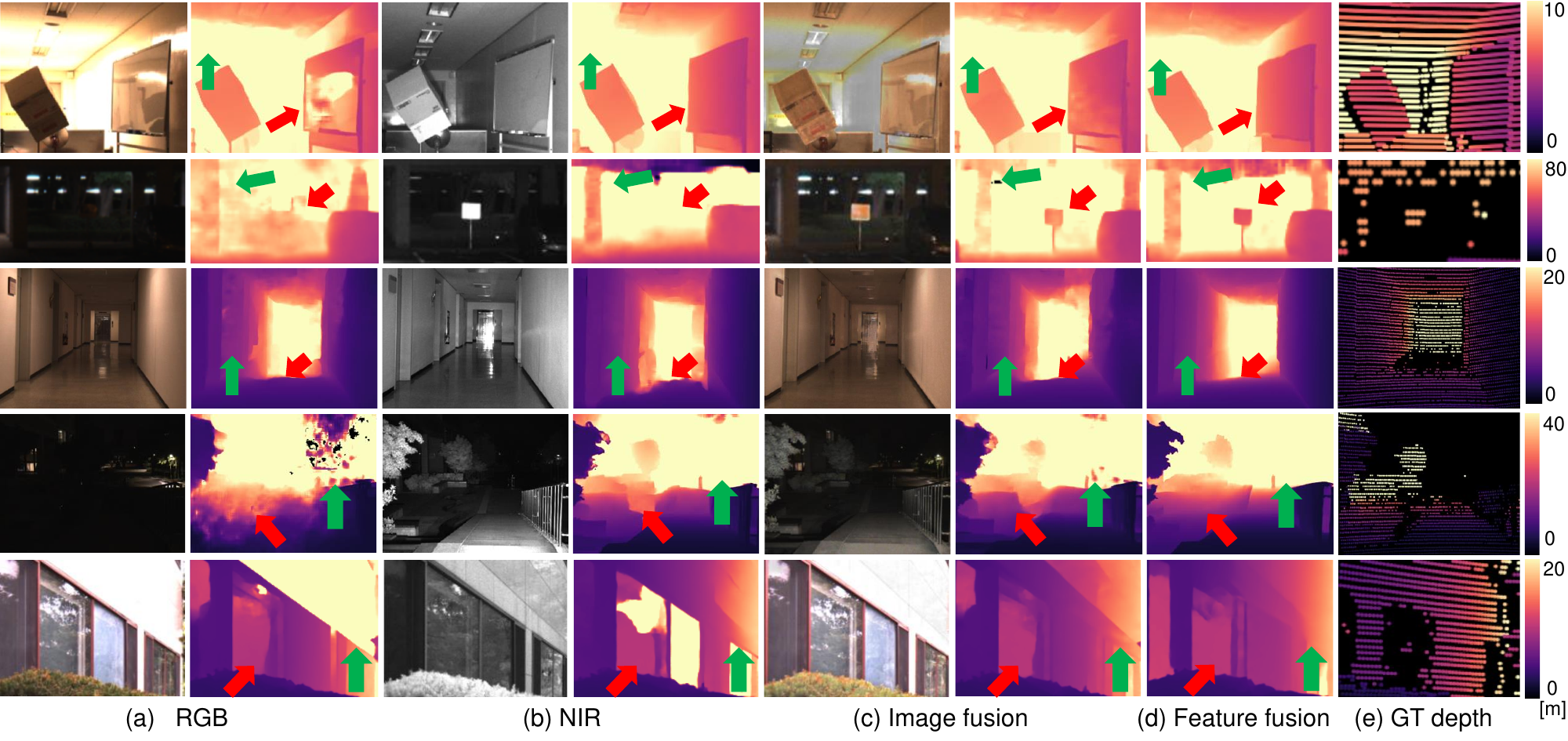}
  \caption{\textbf{RGB-NIR feature-based depth estimation.} 
  (a)\&(b) Using single-modality RGB/NIR images for stereo depth estimation~\cite{lipson2021raft} often fails to obtain high-quality depth. 
  (c) Our RGB-NIR image-fusion method (Section~\ref{sec:image_fusion}) mitigates the issue without finetuning the RAFT stereo, although it does not fully compensate for the strong saturation seen in (a) and (b).
(d) Our RGB-NIR feature-based method (Section~\ref{sec:feature_fusion}) further improves accuracy using an attentional method, at the cost of additional training.}  
  \label{fig:output_depth}
  \vspace{-3mm}
\end{figure*}

\begin{table}[t] 
\small 
\centering 
\resizebox{\linewidth}{!}{%
\begin{tabularx}{\linewidth}{|c|>{\centering\arraybackslash}X|} \hline

\textbf{Depth estimation model} & \textbf{Depth RMSE [m]}$\downarrow$ \\ \hline

RAFT-Stereo (RGB)-\cite{lipson2021raft}  & 8.943  \\ \hline
RAFT-Stereo (NIR)-\cite{lipson2021raft} & 9.646  \\ \hline

CS-Stereo~\cite{Zhi_2018_material_cross_spectral} & 8.941  \\ \hline 
CSPD~\cite{Guo_2023_cross_stereo_est_and_data} & 10.101 \\
\hline 

DPSNet~\cite{tian2023dps_polar_stereo} & 7.633  \\ \hline 
Image fusion (ours)-\cite{lipson2021raft} & 7.567 \\ \hline
\textbf{Ours} & \textbf{6.747}  \\ \hline
\end{tabularx} 
} 
\caption{\textbf{RGB-NIR feature-based depth estimation.} 
Our feature-based depth estimation (Section~\ref{sec:feature_fusion}) outperforms using the single-modality RGB/NIR inputs, different RGB-NIR stereo depth estimation model~\cite{Zhi_2018_material_cross_spectral}, different cross-spectral methods~\cite{tian2023dps_polar_stereo,Guo_2023_cross_stereo_est_and_data} which we fine-tuned with our synthetic and real dataset, and our image-fusion method (Section~\ref{sec:image_fusion}). 
} \label{tab:depth_comparison_multispectral} 
\vspace{-2mm}
\end{table}

\subsection{Pixel-aligned RGB-NIR Feature Fusion}
\label{sec:results_feature_fusion}
Figure~\ref{fig:output_depth} shows that our feature-based RGB-NIR stereo method (Section~\ref{sec:feature_fusion}) outperforms not only the single-modality predictions using either RGB or NIR channels, but also our RGB-NIR image-fusion method (Section~\ref{sec:image_fusion}).
While using single-modality RGB/NIR images easily struggle with saturated areas and shadows, our RGB-NIR fused images provide better depth estimation without finetuning. 
Our feature-based stereo depth estimation method further improves reconstruction quality by effectively exploiting the strengths of RGB and NIR channels via finetuning.
Table~\ref{tab:depth_comparison_multispectral} shows quantitative comparison over six baseline methods: RGB-only, NIR-only, our image-fusion method, and four multi-spectral fusion architectures~\cite{Zhi_2018_material_cross_spectral,Guo_2023_cross_stereo_est_and_data,tian2023dps_polar_stereo}. 
For the multi-spectral fusion methods~\cite{Guo_2023_cross_stereo_est_and_data,tian2023dps_polar_stereo}, we use their fusion architectures and retrain them on our dataset for fair comparison.
Refer to the Supplemental Document for details.

\begin{table}[t]
\centering
\resizebox{\linewidth}{!}{%
\begin{tabular}{c|c}
\hline
\textbf{Correlation volumes for disparity estimation} & \makecell{ \textbf{Depth}\\ \textbf{error } } [m]$\downarrow$  \\
\hline

Fusion correlation volumes only  & 7.440 \\
Alternating RGB-NIR correlation volumes & 8.571 \\
Alternating Fusion-RGB-NIR correlation volumes & 7.426 \\
\textbf{Alternating Fusion-NIR correlation volumes} & \textbf{6.747} \\
\hline
\end{tabular}
}
\caption{\textbf{Ablation study on alternating correlation volumes} 
Using fused and NIR correlation volumes in an alternating manner provides the highest depth-estimation accuracy.} 
\vspace{-3mm}
\label{tab:ablation_costvolume}
\end{table}

\paragraph{Alternating Correlation Volumes} 
In Section~\ref{sec:feature_fusion}, we investigate the effect of using {different types of correlation volumes for stereo disparity matching}. Specifically, we experiment with several strategies: using only the fusion correlation volume, alternating between RGB and NIR correlation volumes, alternating among fusion, RGB, and NIR correlation volumes, and alternating between fusion and NIR correlation volumes. Table~\ref{tab:ablation_costvolume} shows that alternating fusion-NIR correlation volumes yields the highest accuracy compared to the other methods.

\section{Conclusion}
\label{sec:conclusion}

In this paper, we have presented a pixel-aligned RGB-NIR stereo imaging system, collect datasets, and develop RGB-NIR image fusion for pretrained models and feature fusion methods with additional training. 
We demonstrate the effectiveness of pixel-aligned RGB-NIR images and our RGB-NIR fusion methods for three downstream tasks: depth estimation, object detection, and structure from motion.  
We hope our work shows the potential of pixel-aligned RGB-NIR imaging.

\paragraph{Limitations and Future Work}
We have demonstrated 3D imaging applications of pixel-aligned RGB-NIR images. 
A potentially more interesting future work is to reproduce the success of RGB-image generative models for pixel-aligned RGB-NIR images. 

Another interesting future research is to improve the accuracy of pixel-aligned RGB-NIR 3D imaging with the LiDAR measurements that provide complementary information about the scene via time-of-flight imaging.

\clearpage
{\small
\bibliographystyle{ieee_fullname}
\bibliography{references}

\begin{thebibliography}{10}\itemsep=-1pt

\bibitem{awad2019adaptive_rgb_nir}
Mohamed Awad, Ahmed Elliethy, and Hussein~A Aly.
\newblock Adaptive near-infrared and visible fusion for fast image enhancement.
\newblock {\em IEEE Transactions on Computational Imaging}, 6:408--418, 2019.

\bibitem{bouguet2004camera_calibration}
Jean-Yves Bouguet.
\newblock Camera calibration toolbox for matlab.
\newblock {\em http://www. vision. caltech. edu/bouguetj/calib\_doc/}, 2004.

\bibitem{broadbent2023cognitive_driver_eyetracking}
David~P Broadbent, Giorgia D'Innocenzo, Toby~J Ellmers, Justin Parsler, Andre~J Szameitat, and Daniel~T Bishop.
\newblock Cognitive load, working memory capacity and driving performance: A preliminary fnirs and eye tracking study.
\newblock {\em Transportation research part F: traffic psychology and behaviour}, 92:121--132, 2023.

\bibitem{brown2011_rgbnir_filter}
Matthew Brown and Sabine S{\"u}sstrunk.
\newblock Multi-spectral sift for scene category recognition.
\newblock In {\em CVPR 2011}, pages 177--184. IEEE, 2011.

\bibitem{brucker2024cross}
Samuel Brucker, Stefanie Walz, Mario Bijelic, and Felix Heide.
\newblock Cross-spectral gated-rgb stereo depth estimation.
\newblock In {\em Proceedings of the IEEE/CVF Conference on Computer Vision and Pattern Recognition}, pages 21654--21665, 2024.

\bibitem{chebrolu2017agricultural}
Nived Chebrolu, Philipp Lottes, Alexander Schaefer, Wera Winterhalter, Wolfram Burgard, and Cyrill Stachniss.
\newblock Agricultural robot dataset for plant classification, localization and mapping on sugar beet fields.
\newblock {\em The International Journal of Robotics Research}, 36(10):1045--1052, 2017.

\bibitem{choe2018ranus}
Gyeongmin Choe, Seong-Heum Kim, Sunghoon Im, Joon-Young Lee, Srinivasa~G Narasimhan, and In~So Kweon.
\newblock Ranus: Rgb and nir urban scene dataset for deep scene parsing.
\newblock {\em IEEE Robotics and Automation Letters}, 3(3):1808--1815, 2018.

\bibitem{Choi_Kim_Hwang_Park_Yoon_An_Kweon_2018}
Yukyung Choi, Namil Kim, Soonmin Hwang, Kibaek Park, Jae~Shin Yoon, Kyounghwan An, and In~So Kweon.
\newblock Kaist multi-spectral day/night data set for autonomous and assisted driving.
\newblock {\em IEEE Transactions on Intelligent Transportation Systems}, 19(3):934–948, Mar. 2018.

\bibitem{Connah_Drew_Finlayson_2014_grad_poisson}
David Connah, Mark~Samuel Drew, and Graham~David Finlayson.
\newblock Spectral edge image fusion: Theory and applications.
\newblock In David Fleet, Tomas Pajdla, Bernt Schiele, and Tinne Tuytelaars, editors, {\em Computer Vision – ECCV 2014}, page 65–80, Cham, 2014. Springer International Publishing.

\bibitem{couprie2013nyu_dataset}
Camille Couprie, Cl{\'e}ment Farabet, Laurent Najman, and Yann Lecun.
\newblock Indoor semantic segmentation using depth information.
\newblock In {\em First International Conference on Learning Representations (ICLR 2013)}, pages 1--8, 2013.

\bibitem{dai2017scannet}
Angela Dai, Angel~X Chang, Manolis Savva, Maciej Halber, Thomas Funkhouser, and Matthias Nie{\ss}ner.
\newblock Scannet: Richly-annotated 3d reconstructions of indoor scenes.
\newblock In {\em Proceedings of the IEEE conference on computer vision and pattern recognition}, pages 5828--5839, 2017.

\bibitem{Dai_2021_aff}
Yimian Dai, Fabian Gieseke, Stefan Oehmcke, Yiquan Wu, and Kobus Barnard.
\newblock Attentional feature fusion.
\newblock In {\em 2021 IEEE Winter Conference on Applications of Computer Vision (WACV)}, page 3559–3568, Waikoloa, HI, USA, Jan. 2021. IEEE.

\bibitem{deevi2024rgbx}
Sri~Aditya Deevi, Connor Lee, Lu Gan, Sushruth Nagesh, Gaurav Pandey, and Soon-Jo Chung.
\newblock Rgb-x object detection via scene-specific fusion modules.
\newblock In {\em Proceedings of the IEEE/CVF Winter Conference on Applications of Computer Vision}, pages 7366--7375, 2024.

\bibitem{fredembach2008colouring_hsv}
Cl{\'e}ment Fredembach and Sabine S{\"u}sstrunk.
\newblock Colouring the near-infrared.
\newblock In {\em Color and imaging conference}, volume~16, pages 176--182. Society of Imaging Science and Technology, 2008.

\bibitem{Gruber_2019_gated2depth}
Tobias Gruber, Frank Julca-Aguilar, Mario Bijelic, and Felix Heide.
\newblock Gated2depth: Real-time dense lidar from gated images.
\newblock In {\em Proceedings of the IEEE/CVF International Conference on Computer Vision}, pages 1506--1516, 2019.

\bibitem{Guo_2023_cross_stereo_est_and_data}
Yubin Guo, Xinlei Qi, Jin Xie, Cheng-Zhong Xu, and Hui Kong.
\newblock Unsupervised cross-spectrum depth estimation by visible-light and thermal cameras.
\newblock {\em IEEE Transactions on Intelligent Transportation Systems}, 24(10):10937–10947, Oct. 2023.

\bibitem{hansen2010_rgbnirfacephotometric}
Mark~F Hansen, Gary~A Atkinson, Lyndon~N Smith, and Melvyn~L Smith.
\newblock 3d face reconstructions from photometric stereo using near infrared and visible light.
\newblock {\em Computer Vision and Image Understanding}, 114(8):942--951, 2010.

\bibitem{herrera2021tophatnir}
Mar{\'\i}a Herrera-Arellano, Hayde Peregrina-Barreto, and Iv{\'a}n Terol-Villalobos.
\newblock Visible-nir image fusion based on top-hat transform.
\newblock {\em IEEE Transactions on Image Processing}, 30:4962--4972, 2021.

\bibitem{herrera2019color}
Mar{\'\i}a~A Herrera-Arellano, Hayde Peregrina-Barreto, and Iv{\'a}n Terol-Villalobos.
\newblock Color outdoor image enhancement by v-nir fusion and weighted luminance.
\newblock In {\em 2019 IEEE International Autumn Meeting on Power, Electronics and Computing (ROPEC)}, pages 1--6. IEEE, 2019.

\bibitem{hong2022reflection}
Yuchen Hong, Youwei Lyu, Si Li, Gang Cao, and Boxin Shi.
\newblock Reflection removal with nir and rgb image feature fusion.
\newblock {\em IEEE Transactions on Multimedia}, 2022.

\bibitem{huang2023polarization_nir_3d}
Xuanlun Huang, Chenyang Wu, Xiaolan Xu, Baishun Wang, Sui Zhang, Chihchiang Shen, Chiennan Yu, Jiaxing Wang, Nan Chi, Shaohua Yu, et~al.
\newblock Polarization structured light 3d depth image sensor for scenes with reflective surfaces.
\newblock {\em Nature Communications}, 14(1):6855, 2023.

\bibitem{hwang2015multispectral}
Soonmin Hwang, Jaesik Park, Namil Kim, Yukyung Choi, and In So~Kweon.
\newblock Multispectral pedestrian detection: Benchmark dataset and baseline.
\newblock In {\em Proceedings of the IEEE conference on computer vision and pattern recognition}, pages 1037--1045, 2015.

\bibitem{jang2017colour_fusion_dehazing}
Dong-Won Jang and Rae-Hong Park.
\newblock Colour image dehazing using near-infrared fusion.
\newblock {\em IET Image Processing}, 11(8):587--594, 2017.

\bibitem{Jin_2023_DarkVision}
Shuangping Jin, Bingbing Yu, Minhao Jing, Yi Zhou, Jiajun Liang, and Renhe Ji.
\newblock Darkvisionnet: Low-light imaging via rgb-nir fusion with deep inconsistency prior.
\newblock In {\em Proceedings of the AAAI Conference on Artificial Intelligence}, volume~36, pages 1104--1112, 2022.

\bibitem{Jung_2020_FusionNet}
Cheolkon Jung, Kailong Zhou, and Jiawei Feng.
\newblock Fusionnet: Multispectral fusion of rgb and nir images using two stage convolutional neural networks.
\newblock {\em IEEE Access}, 8:23912–23919, 2020.

\bibitem{kim2012_3dhyperspectral}
Min~H Kim, Todd~Alan Harvey, David~S Kittle, Holly Rushmeier, Julie Dorsey, Richard~O Prum, and David~J Brady.
\newblock 3d imaging spectroscopy for measuring hyperspectral patterns on solid objects.
\newblock {\em ACM Transactions on Graphics (TOG)}, 31(4):1--11, 2012.

\bibitem{kim2018multispectral}
Namil Kim, Yukyung Choi, Soonmin Hwang, and In~So Kweon.
\newblock Multispectral transfer network: Unsupervised depth estimation for all-day vision.
\newblock In {\em Proceedings of the AAAI Conference on Artificial Intelligence}, volume~32, 2018.

\bibitem{Lee_Cho_Shin_Kim_Myung_2022_Vivid}
Alex~Junho Lee, Younggun Cho, Young-sik Shin, Ayoung Kim, and Hyun Myung.
\newblock Vivid++ : Vision for visibility dataset.
\newblock {\em IEEE Robotics and Automation Letters}, 7(3):6282–6289, July 2022.

\bibitem{li2018infrared_vgg}
Hui Li, Xiao-Jun Wu, and Josef Kittler.
\newblock Infrared and visible image fusion using a deep learning framework.
\newblock In {\em 2018 24th international conference on pattern recognition (ICPR)}, pages 2705--2710. IEEE, 2018.

\bibitem{Li_2022_CREStereo}
Jiankun Li, Peisen Wang, Pengfei Xiong, Tao Cai, Ziwei Yan, Lei Yang, Jiangyu Liu, Haoqiang Fan, and Shuaicheng Liu.
\newblock Practical stereo matching via cascaded recurrent network with adaptive correlation.
\newblock In {\em 2022 IEEE/CVF Conference on Computer Vision and Pattern Recognition (CVPR)}, page 16242–16251, New Orleans, LA, USA, June 2022. IEEE.

\bibitem{Li_2013_guidedfiltering_fusion}
Shutao Li, Xudong Kang, and Jianwen Hu.
\newblock Image fusion with guided filtering.
\newblock {\em IEEE Transactions on Image Processing}, 22(7):2864–2875, July 2013.

\bibitem{li2007nirilumface}
Stan~Z Li, RuFeng Chu, ShengCai Liao, and Lun Zhang.
\newblock Illumination invariant face recognition using near-infrared images.
\newblock {\em IEEE Transactions on pattern analysis and machine intelligence}, 29(4):627--639, 2007.

\bibitem{li2021near_polar_3d}
Xuan Li, Fei Liu, Pingli Han, Shichao Zhang, and Xiaopeng Shao.
\newblock Near-infrared monocular 3d computational polarization imaging of surfaces exhibiting nonuniform reflectance.
\newblock {\em Optics Express}, 29(10):15616--15630, 2021.

\bibitem{li_2021_viz_nir_fusion_spectrum}
Zhuo Li, Hai-Miao Hu, Wei Zhang, Shiliang Pu, and Bo Li.
\newblock Spectrum characteristics preserved visible and near-infrared image fusion algorithm.
\newblock {\em IEEE Transactions on Multimedia}, 23:306--319, 2021.

\bibitem{liang2019rgbnir_UCSSM}
Mingyang Liang, Xiaoyang Guo, Hongsheng Li, Xiaogang Wang, and You Song.
\newblock Unsupervised cross-spectral stereo matching by learning to synthesize.
\newblock In {\em Proceedings of the AAAI Conference on Artificial Intelligence}, volume~33, pages 8706--8713, 2019.

\bibitem{lipson2021raft}
Lahav Lipson, Zachary Teed, and Jia Deng.
\newblock Raft-stereo: Multilevel recurrent field transforms for stereo matching.
\newblock In {\em 2021 International Conference on 3D Vision (3DV)}, pages 218--227. IEEE, 2021.

\bibitem{Liu_Huang_2010_EM}
Gang Liu and Guohong Huang.
\newblock Color fusion based on em algorithm for ir and visible image.
\newblock In {\em 2010 The 2nd International Conference on Computer and Automation Engineering (ICCAE)}, volume~2, page 253–258, Feb. 2010.

\bibitem{liu2016multispectral_pedestrian}
Jingjing Liu, Shaoting Zhang, Shu Wang, and Dimitris~N Metaxas.
\newblock Multispectral deep neural networks for pedestrian detection.
\newblock In {\em 27th British Machine Vision Conference, BMVC 2016}, 2016.

\bibitem{mayer2016sceneflow}
Nikolaus Mayer, Eddy Ilg, Philip Hausser, Philipp Fischer, Daniel Cremers, Alexey Dosovitskiy, and Thomas Brox.
\newblock A large dataset to train convolutional networks for disparity, optical flow, and scene flow estimation.
\newblock In {\em Proceedings of the IEEE conference on computer vision and pattern recognition}, pages 4040--4048, 2016.

\bibitem{monno2018single_sensor_rgb_nir}
Yusuke Monno, Hayato Teranaka, Kazunori Yoshizaki, Masayuki Tanaka, and Masatoshi Okutomi.
\newblock Single-sensor rgb-nir imaging: High-quality system design and prototype implementation.
\newblock {\em IEEE Sensors Journal}, 19(2):497--507, 2018.

\bibitem{mortimer_2024_goose}
Peter Mortimer, Raphael Hagmanns, Miguel Granero, Thorsten Luettel, Janko Petereit, and Hans-Joachim Wuensche.
\newblock The goose dataset for perception in unstructured environments.
\newblock 2024.

\bibitem{park2016color_restore_rgbn}
Chulhee Park and Moon~Gi Kang.
\newblock Color restoration of rgbn multispectral filter array sensor images based on spectral decomposition.
\newblock {\em Sensors}, 16(5):719, 2016.

\bibitem{poggi2022cross}
Matteo Poggi, Pierluigi~Zama Ramirez, Fabio Tosi, Samuele Salti, Stefano Mattoccia, and Luigi Di~Stefano.
\newblock Cross-spectral neural radiance fields.
\newblock In {\em 2022 International Conference on 3D Vision (3DV)}, pages 606--616. IEEE, 2022.

\bibitem{quan1999linear_pnp}
Long Quan and Zhongdan Lan.
\newblock Linear n-point camera pose determination.
\newblock {\em IEEE Transactions on pattern analysis and machine intelligence}, 21(8):774--780, 1999.

\bibitem{yolov8_ultralytics}
Joseph Redmon, Santosh Divvala, Ross Girshick, and Ali Farhadi.
\newblock You only look once: Unified, real-time object detection.
\newblock In {\em 2016 IEEE Conference on Computer Vision and Pattern Recognition (CVPR)}, pages 779--788, 2016.

\bibitem{schonberger2016sfm_colmap}
Johannes~L Schonberger and Jan-Michael Frahm.
\newblock Structure-from-motion revisited.
\newblock In {\em Proceedings of the IEEE conference on computer vision and pattern recognition}, pages 4104--4113, 2016.

\bibitem{shibata2016versatile_nir}
Takashi Shibata, Masayuki Tanaka, and Masatoshi Okutomi.
\newblock Versatile visible and near-infrared image fusion based on high visibility area selection.
\newblock {\em Journal of Electronic Imaging}, 25(1):013016--013016, 2016.

\bibitem{Shin_Park_Kweon_2023}
Ukcheol Shin, Jinsun Park, and In~So Kweon.
\newblock Deep depth estimation from thermal image.
\newblock In {\em 2023 IEEE/CVF Conference on Computer Vision and Pattern Recognition (CVPR)}, page 1043–1053, Vancouver, BC, Canada, June 2023. IEEE.

\bibitem{su2021multidenoising}
Haonan Su, Cheolkon Jung, and Long Yu.
\newblock Multi-spectral fusion and denoising of color and near-infrared images using multi-scale wavelet analysis.
\newblock {\em Sensors}, 21(11):3610, 2021.

\bibitem{takumi2017multispectral_detection}
Karasawa Takumi, Kohei Watanabe, Qishen Ha, Antonio Tejero-De-Pablos, Yoshitaka Ushiku, and Tatsuya Harada.
\newblock Multispectral object detection for autonomous vehicles.
\newblock In {\em Proceedings of the on Thematic Workshops of ACM Multimedia 2017}, pages 35--43, 2017.

\bibitem{tian2023dps_polar_stereo}
Chaoran Tian, Weihong Pan, Zimo Wang, Mao Mao, Guofeng Zhang, Hujun Bao, Ping Tan, and Zhaopeng Cui.
\newblock Dps-net: Deep polarimetric stereo depth estimation.
\newblock In {\em Proceedings of the IEEE/CVF International Conference on Computer Vision}, pages 3569--3579, 2023.

\bibitem{toet2016dataset_nir_fusion_aligned}
Alexander Toet, Maarten~A Hogervorst, and Alan~R Pinkus.
\newblock The triclobs dynamic multi-band image data set for the development and evaluation of image fusion methods.
\newblock {\em PloS one}, 11(12):e0165016, 2016.

\bibitem{toet1996false_color_fusion}
Alexander Toet and Jan Walraven.
\newblock New false color mapping for image fusion.
\newblock {\em Optical engineering}, 35(3):650--658, 1996.

\bibitem{valada2017rgb_nir_seg_robot}
Abhinav Valada, Gabriel~L Oliveira, Thomas Brox, and Wolfram Burgard.
\newblock Deep multispectral semantic scene understanding of forested environments using multimodal fusion.
\newblock In {\em 2016 International Symposium on Experimental Robotics}, pages 465--477. Springer, 2017.

\bibitem{Walz_2023_gatedstereo}
Stefanie Walz, Mario Bijelic, Andrea Ramazzina, Amanpreet Walia, Fahim Mannan, and Felix Heide.
\newblock Gated stereo: Joint depth estimation from gated and wide-baseline active stereo cues.
\newblock In {\em 2023 IEEE/CVF Conference on Computer Vision and Pattern Recognition (CVPR)}, page 13252–13262, Vancouver, BC, Canada, June 2023. IEEE.

\bibitem{Wang_Wang_Yang_Fang_Wan_2023}
Linbo Wang, Tao Wang, Deyun Yang, Xianyong Fang, and Shaohua Wan.
\newblock Near-infrared fusion for deep lightness enhancement.
\newblock {\em International Journal of Machine Learning and Cybernetics}, 14(5):1621–1633, May 2023.

\bibitem{xia2018gibson_mobilerobot}
Fei Xia, Amir~R Zamir, Zhiyang He, Alexander Sax, Jitendra Malik, and Silvio Savarese.
\newblock Gibson env: Real-world perception for embodied agents.
\newblock In {\em Proceedings of the IEEE conference on computer vision and pattern recognition}, pages 9068--9079, 2018.

\bibitem{yadav2023habitat}
Karmesh Yadav, Ram Ramrakhya, Santhosh~Kumar Ramakrishnan, Theo Gervet, John Turner, Aaron Gokaslan, Noah Maestre, Angel~Xuan Chang, Dhruv Batra, Manolis Savva, et~al.
\newblock Habitat-matterport 3d semantics dataset.
\newblock In {\em Proceedings of the IEEE/CVF Conference on Computer Vision and Pattern Recognition}, pages 4927--4936, 2023.

\bibitem{yang2023depthanything}
Lihe Yang, Bingyi Kang, Zilong Huang, Xiaogang Xu, Jiashi Feng, and Hengshuang Zhao.
\newblock Depth anything: Unleashing the power of large-scale unlabeled data.
\newblock In {\em CVPR}, 2024.

\bibitem{Zhang_2024_TFDet}
Xue Zhang, Xiaohan Zhang, Jiangtao Wang, Jiacheng Ying, Zehua Sheng, Heng Yu, Chunguang Li, and Hui-Liang Shen.
\newblock Tfdet: Target-aware fusion for rgb-t pedestrian detection.
\newblock {\em IEEE Transactions on Neural Networks and Learning Systems}, page 1–15, 2024.

\bibitem{Gao_2019_hybrid_rgb_thermal_cam}
Yigong Zhang, Yicheng Gao, Shuo Gu, Yubin Guo, Minghao Liu, Zezhou Sun, Zhixing Hou, Hang Yang, Ying Wang, Jian Yang, Jean Ponce, and Hui Kong.
\newblock Build your own hybrid thermal/eo camera for autonomous vehicle.
\newblock In {\em 2019 International Conference on Robotics and Automation (ICRA)}, page 6555–6560, May 2019.

\bibitem{zhao2023metafusion}
Wenda Zhao, Shigeng Xie, Fan Zhao, You He, and Huchuan Lu.
\newblock Metafusion: Infrared and visible image fusion via meta-feature embedding from object detection.
\newblock In {\em Proceedings of the IEEE/CVF Conference on Computer Vision and Pattern Recognition}, pages 13955--13965, 2023.

\bibitem{zhao2020bayesian_fusion}
Zixiang Zhao, Shuang Xu, Chunxia Zhang, Junmin Liu, and Jiangshe Zhang.
\newblock Bayesian fusion for infrared and visible images.
\newblock {\em Signal Processing}, 177:107734, 2020.

\bibitem{Zhi_2018_material_cross_spectral}
Tiancheng Zhi, Bernardo~R Pires, Martial Hebert, and Srinivasa~G Narasimhan.
\newblock Deep material-aware cross-spectral stereo matching.
\newblock In {\em Proceedings of the IEEE conference on computer vision and pattern recognition}, pages 1916--1925, 2018.

\end{thebibliography}
}

\end{CJK}
\end{document}


\begin{CJK}{UTF8}{}
\CJKfamily{mj}

\title{
Pixel-aligned RGB-NIR Stereo Imaging and Dataset for Robot Vision
\\
$\text{ }$
\\
\large Supplemental Document
}

\author{
Jinnyeong Kim ~ ~ ~
Seung-Hwan Baek \\
POSTECH\\
}

\maketitle

In this supplemental document, we provide additional results and details in support of our findings in the main manuscript. 

\tableofcontents

\newpage

\section{Experimental Prototype}
\subsection{Hardware Stack}

\subsubsection{Imaging Setup}

Table~\ref{table:sensor_spec} summarizes the detailed specifications of the sensors used in our experimental prototype. To achieve RGB-NIR multispectral imaging and 3D geometric reconstruction, we implemented two pixel-aligned RGB-NIR cameras and a LiDAR system.

\begin{table}[h]
\centering

\begin{tabular}{|p{0.2\columnwidth}|p{0.08\columnwidth}|p{0.2\columnwidth}|p{0.4\columnwidth}|}
\hline
 \textbf{Sensor} & \textbf{Quantity} & \textbf{Product} & \textbf{Resolution}  \\ \hline
\textbf{RGB-NIR Camera} & 2 & JAI FS-1600D-10GE & \begin{tabular}[c]{@{}l@{}}
     
    180Hz 8bit 1440x1080 BayerRG image \\
    180Hz 8bit 1440x1080 NIR image
    \end{tabular} \\ \hline
\textbf{LiDAR} & 1 &  Ouster OS-1 & \begin{tabular}[c]{@{}l@{}}
    20Hz 2048x128 point cloud
    \end{tabular} \\ \hline
\textbf{IMU} & 1 & Ouster OS-1 IMU &  \begin{tabular}[c]{@{}l@{}}
    100Hz inertial data
    \end{tabular} \\ \hline
\end{tabular}
\caption{\textbf{Sensor specs of our imaging system}. }
  \label{table:sensor_spec}
\end{table}

\paragraph{RGB-NIR Stereo Camera Setup}
The RGB-NIR cameras (JAI FS-1600D-10GE) leverage a dichroic prism to simultaneously capture visible (RGB) and near-infrared (NIR) images, offering distinct advantages in robust feature extraction under varying illumination conditions. 
This dual-spectrum imaging capability facilitates applications such as material classification, vegetation analysis, and object detection in low-light environments.

Our system integrates two RGB-NIR cameras connected via RJ-45 interfaces using Ethernet cables.
These cameras support high frame rates and high-resolution image acquisition, with performance primarily determined by the transmission link speed. 
The JAI FS-1600D-10GE officially achieves up to 100 fps for RGB-NIR pixel-aligned imaging when operating on a 10 Gbps Ethernet connection.

\paragraph{NIR Active Illumination}
To enhance imaging in the NIR spectrum, an active illumination source, Advanced Illumination AL295-150850IC, emitting at an 850 nm wavelength, was employed. Table~\ref{table:active_illumination} shows detail specification of the NIR illumination. 
This illumination compensates for ambient lighting variability and improves the quality of the NIR channel, particularly in controlled or low-light environments.

\begin{table}[h] 
\centering 
\begin{tabular}{|p{0.4\columnwidth}|p{0.4\columnwidth}|} 
\hline 
\textbf{Specification} & \textbf{Parameter} \\ \hline
Length & 165.6 mm \\ \hline 
Weight &  68.9 g \\ \hline 
Wavelength & 850 nm \\ \hline 
Photobiological Risk Factor & Exempt (850 nm) \\ \hline 
Operating Temperature & 0 - 60°C \\ \hline 
Compliance & CE, RoHS, IEC 62471 \\ \hline 
IP Rating & IP50 \\ \hline 
Lumen Maintenance & L70 = 50,000 Hours \\ \hline 
\end{tabular} 
\caption{Specifications of the active illumination source (AL295-150850IC).} \label{table:active_illumination} 
\end{table}

\paragraph{3D LiDAR for Depth Ground Truth}
We use the Ouster OS-1 LiDAR to obtain accurate depth ground truth. 
This LiDAR supports up to 2048 samples with 128 channels, providing a depth resolution of 2048 × 128 for a full 360-degree rotation. 
The LiDAR can achieve a maximum frame rate of 20 fps at 1024 × 128 resolution and 10 fps at 2048 × 128 resolution. We utilized the 20 fps configuration for high-frequency datasets and the 10 fps configuration for lower-frequency datasets.
The built-in inertial measurement unit (IMU) measures angular velocity (radians/second) and linear acceleration (G) along the x, y, and z axes at up to 100 Hz, offering additional data for refining LiDAR point clouds. 
The LiDAR operates using 865 nm structured light, which interacts with NIR cameras and may be affected by external NIR illumination.
Nevertheless, advanced features, including multi-sensor crosstalk suppression and programmable settings, mitigate such interference, ensuring high-accuracy depth measurements under challenging illumination conditions.

\subsubsection{Mobile Robot}
Our imaging system is mounted on a mobile wheeled robot, the Agile-X Ranger Mini 2.0, which provides stable and efficient operation for large-scale data collection in both indoor and outdoor environments. 
The robot is equipped with a 4-wheel drive (4WD) system and features wheels capable of 180-degree rotation, offering exceptional maneuverability and the capability to navigate tight and complex spaces.

Table~\ref{table:ranger_spec} shows specifications of the mobile robot.
The Agile-X Ranger Mini 2.0 has a compact design with overall dimensions of $738 \text{mm}$ × $500 \text{mm}$ × $338 \text{mm}$ and an axle track of 494 mm, enabling smooth traversal across various terrains. 
It is powered by four 48 $V$ brushless toothed motors, each delivering a rated power of 600 $W$ and a torque of 22 $Nm$. 
The robot achieves a maximum speed of $2.6 m/s$ and can climb inclines up to 10° while carrying a maximum load of $150 \text{kg}$.
To accommodate different operational needs, the Ranger Mini 2.0 offers two battery configurations: a single battery setup providing 2–8 hours of operation and a multi-battery configuration supporting extended endurance. 
The lithium-ion batteries can be charged in as little as 1 hour, ensuring minimal downtime. The robot’s advanced suspension system and independent 4-wheel steering enable optimal performance on uneven surfaces and during high-precision maneuvers.

Our setup leverages the Ranger Mini 2.0's versatility to conduct imaging tasks across diverse settings, including roads, sidewalks, and interior spaces, ensuring comprehensive data collection for research in urban and natural environments. 
Its rugged design and IP54 rating make it suitable for challenging outdoor conditions while maintaining high stability and reliability for precise imaging.

\begin{table}[h]
\centering

\begin{tabular}{|c|c|}
\hline
\textbf{Specification} & \textbf{Parameter} \\ \hline
Maximum Payload        & 150 kg             \\ \hline
Maximum Speed          & 2.6 m/s            \\ \hline
Control Mode           & Remote controller / ROS \\ \hline
Steering Type          & 4-wheel steering   \\ \hline
Turning Radius         & 0 mm (Spin mode) / 810 mm (Ackermann mode) \\ \hline
Battery Life           & 2–8 hours          \\ \hline
Charging Time          & 1 hour             \\ \hline
\end{tabular}
\caption{\textbf{Specifications of Agile-X Ranger Mini 2.0}.}
\label{table:ranger_spec}
\end{table}

\subsubsection{Computational Link and Power Supply}

To manage the capture pipeline efficiently, we integrated a high-performance laptop (Asus ROG Zephyrus G14) equipped with an Nvidia RTX 4060 GPU (8GB). 
This setup provides the necessary computational power for real-time data processing and management of our imaging and LiDAR systems.
Since both the RGB-NIR cameras and the LiDAR connect via RJ-45 interfaces, we utilized an RJ-45 network switch hub to extend connectivity, accommodating up to 10 devices. 
This ensures seamless data transfer and system integration, despite the laptop's limited number of Ethernet ports.
Furthermore, to support the power demands of the entire imaging system, we mounted an external AC power bank onto the robot. 
This setup allows our system to operate continuously for over 3 hours without the need for recharging, ensuring sustained data collection in diverse environments.

\subsubsection{Calibration}

\paragraph{Stereo Pose Calibration}

To calibrate a stereo camera system comprising left and right cameras, a standard chessboard pattern is used as a known reference object. 
The calibration process begins by capturing a series of images of the chessboard from both cameras at multiple orientations. Using these images, point correspondences between the observed 2D chessboard corners in each camera image and the known 3D coordinates of the corners in the chessboard's coordinate system are established. 
The intrinsic parameters $K_\text{left}$ and $K_\text{right}$ for the left and right cameras, respectively, are estimated through this correspondence, capturing the focal length and principal point of each camera.
Next, the relative pose between the left and right cameras is determined. 
The extrinsic parameters $E_\text{right}$, comprising the rotation and translation that map 3D points from the left camera coordinate system to the right camera coordinate system, are computed. 
This calibration is essential for rectifying stereo image pairs and for accurate 3D reconstruction.
By using well-established algorithms~\cite{bouguet2004camera_calibration}, the intrinsic matrices $K_\text{left}$, $K_\text{right}$, and the extrinsic transformation $E_\text{right}$ are accurately estimated.

\paragraph{LiDAR Pose Calibration}

To calibrate the LiDAR and left camera, a well-structured real 3D scene is designed and captured to facilitate precise point correspondences between the LiDAR and camera images. 
Specifically, the left RGB image captured by the camera and the plane image generated from the LiDAR are analyzed to manually annotated corresponding points across both modalities. 
Each LiDAR coordinate, $(x_\text{LiDAR}, y_\text{LiDAR}, z_\text{LiDAR})$, is mapped into the camera coordinate system as $(x_\text{camera}, y_\text{camera}, z_\text{camera})$ using a transformation matrix $M_{\text{LiDAR}\rightarrow\text{camera}}$. 
The intrinsic camera matrix $K$ then projects these 3D points onto the 2D image plane, yielding image coordinates $(u, v)$, as expressed by the following equation:
\begin{equation}
    \begin{bmatrix} \hat{u} \\ \hat{v} \\ z_\text{left} \end{bmatrix} = K_\text{left} M_{\text{LiDAR}\rightarrow\text{camera}} \begin{bmatrix} x_\text{LiDAR} \\ y_\text{LiDAR} \\ z_\text{LiDAR} \\ 1 \end{bmatrix}, \begin{bmatrix}
        u \\ v
    \end{bmatrix} = \begin{bmatrix}
        \frac{\hat{u} }{z_\text{left}} \\
        \frac{\hat{v}}{z_\text{left}}
    \end{bmatrix}
\end{equation}

By collecting a sufficient number of $(x_\text{LiDAR}, y_\text{LiDAR}, z_\text{LiDAR})$ and $(u, v)$ correspondences and leveraging the known intrinsic parameters $K$, the transformation matrix $M_{\text{LiDAR}\rightarrow\text{camera}}$ can be estimated. 
This estimation process can be formulated as a Perspective-n-Point (PnP) problem, which seeks to determine the camera's extrinsic parameters (rotation $R$ and translation $t$) 
by minimizing the reprojection error between a set of 3D points and their corresponding 2D image projections.
The solution to the PnP problem, including the estimation of
$M_{\text{LiDAR}\rightarrow\text{camera}}$ follows the approach described in~\cite{quan1999linear_pnp}, which employs a linear formulation to efficiently estimate the transformation while minimizing the reprojection error.
Advanced solvers or iterative methods such as RANSAC can also be applied to enhance robustness in the presence of outliers.
\paragraph{Timing Calibration}

The JAI MultiSpectral Camera utilized in our system features Precision Timing Protocol (PTP), enabling time measurement with nanosecond-level accuracy. 
Internally, RGB CMOS and NIR CMOS of a camera are synchronized using PTP.
As we employed two separate cameras, it was essential to achieve synchronization between them. 
To this end, we activated a PTP-based Pulse Generator on each camera, which emits a trigger signal at precisely defined periods. 
By adjusting the delay settings of the Pulse Generators, we compared timestamps recorded in the response packets of both cameras to align them. 
This approach ensured that the synchronization between the cameras remained within 100 microseconds. 
Such a low level of time difference is negligible for 3D vision applications, eliminating the need for further temporal correction in downstream tasks. 
Additionally, as a precautionary guideline, we discarded any packets that exhibited a timing difference exceeding 1 millisecond.

\subsection{Software Stack}
\subsubsection{Sensor SDK Implementation}
\paragraph{JAI Fusion Camera} The JAI Fusion camera utilizes the eBUS SDK, which enables configuration of Ethernet-connected cameras and the creation of receiver sockets for the camera's data streams. 
Specifically, since this camera has two independent CMOS sensors within a single unit, configuring it as a stereo camera requires handling data from four separate streams. 
We integrated the C++ SDK library into our CMake project, setting up four distinct ports to create separate stream receiver objects. 
Each of these receiver objects operates on an independent thread, ensuring smooth data acquisition without blocking. 
The main thread monitors the timing of the collected image packets from the four streams, grouping them into a single frame if the timestamp difference is less than 1 ms. Once a complete frame is formed, it is published as a ROS2 topic for use by other processes. 
To minimize the time difference between the stereo cameras, we adjust the delay of the internal pulse generator of the cameras, based on the timing discrepancies measured for each frame.

\paragraph{Ouster OS-1} The Ouster LiDAR sensor is interfaced using its native Python3 SDK library. 
The LiDAR device is assigned a unique IP address, and its data packets are combined into a 360-degree scan using the SDK's packet integration tool. 
However, this integration tool does not accumulate IMU packets. 
To address this, we extended the packet integration functionality of the SDK, enabling it to integrate accumulated IMU packets into the 360-degree scanning data.

\subsubsection{Imaging System Pipeline Architecture}
Our imaging system operates on Ubuntu 22.04, utilizing Python 3.12 and ROS2 Humble to provide a robust and efficient platform for multi-sensor data acquisition and processing. 
The software integrates SDKs for stereo cameras and LiDAR sensors, facilitating seamless file I/O operations and an intuitive UX/UI for user interaction. 
The system is modularized into four primary components: Capture Trigger, Response Queue, Storage Management, and User Interface/User Experience (UI/UX).

The Capture Trigger module orchestrates the periodic activation of connected sensors, ensuring synchronized data acquisition from stereo cameras and LiDAR. 
It maintains precise time synchronization between the stereo cameras, enabling consistent stereo image capturing essential for depth estimation. 
The LiDAR module processes incoming LiDAR packets, aggregating them into a cohesive 360-degree point cloud representation.
Data from the sensors are funneled into the Response Queue, which manages the incoming RGB and LiDAR data streams. 
This module identifies and pairs sensor data with minimal time discrepancies, adhering to a predefined time difference threshold to maintain data integrity. 
Once paired, the data is forwarded to the Storage Management module, which handles the efficient local storage of synchronized multi-sensor information.
The UI/UX component provides users with a streamlined interface to control the imaging process. 
A single-button interface allows users to toggle sequential frame capturing (video recording) effortlessly, enabling them to concentrate on the robot's navigation tasks. 

The system, deployed on a robot-mounted laptop, supports capturing up to 20 frames per second. When real-time stereo depth estimation is activated, the system maintains a capture rate of up to 10 depth images per second, balancing performance and computational demands.
This software architecture ensures reliable synchronization, efficient data management, and user-friendly operation, making it well-suited for real-time robotic applications requiring high-fidelity multi-sensor imaging.
\section{Pixel-aligned RGB-NIR Datasets}
\subsection{Real Dataset}

\subsubsection{Details on Acquisition}
We captured 720×540 resolution BayerRG stereo images using JAI Fusion camera, restoring them to 3-channel RGB images via Bayer interpolation before saving. 
Simultaneously, 720×540 mono-channel NIR images were captured using JAI FS-1600D-10GE and saved alongside the RGB images. 
Our dataset comprises 28 video sequences captured at 10 Hz and 15 sequences at 5 Hz. Each video ranges from 100 to 6000 frames. 
Longer videos encompass continuous transitions through various environments and can be partitioned into shorter clips for specific applications.
Data collection occurred across diverse settings, including both indoor spaces like laboratories and lecture halls, and outdoor areas such as courtyards and walkways. 
Videos were recorded at different times of the day—morning, noon, evening, and night—to capture a wide range of lighting conditions. 
Some sequences feature transitions between indoor and outdoor environments, while others include dynamic scenarios like a vehicle entering an indoor parking garage from outside. 
Weather conditions during recording varied from sunny to overcast, adding to the dataset's versatility.
Our compact and lightweight camera setup eliminates the need for bulky beam splitters, allowing for extensive and flexible data collection without compromising the field of view. 
This design facilitates large-scale indoor and outdoor recording sessions, making the dataset suitable for applications in machine vision, autonomous navigation, and robotics.

\subsubsection{LiDAR Densification}
\label{sec:bpnet}

\begin{figure*}
\centering
  \includegraphics[width=\textwidth]{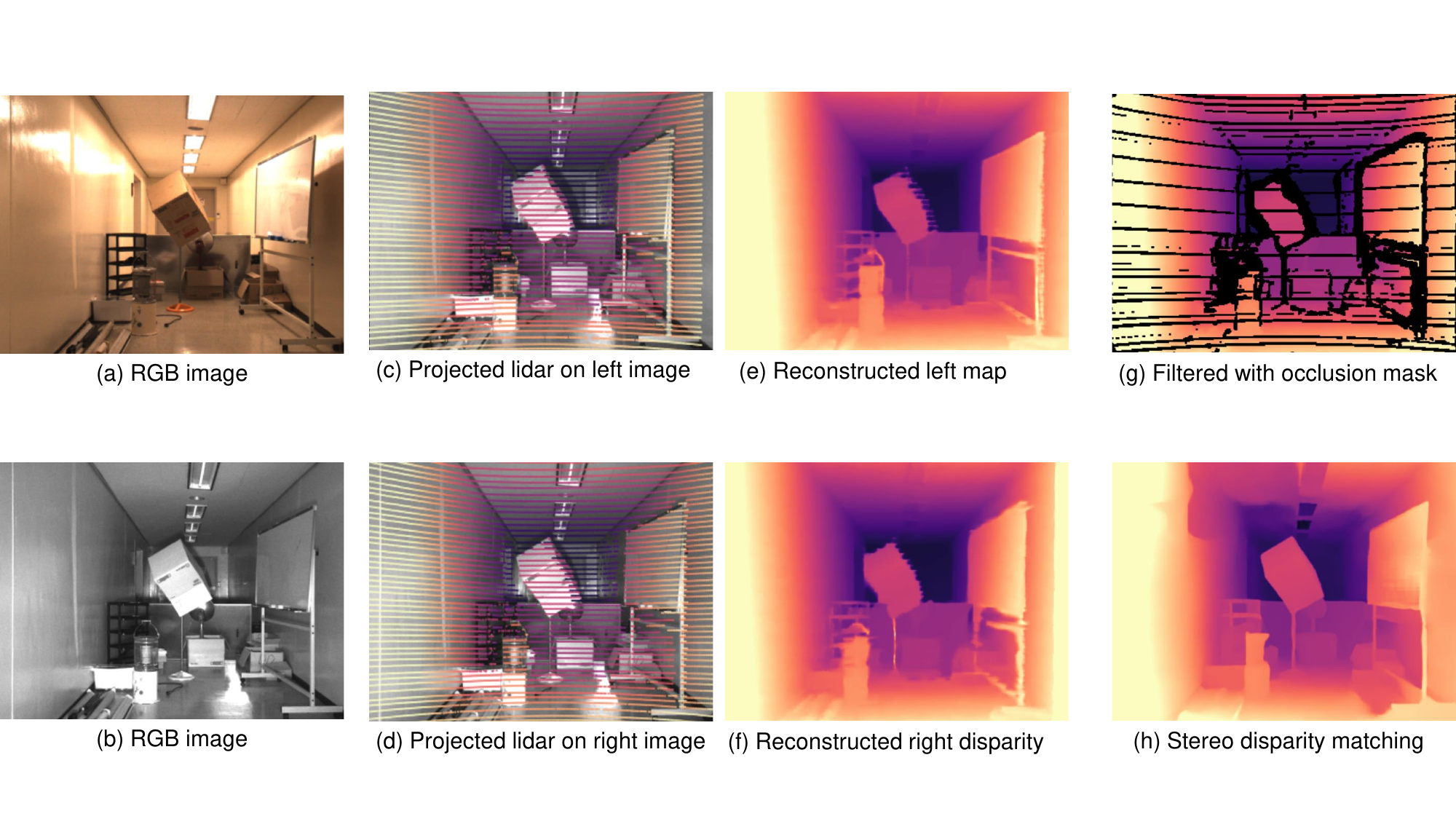}
  \caption{\textbf{Sparse Depth Reconstruction}. (a) RGB image and (b) NIR image are used as guidance for reconstruction. LiDAR points are projected onto the left image plane (c) and the right image plane (d). The points are then densified following ~\cite{Tang_2024_bpnet} in (e) and (f). Left and right consistency are employed to obtain the final disparity map with an occlusion mask (g). The resulting disparity map is compared with stereo disparity matching results (h) from ~\cite{lipson2021raft}, demonstrating an appropriate range of values.}
  \label{fig:bpnet}
  \end{figure*}
We reconstruct a dense depth map from sparse LiDAR data, adhering to the approach proposed in~\cite{Tang_2024_bpnet}. Figure~\ref{fig:bpnet} shows steps of reconstruction. The reconstruction process is delineated in the following steps:

\paragraph{Sparse Depth Map Generation}

We begin by converting LiDAR points, denoted as $ \mathbf{p}_{\text{LiDAR}} = (x_{\text{LiDAR}}, y_{\text{LiDAR}}, z_{\text{LiDAR}}) $, into a sparse depth map. This conversion is achieved through the projection of LiDAR points onto the image plane using the camera intrinsic matrix $ \mathbf{K}_{\text{camera}} $ and the extrinsic transformation matrix $ \mathbf{M}_{\text{LiDAR} \rightarrow \text{camera}} $, as illustrated in Equation~\eqref{eq:projection}:
\begin{equation}
[u_\text{image},v_\text{image},z_\text{image}]^\top = \mathbf{K}_{\text{camera}} \cdot \mathbf{M}_{\text{LiDAR} \rightarrow \text{camera}} \cdot [x_\text{LiDAR}, y_\text{LiDAR}, z_\text{LiDAR}]^\top.
\label{eq:projection}
\end{equation}
The matrices $ \mathbf{K}_{\text{camera}} $ and $ \mathbf{M}_{\text{LiDAR} \rightarrow \text{camera}} $ are precomputed through a calibration process.

\paragraph{Occlusion Handling and Disparity Filtering}

During the projection, occlusions may result in certain LiDAR points being erroneously visible from the camera's viewpoint. To address this, we implement a filtering mechanism based on disparity scaling. The depth obtained from LiDAR is converted to disparity using the relation:
\begin{equation}
d_{\text{sparse}} = \frac{f_x \cdot \text{baseline}}{z_{\text{image}}},
\end{equation}
where $ f_x $ is the focal length in the x-direction and the baseline is the distance between the stereo camera pair, both determined through calibration. The sparse disparity map is represented as a set of 3-channel vectors $ (u_\text{image}, v_\text{image}, d_\text{sparse}) $.

To adjust the density of LiDAR points and mitigate occlusion artifacts, we implement a refinement process on the sparse disparity points. The refinement algorithm is outlined in Algorithm~\ref{alg:refine_disparity_points} and is mathematically detailed below.

\begin{algorithm}[H]
\caption{Disparity Point Refinement}
\label{alg:refine_disparity_points}
\begin{algorithmic}[1]
\Procedure{RefineDisparityPoints}{Points, $\alpha$, $\beta$}
    \State \textbf{Input}: 
    \State \quad Points $\mathbf{P} = \{(u_i, v_i, d_i)\}_{i=1}^N$, 
    \State \quad Distance threshold $\alpha$, 
    \State \quad Disparity ratio threshold $\beta$
    \State \textbf{Output}: 
    \State \quad Filtered Points $\mathbf{P}_{\text{filtered}}$
    
    \For{each point $i$ in $\mathbf{P}$}
        \For{each point $j$ in $\mathbf{P}$, $j \neq i$}
            \State Calculate the Euclidean distance:
            \begin{equation}
            \text{dist}(i,j) = \sqrt{(u_i - u_j)^2 + (v_i - v_j)^2}
            \end{equation}
            \If{$\text{dist}(i,j) \leq \alpha \cdot d_i$ and $d_j < \beta \cdot d_i$}
                \State \textbf{Remove} point $i$ from $\mathbf{P}$
                \State \textbf{Break} inner loop
            \EndIf
        \EndFor
    \EndFor
    
    \State \textbf{return} $\mathbf{P}_{\text{filtered}}$
\EndProcedure
\end{algorithmic}
\end{algorithm}

The algorithm operates as follows:

\begin{enumerate}
    \item \textbf{Point Separation}: The input point set $\mathbf{P} = \{(u_i, v_i, d_i)\}_{i=1}^N$ consists of pixel coordinates $(u_i, v_i)$ and their corresponding disparity values $d_i$.
    
    \item \textbf{Distance Calculation}: For each point $i$, the Euclidean distance to every other point $j$ is computed:
    \begin{equation}
    \text{dist}(i,j) = \sqrt{(u_i - u_j)^2 + (v_i - v_j)^2}
    \end{equation}
    
    \item \textbf{Thresholding}: 
    \begin{itemize}
        \item $\alpha$: A point $j$ is considered a neighbor of point $i$ if $\text{dist}(i,j) \leq \alpha \cdot d_i$.
        \item $\beta$: Among the neighboring points, if the disparity $d_j$ of point $j$ is less than $\beta \cdot d_i$, then point $i$ is marked for removal.
    \end{itemize}
    
    \item \textbf{Point Removal}: If any neighboring point $j$ satisfies both the distance and disparity ratio conditions, point $i$ is removed from the point set $\mathbf{P}$ to obtain the filtered point set $\mathbf{P}_{\text{filtered}}$.
    
\end{enumerate}

In our experiments, we set $\alpha = 0.75$ and $\beta = 0.85$ to effectively control the density of the remaining points. This refinement process ensures that points excessively deep relative to their local neighborhood are eliminated, thereby reducing occlusion artifacts in the sparse disparity map.
The filtered disparity map is then converted back to a refined sparse depth map using the inverse of the disparity relation:
\begin{equation}
z_{\text{refined}} = \frac{f_x \cdot \text{baseline}}{d_{\text{sparse}}}.
\end{equation}

\paragraph{Dense Depth Reconstruction Using BPNet}

Subsequently, we employ BPNet~\cite{Tang_2024_bpnet}, a sparse-to-dense depth reconstruction model. BPNet takes as input the RGB image, camera intrinsic parameters, and the projected sparse depth map to generate a comprehensive dense depth map.

\paragraph{Stereo Camera Setup and Right Camera Depth Map}

Given the stereo camera configuration, we extend the transformation matrix $ \mathbf{M} $ by incorporating the left-right translation vector $ \mathbf{T} $, thereby obtaining $ \mathbf{M}_{\text{LiDAR} \rightarrow \text{right}} $. Using this augmented transformation matrix, we project the LiDAR points to the right camera's image plane and generate a dense depth map for the right view.

\paragraph{Disparity Map Computation and Occlusion Mask Generation}

With dense depth maps for both left and right cameras, we compute their respective disparity maps by applying the disparity-depth relation. 
To facilitate occlusion detection, the right disparity map \( d_{\text{right}} \) is projected onto the left image plane to obtain \( d_{\text{right} \rightarrow \text{left}} \). This projection accounts for the relative position of the stereo cameras and is mathematically represented as:

\begin{equation}
u_{\text{left}} = u_{\text{right}} + \frac{d_{\text{right}}(u_{\text{right}}, v_{\text{right}}) \cdot \text{baseline}}{f_x}
\end{equation}

Discrepancies exceeding a predefined threshold indicate occlusions, which are marked in an occlusion mask as follows:

\begin{equation}
\text{occlusion mask} =
\begin{cases}
1 & \text{if } |d_{\text{left}} - d_{\text{right} \rightarrow \text{left}}| > \text{threshold} \\
0 & \text{otherwise}
\end{cases}
\end{equation}
\paragraph{Final Dense Disparity Map and Application}

Finally, we utilize the occlusion mask to retain only the valid regions in the left dense disparity map, effectively filtering out occluded areas. The resulting refined dense disparity map, in conjunction with the occlusion mask, serves as a robust label for training and testing in stereo depth estimation tasks.

\subsubsection{Samples of Real Dataset}
\begin{figure*}[h]
  \centering
  \begin{subfigure}{\textwidth}
    \centering
    \includegraphics[width=\linewidth]{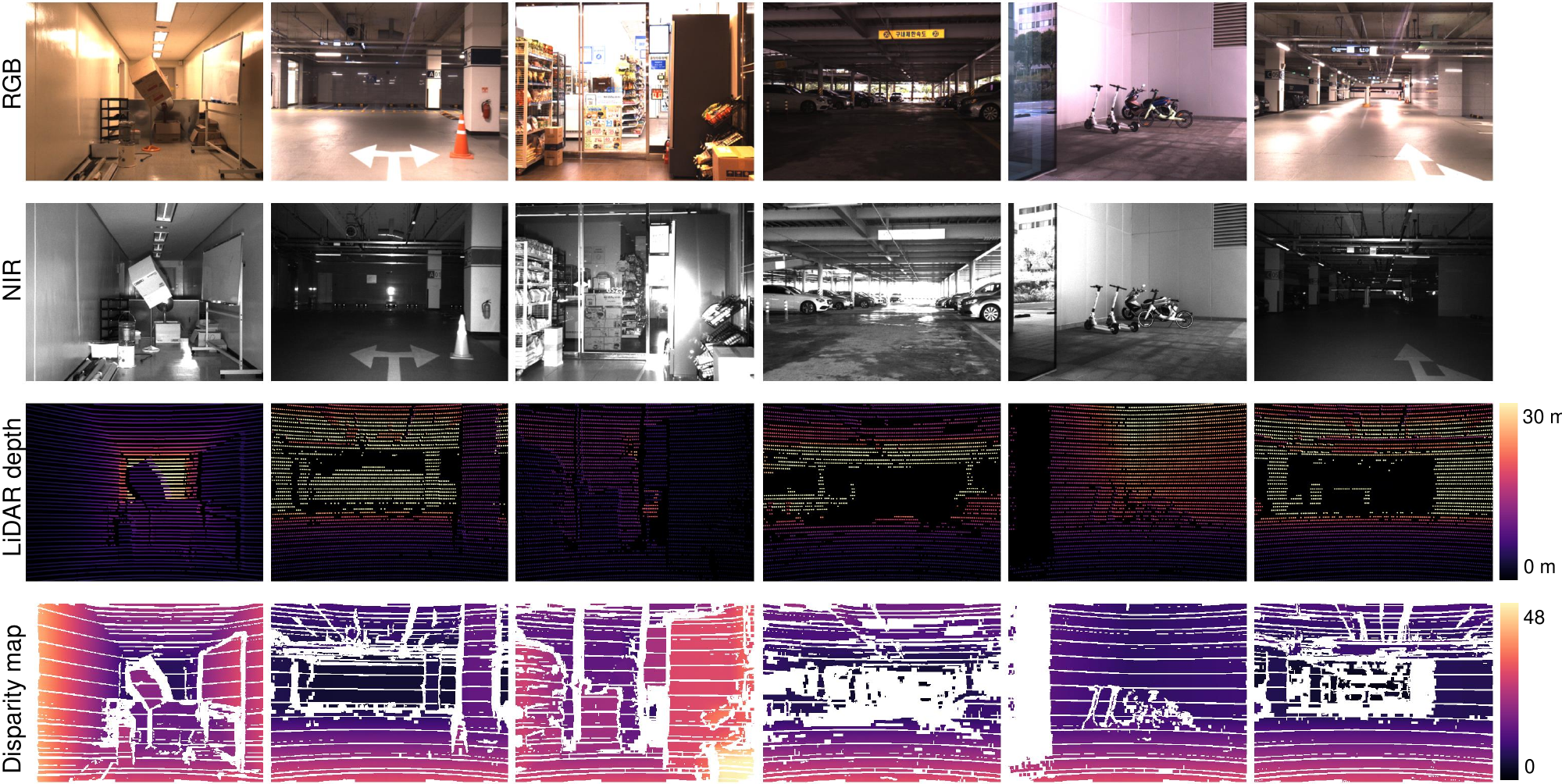}
    \caption{\textbf{Real data of indoor environment}.}
  \end{subfigure}
  \hfill
  \begin{subfigure}{\textwidth}
    \centering
    \includegraphics[width=\linewidth]{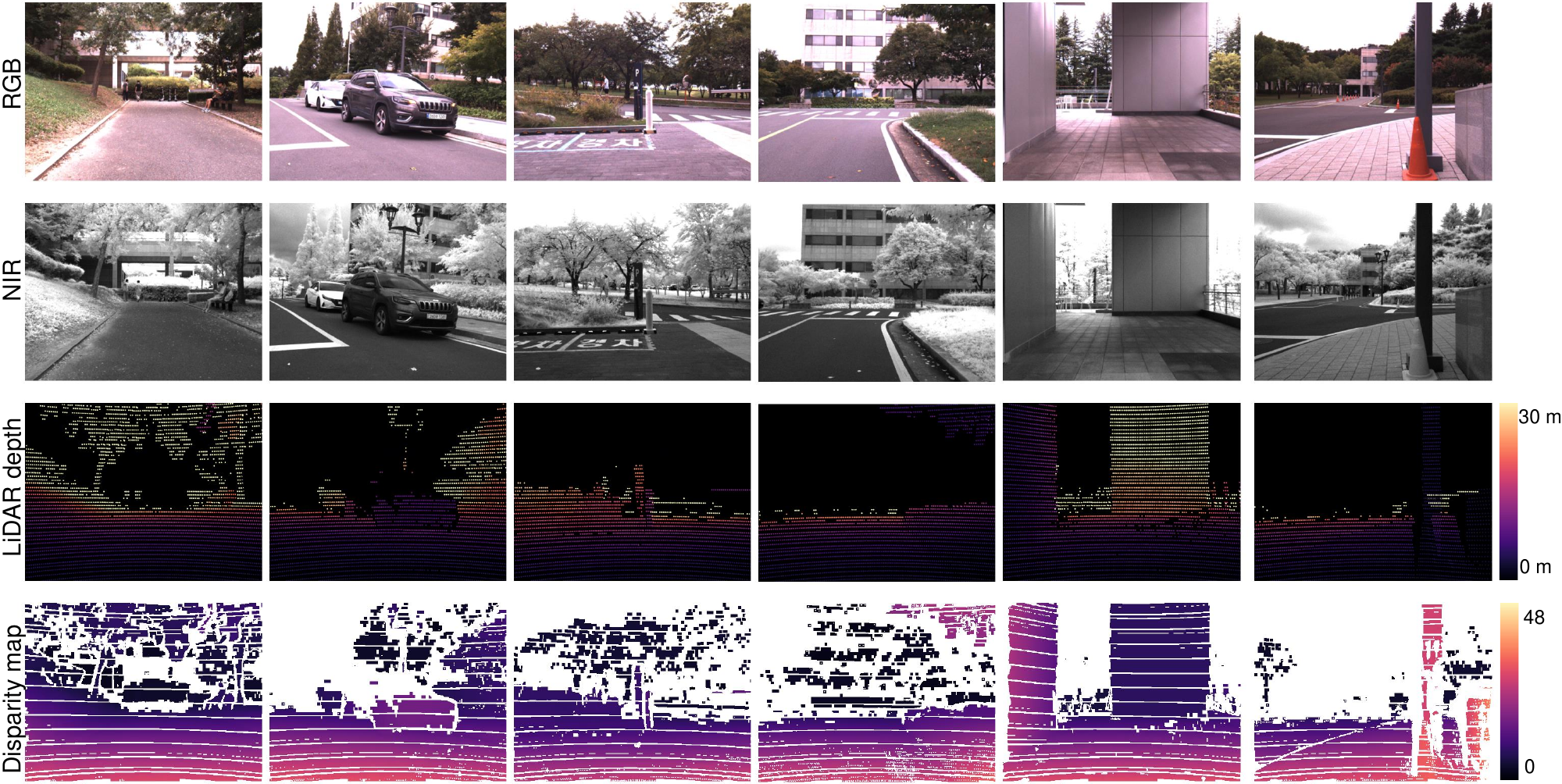}
    \caption{\textbf{Real data of outdoor day}.}
  \end{subfigure}
    \caption{Various real dataset samples}
  \label{fig:all_real_data}
  \end{figure*}

\begin{figure*}[h]
  \begin{subfigure}{\textwidth}
    \centering
    \includegraphics[width=\linewidth]{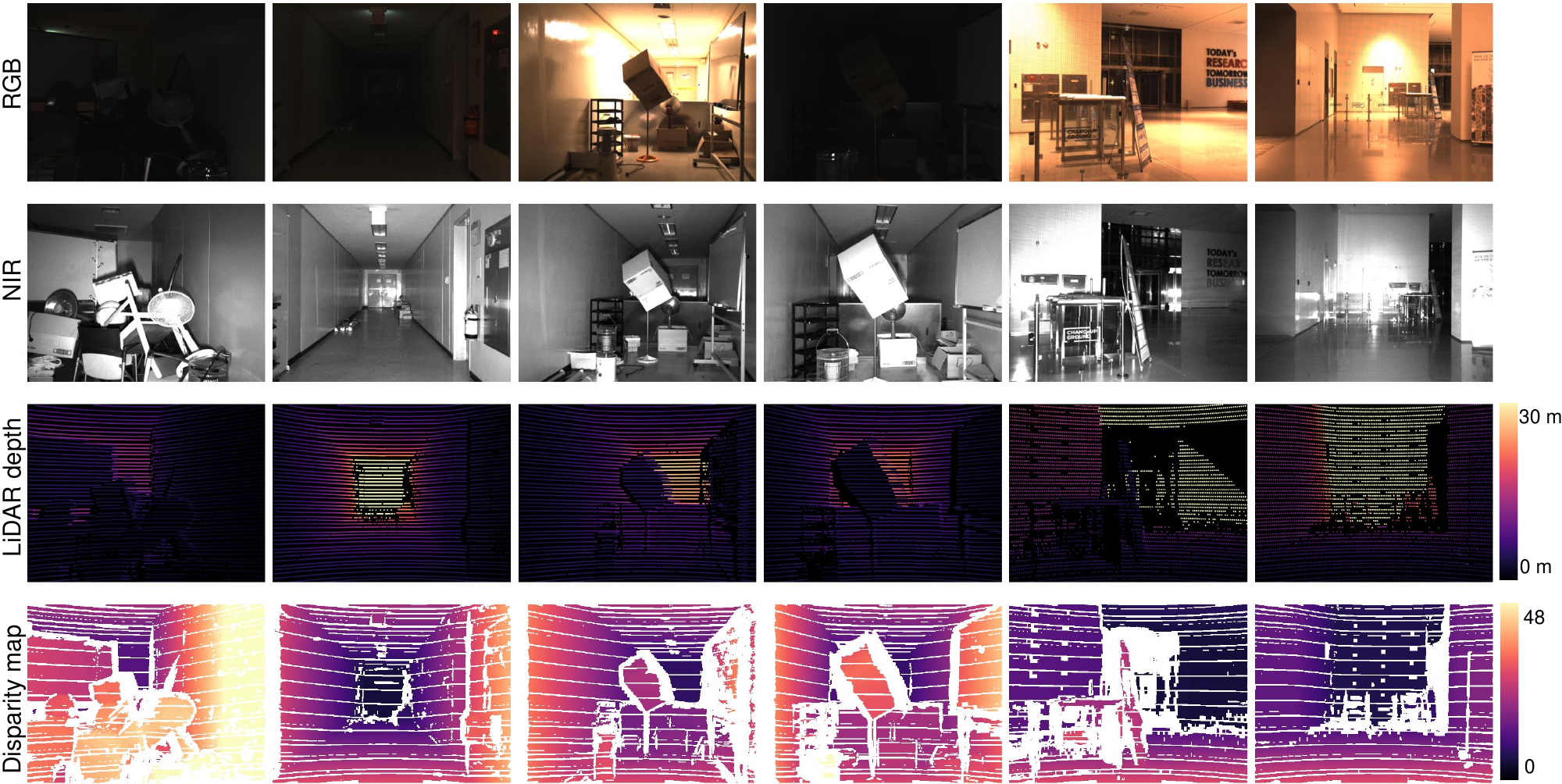}
    \caption{\textbf{Real data of indoor environment with challenging lighting}.}
    \label{fig:real_data_3}
  \end{subfigure}
  \hfill
  \begin{subfigure}{\textwidth}
    \centering
    \includegraphics[width=\linewidth]{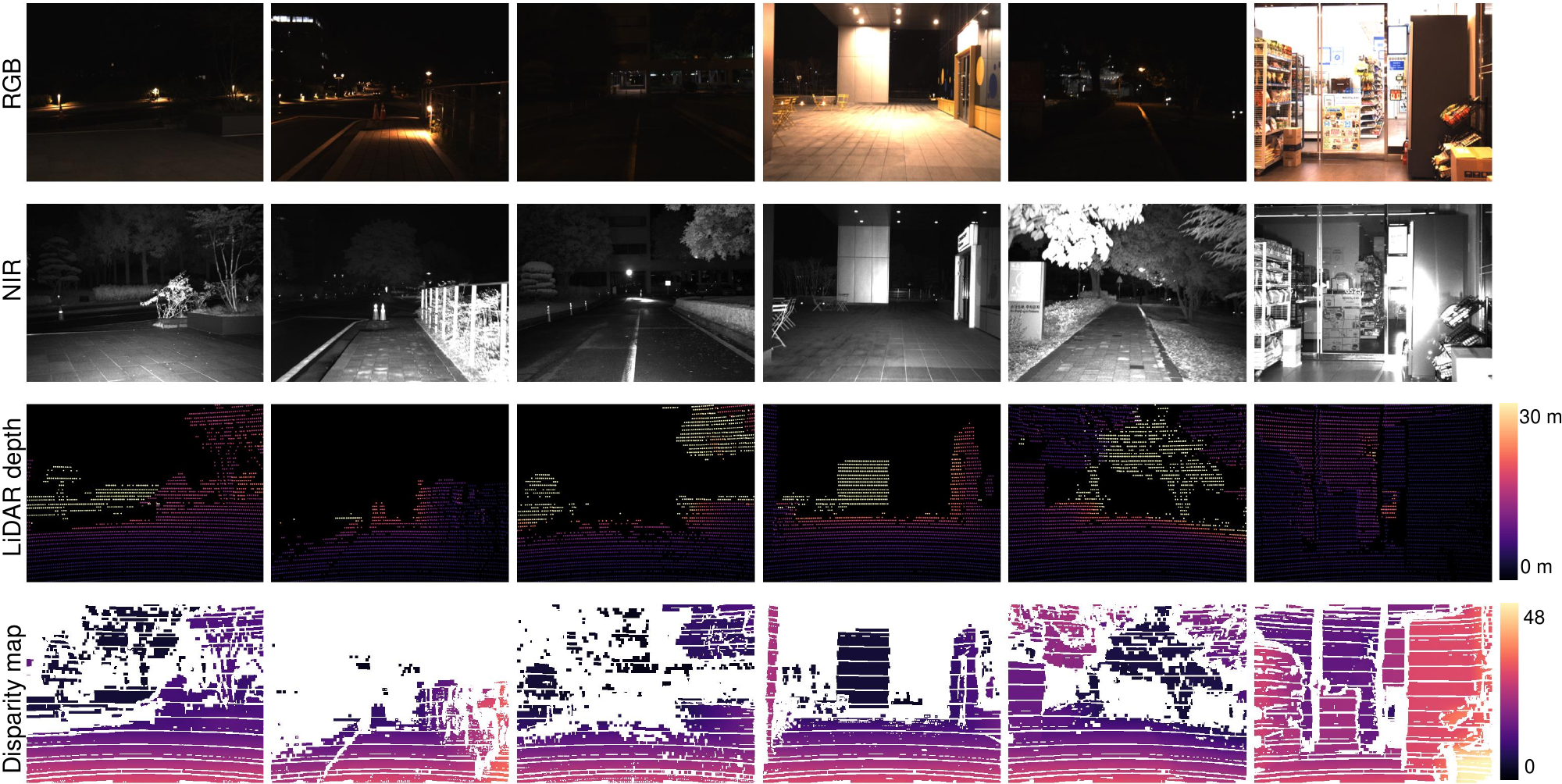}
    \caption{\textbf{Real data of outdoor night}.}
    \label{fig:real_data_4}
  \end{subfigure}
  
  \caption{Various real dataset samples on challenging lighting condition.}
  \label{fig:all_real_data_challenging}
\end{figure*}

Figure~\ref{fig:all_real_data} presents samples from our acquired real dataset, which includes RGB stereo images, NIR stereo images, and sparse LiDAR points. Additionally, we performed sparse LiDAR reconstruction as described in Section~\ref{sec:bpnet}, with the results illustrated in the figure. Figure~\ref{fig:all_real_data_challenging} showcases our dataset under challenging lighting conditions, such as nighttime and poorly lit indoor environments. We collected over 80,000 frames across diverse settings, including indoor and outdoor environments, bright and dark locations, as well as roads and sidewalks. By capturing data across a wide range of real-world environments and times of day, our dataset is expected to support not only depth estimation but also 3D geometry reconstruction techniques, such as structure-from-motion~\cite{schonberger2016sfm_colmap}, and various photometric volume reconstruction methods, including Gaussian splatting~\cite{3dgaussian}.

\subsubsection{Statistics}
\begin{figure}[h]
    \centering
    \includegraphics[width=\linewidth]{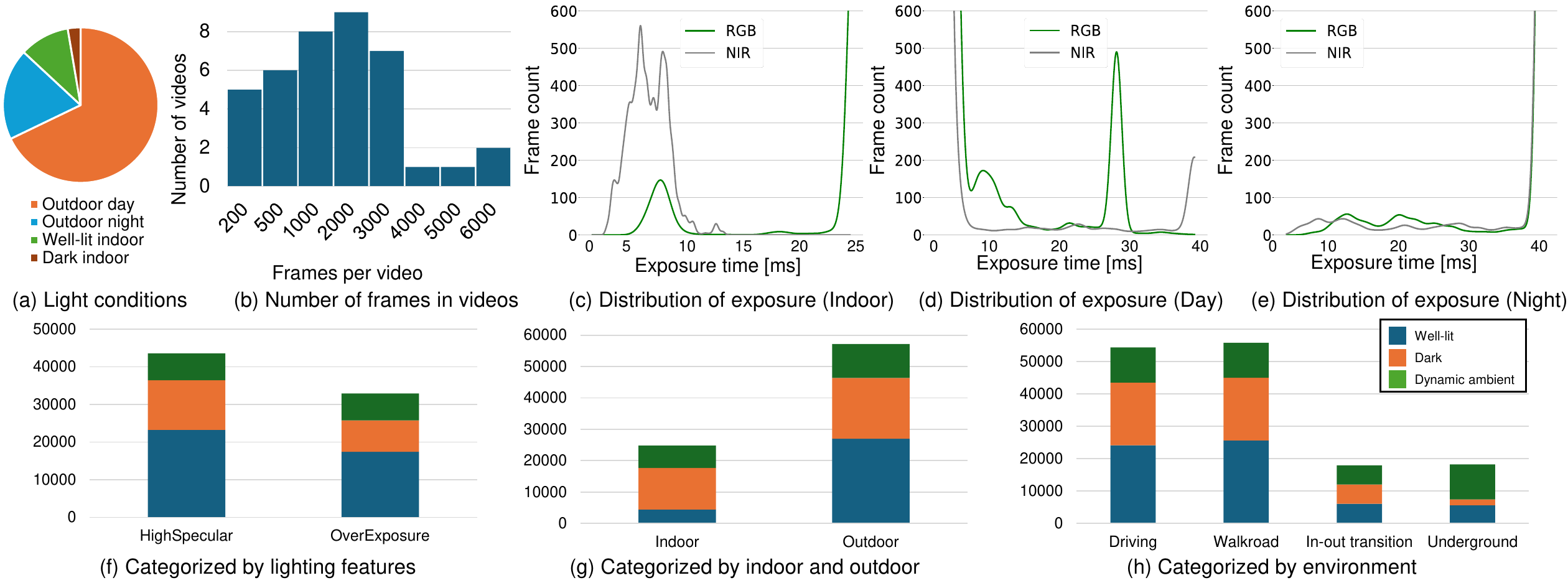}
    \caption{\textbf{Statistics of our real dataset}.}
    \label{fig:dataset_statistics}
\end{figure}

Figure~\ref{fig:dataset_statistics} presents a comprehensive analysis of our dataset.
(a) illustrates the proportion of the dataset categorized into four distinct lighting conditions: Outdoor Day, Outdoor Night, Well-Lit Indoor, and Dark Indoor.
(b) provides a histogram that depicts the distribution of video scenes based on the number of frames. 
(c), (d), and (e) show histograms of the exposure times for RGB and NIR cameras under Outdoor Day, Outdoor Night, and Indoor settings, respectively. Notably, due to the independent operation of auto-exposure for the RGB and NIR sensors, the exposure time distributions differ between the two modalities.
Graph (f), (g), and (h) further classify scenes based on additional criteria. 
(f) evaluates the presence of high-specular scenes and overexposure conditions, while (g) categorizes scenes as either indoor or outdoor. 
(h) details the diverse environments encountered, including road driving, pedestrian pathways, transitions between indoor and outdoor settings, and underground roads. 
These scenes are further subdivided into Well-Lit, Dark, and Dynamic Lighting conditions.
Importantly, the bar graphs in (f), (g), and (h) illustrate that a single scene can belong to multiple categories simultaneously, highlighting the complexity and diversity of our dataset.

\subsection{Synthetic Dataset}

\subsubsection{Synthetic Dataset Augmentation}

\paragraph{Augmentation from Image Formation Equation}
The image formation of our RGB-NIR pixel aligned camera is denotes as following:
\begin{align}
\label{eq:image_formation}
    I_{i}^{c}(p^c) = \eta_1 + g_i \left( \eta_2 + t_i \left( R_i^c(p^c) \left( E_i^c(p^c) + L_i^c(p^c) \right) \right) \right).
\end{align}
To facilitate training of other vision tasks such as stereo depth estimation, it is essential to have a training dataset with precise ground-truth information, which can be achieved by creating a synthetic dataset. 
However, generating a large-scale synthetic dataset from scratch—including rendering 3D scenes and extending the three-channel color space to four channels (R, G, B, NIR)—requires substantial resources that may not be readily available. 
Therefore, we utilized existing large-scale RGB stereo datasets~\cite{mayer2016sceneflow,scharstein2014middlebury} and developed a synthetic rendering pipeline leveraging an image formation model to generate realistic RGB images based on provided depth maps and material segmentation maps. 

\begin{figure*}[h]
  \centering
  \includegraphics[width=\textwidth]{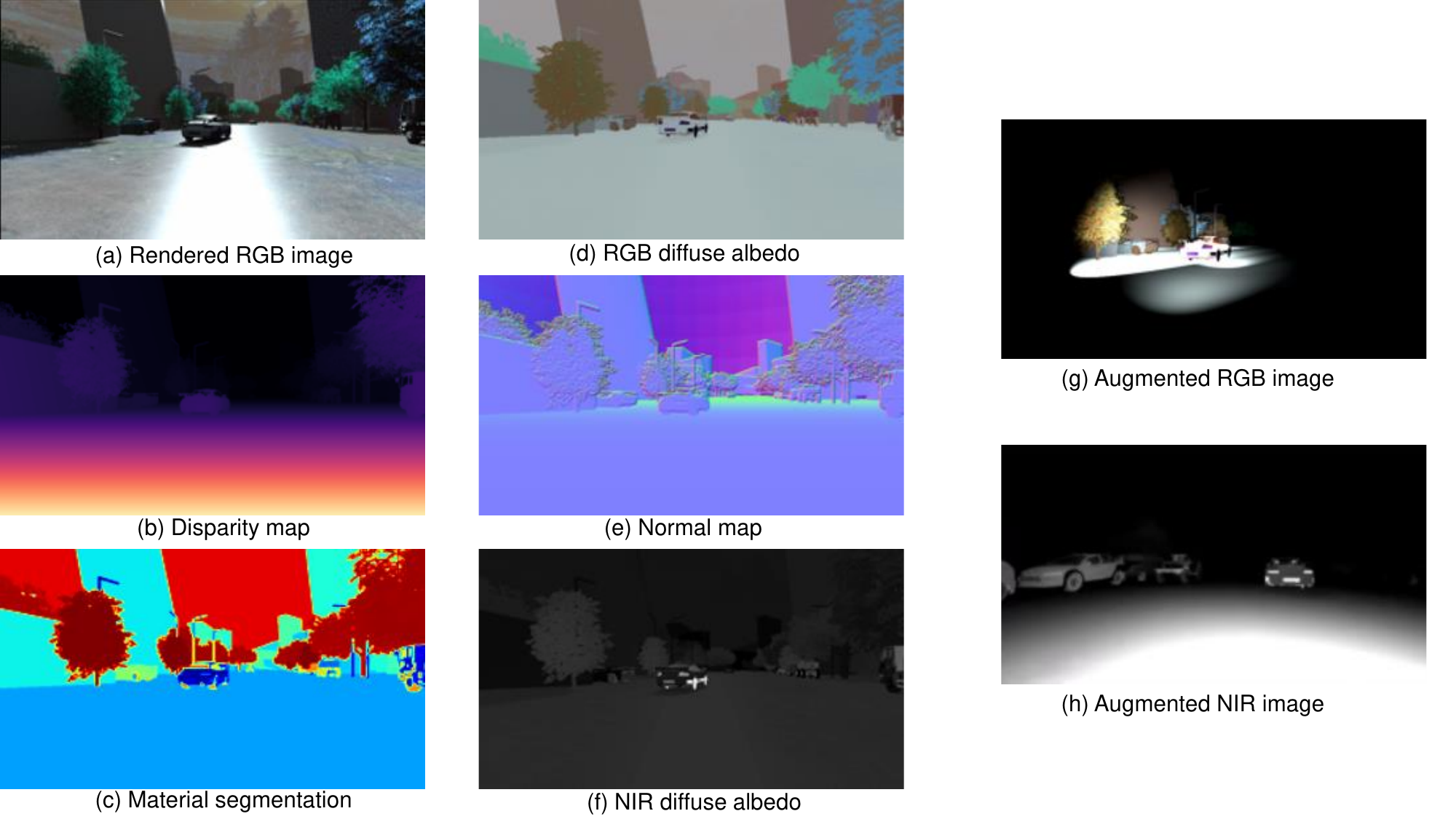}
  \caption{\textbf{RGB-NIR synthetic data augmentation}.  Sceneflow dataset~\cite{mayer2016sceneflow} provide (a) rendered RGB image, (b) disparity map and (c) material index. We assumed RGB albedo (d) with (a) \& (b), normal map (e) from (b). We simulated NIR albedo (f) from (d) by~\cite{Gruber_2019_gated2depth}. We rendered some light source for RGB (g) and NIR (h).
  } 
  \label{fig:synthetic}
\end{figure*}

Figure~\ref{fig:synthetic} shows image components in augmentation pipeline. The baseline dataset provides RGB images (a), depth (b) and material index (c). 
The albedo $R_{i\in\text{\{R,G,B\}}}^c(p^c)$ (Fig.~\ref{fig:synthetic}(d)) is accurately simulated by assigning distinct reflectance values to each material class identified in the segmentation map, thereby ensuring material-specific color representation. NIR albedo $R_{i\in\text{\{NIR\}}}^c(p^c)$ (Fig.~\ref{fig:synthetic}(f)) is pseudo driven from $R_{i\in\text{\{R,G,B\}}}^c(p^c)$ by~\cite{Gruber_2019_gated2depth}.

\paragraph{Normal Map Reconstruction} To accurately compute the lighting interactions, normal maps (Fig.~\ref{fig:synthetic}(e)) are derived from the depth maps (Fig.~\ref{fig:synthetic}(b)), enabling precise calculation of the incident angles between the light sources and the surface normals for each pixel. 
The conversion of a depth map into a normal map involves calculating the gradients in the $x$ and $y$ directions to determine the surface normals. 
This process starts by computing the partial derivatives of the depth values, which are then used to construct the normal vector at each point. 
Specifically, the normal vector $\mathbf{N}$ can be derived as 
\begin{equation}
\label{eqa:depth2normal}
    \mathbf{N} = \left(-\frac{\partial z}{\partial x}, -\frac{\partial z}{\partial y}, z\right),
\end{equation}
where $x,y,z$ consist of point cloud on camera coordinate, driven by projecting depth map into camera coordinate, followed by normalization to ensure unit length. Detail implementation follows~\cite{hoppe1992surface_recon_from_points}.

\paragraph{Ambient Lighting} Ambient lighting $ E_i^c(p) $ is introduced to emulate diverse environmental illumination conditions by utilizing multiple ambient light sources. Each ambient light source is characterized by its unique position and brightness, contributing cumulatively to the overall ambient illumination at each pixel. Specifically, the ambient lighting is modeled as the sum of contributions from $ n $ ambient light sources, as described by the following equation:

\begin{equation}
E_i^c(p) = \sum_{j=1}^{n} \phi_j \cdot \max\left(0, \mathbf{N}(p) \cdot \mathbf{L}_j(p)\right),
\label{eq:ambient_lighting}
\end{equation}
where
\begin{itemize}
    \item $ \phi_j $ denotes the brightness of the $ j $-th ambient light source.
    \item $ \mathbf{N}(p) $ represents the normal vector at pixel $ p $, derived by~\ref{eqa:depth2normal}.
    \item $ \mathbf{L}_j(p) $ is the unit vector pointing from the surface point $ p $ to the position of the $ j $-th ambient light source.
\end{itemize}

This formulation allows for realistic simulation of ambient lighting by accounting for the direction and intensity of multiple light sources, thereby enhancing the visual fidelity of the rendered images.

\paragraph{Active Illumination}

Active illumination $ L_i^c(p) $ is incorporated to provide consistent direct illumination from a single, fixed light source. 
Unlike ambient lighting, active illumination affects only the Near-Infrared (NIR) channel, leaving the Red (R), Green (G), and Blue (B) channels unaffected. 
This selective illumination is particularly useful for applications requiring multi-spectral data. 
The active lighting is modeled without summation, as only one active light source is present, and is defined as follows:

\begin{equation}
L_i^c(p) =
\begin{cases}
0, & \text{for } i \in \{R, G, B\} \\
\phi_{\text{active}} \cdot \max\left(0, \mathbf{N}(p) \cdot \mathbf{L}_{\text{active}}(p)\right), & \text{for } i = \text{NIR}
\end{cases}
\label{eq:active_illumination}
\end{equation}

where:
\begin{itemize}
    \item $ \phi_{\text{active}} $ denotes the brightness of the active light source.
    \item $ \mathbf{L}_{\text{active}}(p) $ is the unit vector pointing from the surface point $ p $ to the fixed position of the active light source.
    \item The term $ \max\left(0, \mathbf{N}(p) \cdot \mathbf{L}_{\text{active}}(p)\right) $ ensures that only positive contributions to the illumination are considered, adhering to the Lambertian reflectance model.
\end{itemize}

By restricting active illumination to the NIR channel and maintaining a fixed light position and intensity, the model ensures that direct illumination is consistently applied without altering the RGB channels. 

The image formation equation incorporates fixed exposure time $t_i$ and gain $g_i$ parameters, which are held constant throughout the simulation to maintain uniform exposure settings. 
Gaussian noise is systematically added both pre- and post-processing to emulate realistic sensor noise, thereby enhancing the fidelity of the synthetic images. 
This comprehensive approach integrates material properties, complex lighting interactions, and realistic noise modeling, resulting in high-quality synthetic renderings that are suitable for various applications in computer vision and graphics research.

\subsubsection{Samples of Synthetic Dataset}

\begin{figure*}[h]
  \centering
  \includegraphics[width=\textwidth]{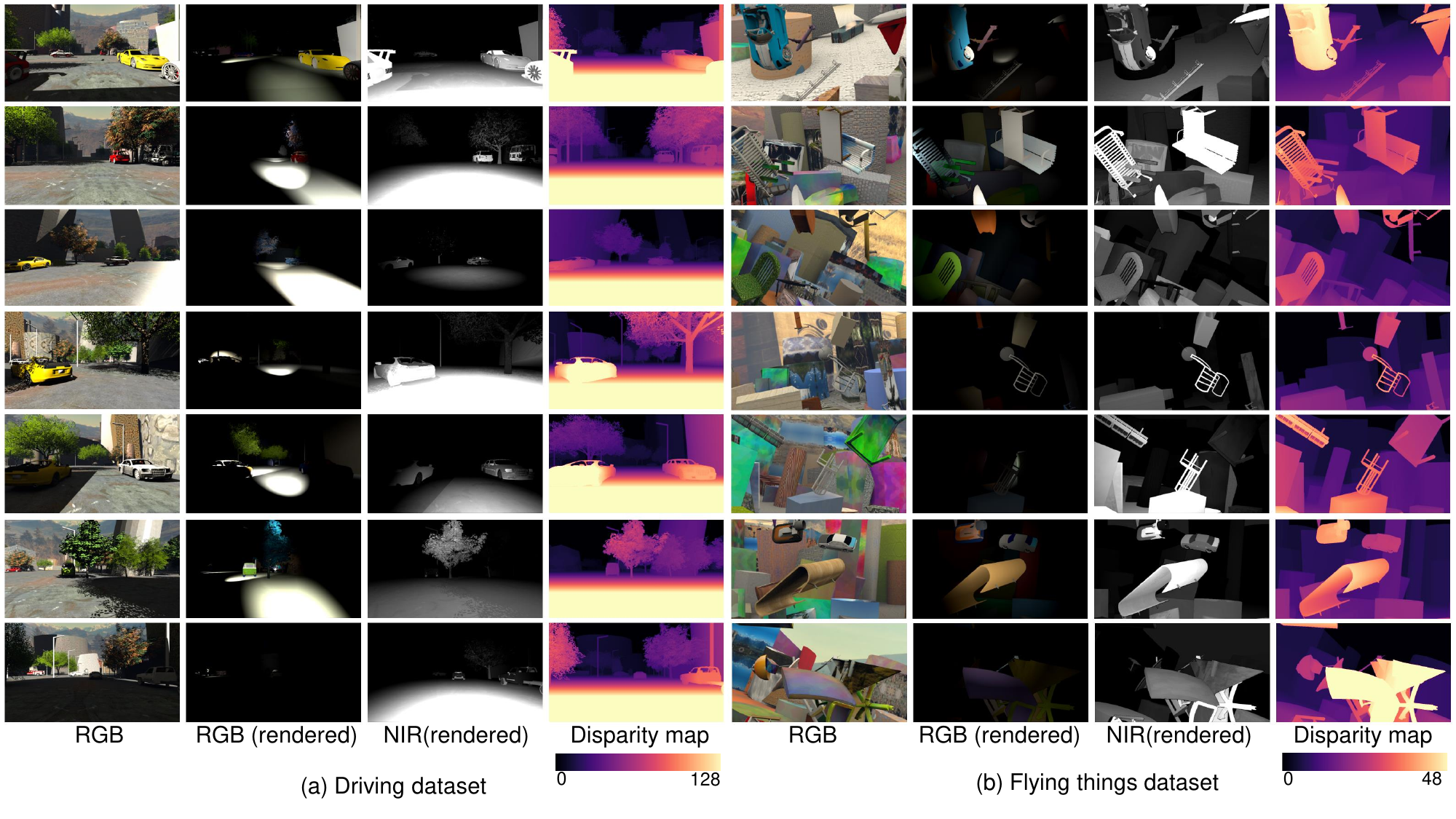}
  \caption{\textbf{Samples of augmented synthetic dataset}. Sceneflow~\cite{mayer2016sceneflow} includes both ourdoor environment (a) and indoor environment (b) dataset. We augmented these datasets relighting RGB stereo images and NIR stereo images.
  } 
  \label{fig:synthetic_samples}
\end{figure*}

Figure~\ref{fig:synthetic_samples} presents samples from our augmented synthetic dataset. We enhanced two distinct environments from the Sceneflow dataset~\cite{mayer2016sceneflow}: a driving scene and an indoor environment featuring randomly flying objects. 
The augmented dataset includes both RGB stereo and Near-Infrared (NIR) stereo images, utilizing existing components of the original dataset such as RGB stereo pairs and disparity maps. 
This augmented dataset was employed to train a stereo depth estimation network using the original disparity map labels and an image fusion model with the original RGB images.

\subsection{Comparison to Other Datasets}

\begin{table*}[htbp]
    \scriptsize
    \newcommand{\GCL}{\cellcolor[HTML]{C6EFCE}}
    \newcommand{\GCR}{\cellcolor[HTML]{FFC7CE}}
    \newcommand{\GCY}{\cellcolor[HTML]{FFF2CC}}
    
\centering
\renewcommand{\tabcolsep}{3pt}
\begin{tabularx}{\textwidth}{|c|c|c|c|c|c|c|c|c|c|c|c|c|l|l|}
    \hline
    \textbf{Dataset} & \makecell{\textbf{Pixel} \\ \textbf{-aligned}} & \makecell{\textbf{Multi} \\ \textbf{-View}} & \makecell{\textbf{RGB} \\ \textbf{Stereo}} & \makecell{\textbf{NIR} \\ \textbf{Stereo}} & \makecell{\textbf{GT} \\ \textbf{Depth}} & \makecell{\textbf{Lidar} \\ \textbf{Depth}} & \makecell{\textbf{Video}} & \makecell{\textbf{Spectral} \\ \textbf{Bands}} & \textbf{Indoor} & \textbf{Outdoor} & \textbf{Day} & \textbf{Night} & \textbf{Base Platform} & \makecell{\textbf{Pixel-align} \\ \textbf{Implementation}} \\ \hline
    ~\cite{Geiger_KITTY_2012}& \GCR X & \GCL O & \GCL O & \GCR X & \GCL O & \GCL O & \GCL O & RGB & \GCR X & \GCL O & \GCL O & \GCL O & Vehicle & N/A \\ \hline
    ~\cite{couprie2013nyu_dataset} & \GCR X & \GCL O & \GCR X & \GCR X & \GCL O & \GCR X & \GCR O & RGB & \GCL O & \GCR X & \GCR O & \GCR X & Hand-carried & N/A \\ \hline
    
    ~\cite{Cordts_Omran_2016_Cityscapes} & \GCR X & \GCL O & \GCL O & \GCR X & \GCL O & \GCL O & \GCL O & RGB & \GCR X & \GCL O & \GCL O & \GCL O & Vehicle & N/A \\ \hline
    
    ~\cite{toet2016dataset_nir_fusion_aligned} & \GCL O & \GCR X & \GCR X & \GCR X & \GCR X & \GCR X & \GCR X & RGB, NIR, Thermal & \GCR X & \GCL O & \GCL O & \GCL O & N/A & Beam splitter \\ \hline
    ~\cite{liu2016multispectral_pedestrian} & Thermal & \GCL O & \GCL O & \GCR X & \GCL O & \GCL O & \GCL O & RGB, Thermal & \GCR X & \GCL O & \GCL O & \GCL O & Vehicle & Beam splitter \\ \hline
    
    ~\cite{dai2017scannet} & \GCR X & \GCL O & \GCR X & \GCR X & \GCL O & \GCR X & \GCR O & RGB & \GCL O & \GCR X & \GCR O & \GCR X & Hand-carried & N/A \\ \hline
    
    ~\cite{chebrolu2017agricultural} & \GCL O & \GCL O & \GCR X & \GCR X & \GCL O & \GCL O & \GCL O & RGB, NIR & \GCR X & \GCL O & \GCL O & \GCR X & Tractor & Prism camera \\ \hline
    ~\cite{takumi2017multispectral_detection} & \GCR X & \GCL O & \GCR X & \GCR X & \GCR X & \GCR X & \GCR O & RGB, NIR, FIR, MIR & \GCR X & \GCL O & \GCL O & \GCL O & Hand-carried & N/A \\ \hline
    ~\cite{valada2017rgb_nir_seg_robot} & \GCR X & \GCL O & \GCL O & \GCR X & \GCL O & \GCR X & \GCR O & RGB, NIR & \GCR X & \GCL O & \GCL O & \GCR X & Mobile robot & N/A \\ \hline
    ~\cite{Zhi_2018_material_cross_spectral} & \GCR X & \GCL O & \GCR X & \GCR X & \GCR X & \GCR X & \GCL O & RGB, NIR & \GCR X & \GCL O & \GCL O & \GCL O & Vehicle & N/A \\ \hline
    ~\cite{choe2018ranus} & \GCR X & \GCL O & \GCR X & \GCR X & \GCR X & \GCR X & \GCL O & RGB, NIR & \GCR X & \GCL O & \GCL O & \GCL O & Vehicle & N/A \\ \hline
    ~\cite{Choi_Kim_Hwang_Park_Yoon_An_Kweon_2018} & Thermal & \GCL O & \GCL O & \GCR X & \GCL O & \GCL O & \GCL O & RGB, Thermal & \GCR X & \GCL O & \GCL O & \GCL O & Vehicle & Beam splitter \\ \hline
    ~\cite{Lee_Cho_Shin_Kim_Myung_2022_Vivid} & \GCR X & \GCL O & \GCL O & \GCR X & \GCL O & \GCL O & \GCL O & RGB, Thermal & \GCL O & \GCL O & \GCL O & \GCL O & Vehicle & N/A \\ \hline
    ~\cite{Li_Cai_Saputra_Dai_Lu_2022} & \GCR X & \GCL O & \GCL O & \GCR X & \GCL O & \GCL O & \GCL O & RGB, LWIR & \GCL O & \GCL O & \GCL O & \GCL O & Mobile robot, drone & N/A \\ \hline
    ~\cite{Guo_2023_cross_stereo_est_and_data} & Thermal & \GCL O & \GCL O & Thermal & \GCL O & \GCL O & \GCL O & RGB, Thermal & \GCR X & \GCL O & \GCL O & \GCL O & Vehicle & Beam splitter \\ \hline
    ~\cite{Shin_Park_Kweon_2023} & \GCR X & \GCL O & \GCL O & \GCL O & \GCL O & \GCL O & \GCL O & RGB, NIR, Thermal & \GCR X & \GCL O & \GCL O & \GCL O & Vehicle & N/A \\ \hline
    ~\cite{mortimer_2024_goose} & \GCL O & \GCL O & \GCL O & \GCR X & \GCL O & \GCL O & \GCL O & RGB, NIR & \GCR X & \GCL O & \GCL O & \GCL O & Vehicle & Prism camera \\ \hline

    \textbf{Ours} & \GCL O & \GCL O & \GCL O & \GCL O & \GCL O & \GCL O & \GCL O & RGB, NIR & \GCL O & \GCL O & \GCL O & \GCL O & Mobile robot & Prism camera \\ \hline
\end{tabularx}
\caption{Summary of RGB-NIR and Multispectral Datasets Characteristics}
\label{tab:rgb_nir_datasets}
\end{table*}

We compared our dataset with existing large-scale datasets, including outdoor RGB stereo depth datasets~\cite{Geiger_KITTY_2012,Cordts_Omran_2016_Cityscapes}, indoor RGB stereo depth datasets~\cite{couprie2013nyu_dataset,dai2017scannet}, datasets with RGB and NIR~\cite{chebrolu2017agricultural,toet2016dataset_nir_fusion_aligned,Zhi_2018_material_cross_spectral,choe2018ranus,mortimer_2024_goose,Shin_Park_Kweon_2023,takumi2017multispectral_detection,valada2017rgb_nir_seg_robot}, and RGB-Thermal datasets~\cite{Choi_Kim_Hwang_Park_Yoon_An_Kweon_2018,Lee_Cho_Shin_Kim_Myung_2022_Vivid,Guo_2023_cross_stereo_est_and_data,liu2016multispectral_pedestrian}. The comparison is presented in Table~\ref{tab:rgb_nir_datasets}, based on several key criteria which are detailed below.

\paragraph{Pixel-aligned RGB-NIR}
Multispectral datasets primarily include different spectral image information alongside RGB. 
When two spectral images are pixel-aligned, they can be effectively fused without the need for pose correction, which greatly enhances efficiency. 
This pixel-level alignment is crucial for applications that leverage multispectral fusion.

\paragraph{Multi-View Camera}
Datasets that include a multi-view camera setup can utilize multi-view geometry, enabling tasks such as depth estimation or optical flow for 3D downstream vision tasks. 
In this comparison, if a dataset includes more than two cameras, it is considered multi-view. RGB-Stereo and NIR-Stereo specifically denote cases where multiple RGB or NIR cameras are included. 
For instance, the dataset from~\cite{Shin_Park_Kweon_2023} includes both RGB stereo and NIR stereo, employing an active stereo camera. On the other hand, datasets such as~\cite{choe2018ranus,Zhi_2018_material_cross_spectral,liu2016multispectral_pedestrian,mortimer_2024_goose,valada2017rgb_nir_seg_robot,takumi2017multispectral_detection} used only one multispectral camera, limiting their 3D geometry capabilities. 
Additionally,~\cite{dai2017scannet,couprie2013nyu_dataset} employed NIR structured light cameras but did not provide the original NIR stereo images, only the estimated depth, which limits further multispectral analysis.

\paragraph{Ground Truth Depth}
Ground truth depth is essential for evaluating depth estimation methods. Some datasets, such as~\cite{dai2017scannet,couprie2013nyu_dataset,valada2017rgb_nir_seg_robot}, used RGB-Depth or structured light cameras to obtain depth measurements. 
However, they do not include more accurate 3D depth measurements from LiDAR, which is a significant limitation for precise ground truth depth generation.

\paragraph{Environment}
The environmental conditions under which data is collected significantly impact dataset usability. 
Therefore, we categorize the datasets based on whether they include data from indoor, outdoor, day, or night scenarios.
Multispectral datasets are especially useful in scenarios such as low-light conditions or environments with varying lighting (indoor to outdoor), as they provide complementary information to enhance RGB data.

\paragraph{Base Platform}
For large-scale data collection, the imaging system requires an appropriate mobile platform. 
Smaller systems can be handheld~\cite{dai2017scannet,takumi2017multispectral_detection,couprie2013nyu_dataset}, while larger, heavier systems are typically mounted on vehicles~\cite{Shin_Park_Kweon_2023,Guo_2023_cross_stereo_est_and_data,Geiger_KITTY_2012}, limiting data collection to areas accessible by vehicle. 
Research using mobile robots or drones~\cite{valada2017rgb_nir_seg_robot,Li_Cai_Saputra_Dai_Lu_2022} improves both mobility and stability, allowing data collection in a wider range of environments.

\paragraph{Pixel-align Implementation}
Implementations of pixel-aligned multispectral imaging can be broadly categorized into beam splitters and prisms. Beam splitters are used by~\cite{toet2016dataset_nir_fusion_aligned,liu2016multispectral_pedestrian,Choi_Kim_Hwang_Park_Yoon_An_Kweon_2018,Guo_2023_cross_stereo_est_and_data}, splitting incoming light into two beams that are then directed to different cameras, achieving pixel-level alignment. 
Alternatively, dichroic prism-based cameras, such as those used by~\cite{chebrolu2017agricultural,mortimer_2024_goose} and our dataset, separate the spectral bands through a prism and direct them to different CMOS sensors. 
This approach offers a more compact design compared to beam splitters, and provides complete spectral separation, making it preferable for many applications.

\section{Details on RGB-NIR Feature Fusion and Attentional Fusion}
\subsection{RGB-NIR Image Fusion}
\subsubsection{Network Architecture}

\paragraph{Residual Block}
We implemented the ResNet~\cite{he2016deep} architecture and its pretrained weights~\cite{lipson2021raft} as feature extractors for both the image fusion method and the feature fusion depth method. Table~\ref{tab:modules_resnet} presents the details of its PyTorch implementation. The basic version of this encoder accepts 3-channel inputs and outputs downsampled feature maps with 256 channels. Additionally, we can adjust the number of input channels, output channels, and the downstream scale factor as needed to accommodate various requirements.

\begin{table}[h]
\centering

\begin{tabular}{|c|c|c|p{6cm}|c|}
\hline
\footnotesize
\textbf{Layer} & \textbf{Layer Name} & \textbf{Input Layer} & \textbf{Description} & \textbf{Output Shape} \\ \hline
1 & Input & - & Input data & $3 \times H \times W$ \\ \hline
2 & Conv1 & Layer 1 & $3 \times 3$ kernel, padding=1, stride=$s$ & $256 \times H/s \times W/s$ \\ \hline
3 & Norm1 & Layer 2 & InstanceNorm2d & $256 \times H/s \times W/s$ \\ \hline
4 & ReLU1 & Layer 3 & ReLU & $256 \times H/s \times W/s$ \\ \hline
5 & Conv2 & Layer 4 & $3 \times 3$ kernel, padding=1, stride=1 & $256 \times H/s \times W/s$ \\ \hline
6 & Norm2 & Layer 5 & InstanceNorm2d & $256 \times H/s \times W/s$ \\ \hline
7 & ReLU2 & Layer 6 & ReLU & $256 \times H/s \times W/s$ \\ \hline
\end{tabular}

\caption{\textbf{Description of ResidualBlock forward sequence}.}
\label{tab:modules_resnet}
\end{table}

\paragraph{Attentional Feature Fusion}

\label{sec:image_fusion}
In our framework, RGB and NIR images are first processed through the ResNet blocks, resulting in two separate 256-channel feature maps, $F_{v}^c$ and $F_{n}^c$. 
To integrate these feature maps into a unified representation, we employ an attentional feature fusion method that leverages spectral information while preserving essential details.
The fusion process comprises two main steps: the self-attention step and the attentional weight summation step.

\textbf{Self-Attention Step.} In this step, channel-level local and global attention mechanisms are applied to the feature maps $F_{v}^c$ and $F_{n}^c$. 
The attention-enhanced features, $A_{v}^c$ and $A_{n}^c$, are computed as follows:
\begin{align} 
A_u^c &= \frac{A_{v}^c + A_{n}^c}{M(A_{v}^c) + M(A_{n}^c)}, \label{eq:attention_map} \\
A_{v}^c &= F_{v}^c \circ M(F_{v}^c), \label{eq:attention_v} \\
A_{n}^c &= F_{n}^c \circ M(F_{n}^c), \label{eq:attention_n}
\end{align}
where $M$ denotes the self-attention module introduced in~\cite{Dai_2021_aff}, and $\circ$ represents element-wise multiplication. 
The unified attention map $A_u^c$ is obtained by normalizing the sum of the attention-enhanced features from both modalities.

\begin{figure}[h]
  \centering
  \includegraphics[width=0.7\linewidth]{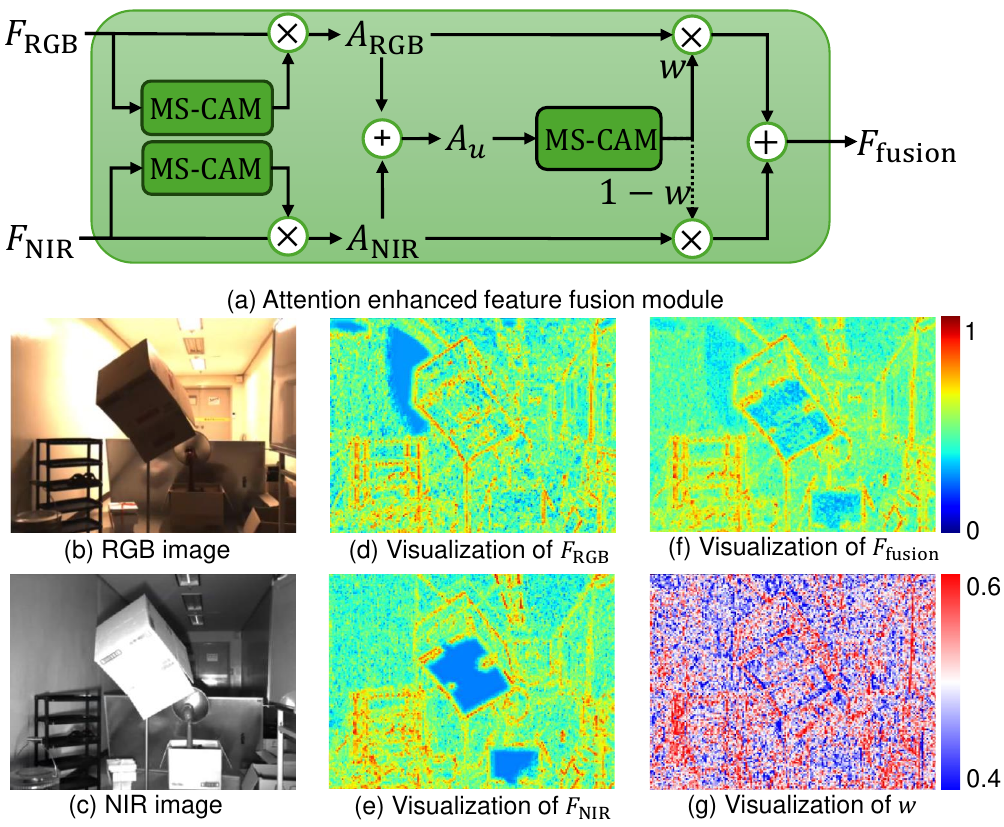}
  \caption{\textbf{Attention-enhanced feature fusion}. (a) The structure of attention based feature fusion module, proposed by ~\cite{Dai_2021_aff}. $\text{MS-CAM}$ is a attention enhancer module. The fusion operation is served by weight sum of attention-enhanced features with $w$.
  (b,c) Examples of input RGB and NIR images. (d,e,f,g) Visualizations of $F_{\text{RGB}}$~(d), $F_{\text{NIR}}$~(e), $F_\text{fusion}$~(f) and $w$ ~(g).} 
  \label{fig:aff}
\end{figure}

\textbf{Attentional Weight Summation Step.} Using the unified attention map $A_u^c$, we compute the fused feature map $F_{f}^c$ by performing a weighted summation of the attention-enhanced features from RGB and NIR:
\begin{equation}
    F_{f}^c = \left(A_{v}^c \circ M(A_u^c)\right) + \left(A_{n}^c \circ \left(1 - M(A_u^c)\right)\right).
\end{equation}
In this equation, $M(A_u^c)$ serves as a weighting factor that dynamically balances the contributions of RGB and NIR features based on the unified attention map. 
This weighted summation effectively integrates features from both modalities, resulting in a single, comprehensive fused feature map.
Figure~\ref{fig:aff} illustrates the detailed progress of attentional feature fusion using RGB and NIR feature maps. To provide a comprehensive and organized overview of the implementation, we have divided the attentional feature fusion module into four distinct PyTorch implementation tables. 
Specifically, Table~\ref{tab:aff_localattention} presents the LocalAttention module, 
Table~\ref{tab:GlobalAttentionModule} details the GlobalAttention module, and Table~\ref{tab:MSCAM} describes the MultiChannelAttention module. These individual components are then integrated to form the complete Attentional Feature Fusion module, as demonstrated in Table~\ref{tab:module_aff}.

\begin{table}[h]
\centering
\footnotesize
\begin{tabular}{|c|c|c|p{5cm}|c|}
\hline
\textbf{Layer} & \textbf{Layer Name} & \textbf{Input Layer} & \textbf{Description} & \textbf{Output Shape} \\ \hline
1 & Input & - & Input tensor & $\text{in\_channels} \times H \times W$ \\ \hline
2 & Local\_Conv1 & Layer 1 & Conv2d with kernel size $1 \times 1$, stride $1$ & $(\text{in\_channels}/\text{reduction}) \times H \times W$ \\ \hline
3 & Local\_BN1 & Layer 2 & BatchNorm applied to the output of Layer 1 & $(\text{in\_channels}/\text{reduction}) \times H \times W$ \\ \hline
4 & Local\_ReLU & Layer 3 & ReLU activation function applied to the output of Layer 2 & $(\text{in\_channels}/\text{reduction}) \times H \times W$ \\ \hline
5 & Local\_Conv2 & Layer 4 & Conv2d with kernel size $1 \times 1$, stride $1$ & $\text{in\_channels} \times H \times W$ \\ \hline
6 & Local\_BN2 & Layer 5 & BatchNorm applied to the output of Layer 4 & $\text{in\_channels} \times H \times W$ \\ \hline
7 & Output & Layer 6 & Output tensor returned by the module & $\text{in\_channels} \times H \times W$ \\ \hline
\end{tabular}
\caption{\textbf{LocalAttentionModule}}
\label{tab:aff_localattention}
\end{table}

\begin{table}[h]
\centering
\footnotesize
\begin{tabular}{|c|c|c|p{5cm}|c|}
\hline
\textbf{Layer} & \textbf{Layer Name} & \textbf{Input Layer} & \textbf{Description} & \textbf{Output Shape} \\ \hline
1 & Input & - & Input tensor & $\text{in\_channels} \times H \times W$ \\ \hline
2 & Global\_AvgPool & Layer 1 & Adaptive Average Pooling to output size $1 \times 1$ & $\text{in\_channels} \times 1 \times 1$ \\ \hline
3 & Global\_Conv1 & Layer 2 & Conv2d with kernel size $1 \times 1$, stride $1$ & $(\text{in\_channels}/\text{reduction}) \times 1 \times 1$ \\ \hline
4 & Global\_BN1 & Layer 3 & BatchNorm applied to the output of Layer 2 & $(\text{in\_channels}/\text{reduction}) \times 1 \times 1$ \\ \hline
5 & Global\_ReLU & Layer 4 & ReLU activation function applied to the output of Layer 3 & $(\text{in\_channels}/\text{reduction}) \times 1 \times 1$ \\ \hline
6 & Global\_Conv2 & Layer 5 & Conv2d with kernel size $1 \times 1$, stride $1$ & $\text{in\_channels} \times 1 \times 1$ \\ \hline
7 & Global\_BN2 & Layer 6 & BatchNorm applied to the output of Layer 6 & $\text{in\_channels} \times 1 \times 1$ \\ \hline
8 & Output & Layer 7 & Output tensor returned by the module & $\text{in\_channels} \times 1 \times 1$ \\ \hline
\end{tabular}
\caption{\textbf{GlobalAttentionModule}.}
\label{tab:GlobalAttentionModule}
\end{table}

\begin{table}[h]
\centering
\footnotesize
\begin{tabular}{|c|c|c|p{5cm}|c|}
\hline
\textbf{Layer} & \textbf{Layer Name} & \textbf{Input Layer} & \textbf{Description} & \textbf{Output Shape} \\ \hline
1 & Input & - & Input tensor & $\text{in\_channels} \times H \times W$ \\ \hline
2a & Local\_Attention & Layer 1 & LocalAttentionModule & $\text{in\_channels} \times H \times W$ \\ \hline
2b & Global\_Attention & Layer 1 & GlobalAttentionModule & $\text{in\_channels} \times 1 \times 1$ \\ \hline
3 & Addition & Layers 2a / 2b & Adds outputs of Layer 2a and Layer 2b & $\text{in\_channels} \times H \times W$ \\ \hline
4 & Sigmoid & Layer 3 & Applies Sigmoid activation & $\text{in\_channels} \times H \times W$ \\ \hline
\end{tabular}
\caption{\textbf{MultiScaleChannelAttentionModule (MS-CAM)}.}
\label{tab:MSCAM}
\end{table}

\begin{table}[h]
\footnotesize
\centering

\label{tab:AttentionFusion}

\centering
\footnotesize
\begin{tabular}{|c|c|c|p{5cm}|c|}
\hline
\textbf{Layer} & \textbf{Layer Name} & \textbf{Input Layer} & \textbf{Description} & \textbf{Output Shape} \\ \hline
1a & Input\_RGB & - & RGB input tensor & $\text{in\_channels} \times H \times W$ \\ \hline
1b & Input\_NIR & - & NIR input tensor & $\text{in\_channels} \times H \times W$ \\ \hline
2a & Attention\_RGB & Layer 1a & MS-CAM & $\text{in\_channels} \times H \times W$ \\ \hline
2b & Attention\_NIR & Layer 1b & MS-CAM & $\text{in\_channels} \times H \times W$ \\ \hline
3 & Addition & Layers 2a/ 2b & Adds Attention\_RGB and Attention\_NIR & $\text{in\_channels} \times H \times W$ \\ \hline
4a & RGB\_Scaled & Layers 1a/ 3 & Multiplies RGB input by (Layer 2a / Layer 3) & $\text{in\_channels} \times H \times W$ \\ \hline
4b & NIR\_Scaled & Layers 1b/ 3 & Multiplies NIR input by (Layer 2b / Layer 3) & $\text{in\_channels} \times H \times W$ \\ \hline
5 & Addition & Layers 4a/ 4b & Adds RGB\_Scaled and NIR\_Scaled & $\text{in\_channels} \times H \times W$ \\ \hline
6 & Attention\_Fusion & Layer 5 & MS-CAM & $\text{in\_channels} \times H \times W$ \\ \hline
7& Feature\_Fusion & Layers 6, 1a, 1b & Combines Attention fusion with original RGB and NIR inputs to produce final output: $\text{Output} = (\text{Layer 5}) \times \text{RGB} + (1 - (\text{Layer 5})) \times \text{NIR}$ & $\text{in\_channels} \times H \times W$ \\ \hline
\end{tabular}
\caption{\textbf{AttentionFeatureFusion}.}
\label{tab:module_aff}
\end{table}

\paragraph{Image Fusion Model}

Table~\ref{tab:HSVNet_forward} presents a detailed implementation of our image fusion network. Feature maps extracted from a pretrained feature encoder~\cite{lipson2021raft} are integrated using an attentional feature fusion mechanism~\cite{Dai_2021_aff}. These fused feature maps are subsequently processed through residual blocks to generate spatially varying weights, $\alpha$ and $\beta$, constrained within the range [0, 1]. Utilizing these weights, the fused image is computed as follows:

\begin{equation} I_\text{fusion} = M_{\text{HSV}\rightarrow\text{RGB}}[I_H,I_S,I_{\hat{V}}]^\intercal = M_{\text{HSV}\rightarrow\text{RGB}}[I_H,I_S,\alpha I_V+\beta I_\text{NIR}]^\intercal \end{equation}

Through this approach, we achieve the fusion of RGB and NIR images based on the derived spatially varying weights

\paragraph{NIR Guided RGB Filtering}
We employ an NIR-guided RGB filtering technique to enhance input RGB images by leveraging a single-channel Near-Infrared (NIR) image as the guiding reference \cite{Li_2013_guidedfiltering_fusion}. Our method is grounded in the guided filter framework, where the NIR image $ I_{\text{NIR}} $ serves as the guide, and the fused image $ I_{\text{fusion}} $ is the input to be filtered. For each color channel  $i$  of $ I_{\text{fusion}} $, we first compute the local means $ \mu_{\text{NIR}} $ and $ \mu_{\text{fusion},i} $, as well as the covariance $ \text{cov}(I_{\text{NIR}}, I_{\text{fusion},i}) $ and the variance $ \text{var}(I_{\text{NIR}}) $ within a window of radius $ r $. The linear coefficients $ a_i $ and $ b_i $ are then determined using the equations
\begin{equation}
    a_i = \frac{\text{cov}(I_{\text{NIR}}, I_{\text{fusion,i}})}{\text{var}(I_{\text{NIR}}) + \epsilon}, \quad b_i = \mu_{\text{fusion,i}} - a_i \mu_{\text{NIR}},
\end{equation}
where $ \epsilon $ is a regularization parameter that ensures numerical stability. Subsequently, we calculate the mean values of $ a^c $ and $ b^c $ within the same window and reconstruct the output RGB image $ \mathbf{q} $ for each channel using
\begin{equation}
    I_{\text{filtered},i} = \text{mean}(a_i) I_{\text{NIR}} + \text{mean}(b_i),
\end{equation}
ensuring that the resulting pixel values are clamped within the valid range [0, 255]. This NIR-guided filtering approach effectively utilizes the structural information from the NIR guide to enhance color fidelity and suppress noise in the images which is generated during fusion process, demonstrating improved performance in various image processing applications.

\begin{table}[h]
\centering
\footnotesize
\begin{tabular}{|c|c|c|p{6cm}|c|}
\hline
\textbf{Layer} & \textbf{Layer Name} & \textbf{Input Layer} & \textbf{Description} & \textbf{Output Shape} \\ \hline
1a & Input\_RGB & - & Receives RGB input tensor & $3 \times H \times W$ \\ \hline
1b & Input\_NIR & - & Receives NIR input tensor & $1 \times H \times W$ \\ \hline
2 & RGB\_to\_HSV & Layer 1a & Converts RGB tensor to HSV color space & $3 \times H' \times W'$ \\ \hline
3a & Encoder\_RGB & Layer 1a (RGB) & Passes normalized RGB through BasicEncoder~\cite{lipson2021raft} & $256 \times \frac{H'}{2} \times \frac{W'}{2}$ \\ \hline
3b & Encoder\_NIR & Layer 1b (NIR) & Passes normalized NIR (repeated across channels) through BasicEncoder & $256 \times \frac{H'}{4} \times \frac{W'}{4}$ \\ \hline
4 & Attentional\_Feature\_Fusion & Layers 3a and 3b & Combines HSV and NIR feature maps & $256 \times \frac{H'}{4} \times \frac{W'}{4}$ \\ \hline
5 & Residual\_Block & Layer 4 & Reduces channel dimensions using ResidualBlocks, followed by Sigmoid activation & $2 \times \frac{H'}{4} \times \frac{W'}{4}$ \\ \hline
6 & Upsample\_Weights & Layer 5 & Upsamples weights by a factor of 4 using bilinear interpolation & $2 \times H' \times W'$ \\ \hline
7 & Combine\_HSV\_and\_NIR & Layers 1b, 2, 6& $[I_H,I_S,I_{\hat{V}}] = [I_H,I_S,\alpha I_V+\beta I_\text{NIR}]$ & $3 \times H' \times W'$ \\ \hline
8 & Convert\_to\_RGB & Layer 7 & Converts combined HSV features back to RGB color space & $3 \times H' \times W'$ \\ \hline
\end{tabular}
\caption{Forward Pass of our Image Fusion Model}
\label{tab:HSVNet_forward}
\end{table}

\subsubsection{Loss Function}

\paragraph{Photometric Loss}

The synthetic dataset includes original RGB images prior to augmentation. 
To ensure that the fused output of augmented RGB and NIR images closely matches the original RGB images, we design a photometric loss function defined as

\begin{equation}
    \mathcal{L}_{\text{photometric}} = \gamma_{\text{L1}} \| I_{\text{fusion}} - I_{\text{original}} \|_1 + \gamma_{\text{ssim}} \, \text{SSIM}(I_{\text{fusion}}, I_{\text{original}}).
\end{equation}

In this formulation, the L1 loss and Structural Similarity Index Measure (SSIM) loss are weighted by coefficients $\gamma_{\text{L1}}$ and $\gamma_{\text{ssim}}$, respectively, set to 0.85 and 0.15. 
This combination ensures that the fused image maintains both pixel-wise accuracy and structural similarity to the original RGB image, facilitating effective training of the fusion model.

\paragraph{Ground-truth Disparity Loss}
To guide our network towards accurate operation, we trained it using a combination of an augmented synthetic dataset and our real dataset, utilizing the $L1$ loss function. tarting from a pretrained model based on \cite{lipson2021raft}, we loaded pretrained weights, excluding the modified structures. Then we fine-tuned our model by freezing weights excluding the modified ones.
For the predicted disparity map $d_\text{pred}$, we compute $L1$ loss by
\begin{equation}
    \mathcal{L}_\text{gt} = \sum_{u,v \in \Omega_\text{gt}}|d_\text{gt}(u,v) - d_\text{pred}(u,v)| 
\end{equation}
where  $\Omega_\text{gt}$ is a binary mask indicating valid regions in the labeled disparity map and $d_\text{gt}$ is ground-truth disparity map from dataset.

\subsubsection{Training Details}

We utilized our augmented RGB-NIR synthetic dataset, comprising approximately 50,000 original images. For each training batch, different augmentations were dynamically applied to the data to enhance diversity. 
The fusion model architecture consists of a pretrained feature encoder and a fusion module. 
During training, the feature encoder was frozen, and only the fusion module was updated to focus on learning effective fusion strategies.
To improve the fidelity of the image fusion restoration process, we employed a photometric loss function. 
Specifically, the fused images were passed through Raft-Stereo to obtain disparity maps, which were then compared to the ground truth disparity maps. 
Although Raft-Stereo was configured to compute gradients, it was frozen and excluded from the optimizer to ensure that only our fusion network was trained.

For optimization, we used the Adam optimizer in conjunction with gradient scaling, initializing the gradient scaler with a value of 1024. 
To accelerate training and optimize memory usage, mixed precision training was enabled. The training process was conducted with a batch size of 4 per GPU, utilizing four RTX 3090 GPUs (24 GB each) in a distributed parallel setup, resulting in an effective batch size of 16. 
The model was trained for approximately 20 epochs using this hardware configuration.

\subsection{RGB-NIR Feature Fusion}
\subsubsection{Network Architecture}

We begin by processing the RGB and NIR stereo images through a shared ResNet-based feature extractor $ f_{\text{enc}} $. For each input image $ I_s^{c} $, where $ s \in \{\text{RGB}, \text{NIR}\} $ and $ c \in \{\text{left}, \text{right}\} $, the encoder transforms it into feature maps:
\begin{equation}
F_s^{c} = f_{\text{enc}}(I_s^{c}), \quad \text{for } s \in \{\text{RGB}, \text{NIR}\}, \, c \in \{\text{left}, \text{right}\}.
\end{equation}
Next, we apply the attention-based fusion method~\cite{Dai_2021_aff} from Section~\ref{sec:image_fusion} to combine the RGB and NIR features:
\begin{equation}
F_{\text{fusion}}^{c} = f_{\text{fusion}}(F_{\text{RGB}}^{c}, F_{\text{NIR}}^{c}).
\end{equation}

The core of our method lies in constructing correlation volumes that capture the relationships between the left and right feature maps. We compute correlation volumes for both the fused features and the NIR features:
\begin{equation}
V_s(x, y, k) = F_s^{\text{left}}(x, y) \cdot F_s^{\text{right}}(x + k, y), \quad s \in \{\text{fusion}, \text{NIR}\},
\end{equation}
where $ (x, y) $ denotes the pixel location, $ k $ is the disparity index, and $ \cdot $ represents the inner product.

We employ an iterative approach using the GRU structure of the RAFT-Stereo network to refine disparity estimates. The process begins with an initial disparity $ d_0 = 0 $ and progressively updates this estimate. At each iteration $ n $, the update model takes the following inputs:
\begin{itemize}
    \item The previous disparity estimate, $ d_n $.
    \item The context feature map from the fused left image, $ F_{\text{fusion}}^{\text{left}} $.
    \item The sampled correlation volume $ V_{\text{sampled}} $, which alternates between $ V_{\text{fusion}} $ and $ V_{\text{NIR}} $.
\end{itemize}
The update model predicts a disparity increment $ \Delta d $, which is added to the previous disparity to obtain the next estimate:
\begin{equation}
H_{n+1} = \text{ConvGRU}(H_n, d_n, [V_{\text{sampled}}, F_{\text{fusion}}^{\text{left}}]),
\end{equation}
where $ H_n $ is the hidden state of the ConvGRU layer, initialized with $ F_{\text{fusion}}^{\text{left}} $.

Then the increment of disparity is computed by a single convolutional layer and it updates the disparity estimate:
\begin{equation}
d_{n+1} = d_n + \text{Conv}(H_{n+1}).
\end{equation}
By alternating between the fused and NIR correlation volumes at each iteration, our method effectively leverages spectral information from both RGB and NIR images. This approach places more emphasis on NIR features that are robust to environmental lighting, resulting in more accurate depth estimation as demonstrated in Table~\ref{tab:ablation_alternation}.

We fine-tune our model on the synthetic and real training datasets using the disparity reconstruction loss and the LiDAR loss, respectively. For more details on the loss functions and optimization, we refer to the Supplemental Document.

\subsubsection{Loss Function}

\paragraph{Ground-truth LiDAR Depth Loss}
We implemented pretrained weight provided by original author, which is trained with RGB stereo dataset, and fine-tuned this using a combination of an augmented synthetic dataset and our real dataset, utilizing the $L1$ loss function.
For a series of disparity estimations $\{d_1,d_2,d_3\dots d_n\}$ we compute $L1$ loss by
\begin{equation}
    L_{\text{gt}} = \sum_i^N w^{n-i}\sum_{u,v \in \Omega_{\text{gt}}}|d_{\text{gt}}(u,v) - d_i(u,v)| 
\end{equation}
where $w$ is weight factor for normalized summation of series of $d$ and $\Omega_{\text{gt}}$ is a binary mask indicating valid regions in the labeled disparity map.
We also use LiDAR points $(u_i,v_i,z_{\text{gt},i}) \in \mathbb{R}_{I^l}^3$, which is projected into image $I^l$ coordinate, as ground truth for depth estimation on real-world data:
\begin{equation}
    L_\text{LiDAR} = \sum_i^N \sum_{(x,y) \in \mathcal{N}(r)} \left| z_{\text{pred}}(u_i + x, v_i + y) - z_{\text{gt},i} \right|
\end{equation}
where $\mathcal{N}(r)$ is a set of point offset, making local neighborhood box and $r$ is radius of it, presently $r=5$, $z_{\text{pred}} = \frac{f_x\cdot B}{d_n}$ and $f_x$ and $B$ is focal length of camera and baseline distance, which is constant of our system.

\subsubsection{Training Detail}
\label{sec:rgb_nir_feature_fusion}

\paragraph{Feature Extraction of RGB and NIR Images}

We employed the same encoder structure and weights for feature extraction from both RGB and NIR images. 
This pretrained encoder, originally trained on RGB images, is effectively applicable to NIR data. 
By replicating the single-channel NIR image into three channels, we achieved stereo depth estimation using RAFT-Stereo~\cite{lipson2021raft} and CREStereo~\cite{Li_2022_CREStereo}. 
Similarly, replicating a single-channel image derived from converting RGB to grayscale or from a monocular visible-light camera into three channels did not present issues within the encoder, regardless of the channel count or target color space. 
This is because stereo depth estimation primarily involves analyzing the spatial correlation between pixels in the two images, and the correlation definition remains unaffected by changes in the spectral domain. 
To leverage the advantages of fine-tuning pretrained weights, we used the same encoder for both RGB and NIR images and kept the encoder frozen during training.
\paragraph{Supervised Training}

We conducted supervised training using our augmented synthetic dataset, which includes RGB-NIR stereo images and highly detailed, pixel-wise disparity maps.
This allowed for supervised learning by directly comparing the predicted disparity maps generated by the model with the ground truth disparity maps. 
To mitigate the domain gap between real and synthetic data, we initially trained for one epoch solely on synthetic data and then incorporated real data into the training dataset. 
The real data includes LiDAR-derived depth information and pseudo disparity maps, along with occlusion maps obtained through sparse LiDAR reconstruction. 
Using these components, we were able to perform supervised learning with the previously defined loss function.

\paragraph{Self-supervised Training}
Once the disparity map $d^\text{left}$ is obtained, it can be used to warp stereo images. The warped image $I^{\text{left}|\text{right}}$ that left image $I^\text{left}$ is warped into right image coordinate, is computed by:
\begin{align}
x' &= x - d^\text{left}(x, y) \\
I^{\text{left}|\text{right}}(x, y) &= I^\text{left}(x - d^\text{left}(x, y), y)
\end{align},
where $x,y \in \Omega$ and $\Omega$ is image plain coordinate.

The photometric consistency between the warped and original images can be leveraged for self-supervised learning:
\begin{equation}
\mathcal{L}_\text{photometric} = \left| I^{\text{left}|\text{right}} - I^\text{right} \right| + \text{SSIM}\left( I^{\text{left}|\text{right}}, I^\text{right} \right)
\end{equation}

However, this approach presents challenges when there are brightness differences between the left and right images, making it difficult for the loss to converge. 
Furthermore, when scene disparities are not pronounced, the warping loss may not effectively differentiate between accurate and erroneous disparity values, even if the disparity estimates are correct or grossly inaccurate. 
To ensure comprehensive data collection across both indoor and outdoor environments, we set the stereo baseline to 133mm. 
This setup results in a relatively narrow disparity range in outdoor scenes.
Consequently, we utilized self-supervised learning only as an auxiliary method to support supervised learning on our real dataset.

\paragraph{Finetuning}

We fine-tuned the pretrained weights of RAFT-Stereo~\cite{lipson2021raft} to adapt and extend its capabilities. 
The correlation sampler, which computes the correlation volume $V$, is not a neural network structure and was therefore excluded from the learning process. 
To leverage the performance of the pretrained feature encoder, its parameters were frozen during training. 
Similarly, the iterative convolutional GRU updater was also frozen to retain its pretrained functionality.

In our design, an attentional feature fusion module~\cite{Dai_2021_aff} was incorporated to generate a fused feature map $F^c_\text{fusion}$ with $F^c_\text{RGB},F^c_\text{NIR}$. 
As such, the attentional feature fusion module was set to trainable mode to enable learning during fine-tuning.

RAFT-Stereo originally implements the context encoder with a structure identical to the feature encoder, differing only in the number of channels to accommodate separate weights.
The context encoder serves as an input to update the hidden state of the GRU updater. 
To optimize the design and reduce the frequency of attentional fusion operations, we removed the context encoder.
Instead, the features obtained from the feature encoder were directly forwarded to the inputs originally designated for context features. 
Consequently, this modification required additional learning in the process of forwarding the feature map to the GRU.

To ensure robust performance across different scenarios, batch normalization layers were frozen throughout the fine-tuning process. 
This approach preserved the stability of normalization statistics while allowing for effective adaptation of the model's new components.

\section{Additional Result}

\subsection{Implementation of RGB-NIR Image Fusion Methods for Comparative Analysis}
\begin{figure}[h]
  \centering
  \includegraphics[width=\linewidth]{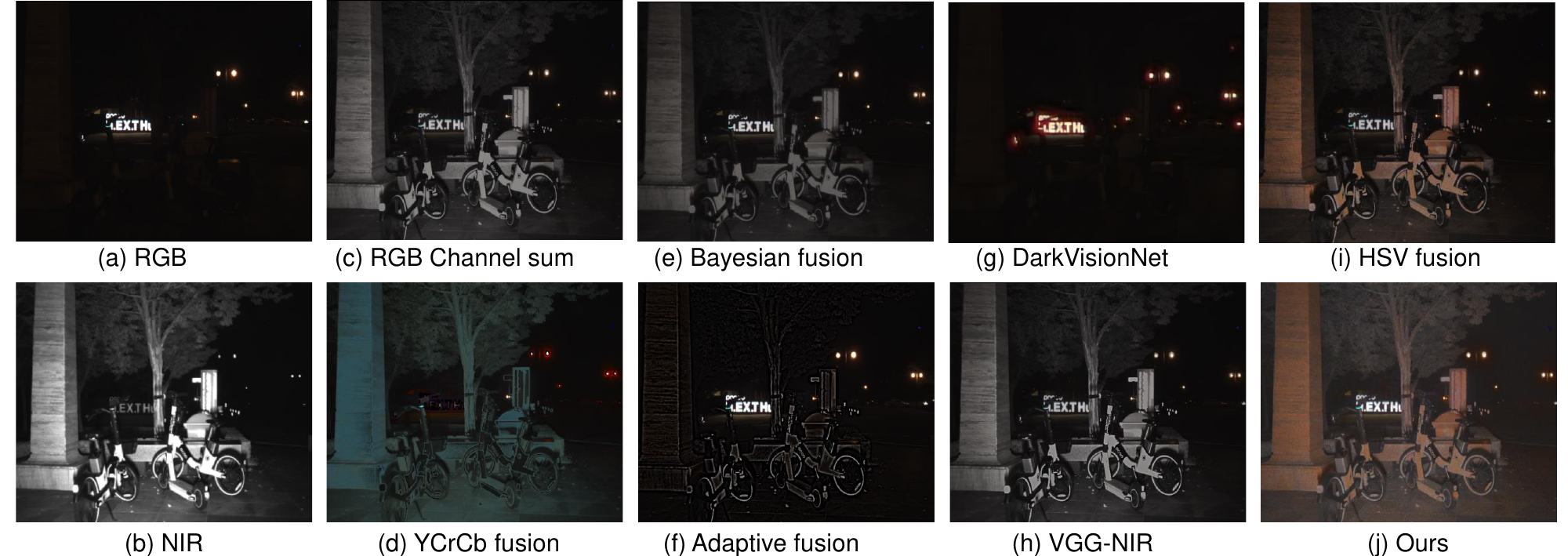}
  \caption{\textbf{Image fusion visualization of ours and comparison methods}. (a,b) Pixel-aligned RGB and NIR images. (c) Image driven by simple channel sum. (d) YCrCb channel fusion~\cite{herrera2019color}. (e) Bayesian estimation based fusion~\cite{zhao2020bayesian_fusion}. (f) Gradient adaptive fusion~\cite{awad2019adaptive_rgb_nir}. (g) DarkVisionNet~\cite{Jin_2023_DarkVision}. (h) VGG based fusion~\cite{li2018infrared_vgg}. (i) HSV chanenl fusion (our baseline)~\cite{fredembach2008colouring_hsv}, (j) Our fusion method. }
  \label{fig:image_fusion_compari}
\end{figure}

\begin{figure}[h]
  \centering
  \includegraphics[width=\linewidth]{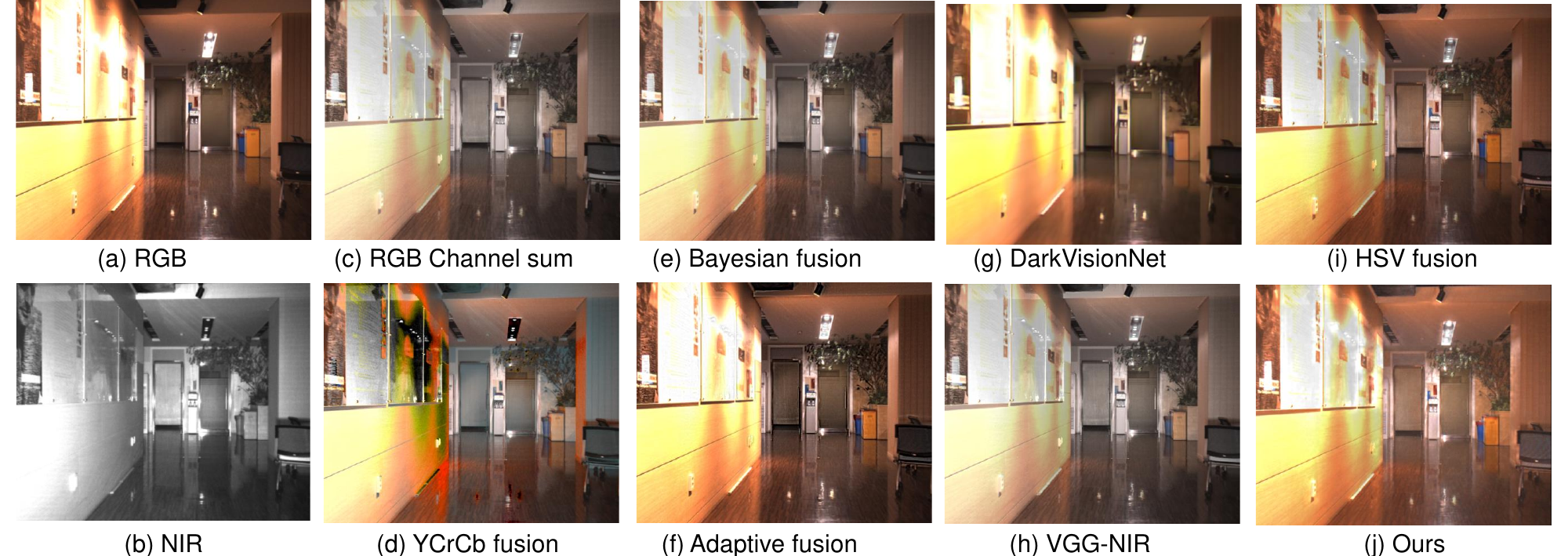}
  \caption{\textbf{Image fusion visualization of ours and comparison methods, indoor sample}.  (a,b) Pixel-aligned RGB and NIR images. (c) Image driven by simple channel sum. (d) YCrCb channel fusion~\cite{herrera2019color}. (e) Bayesian estimation based fusion~\cite{zhao2020bayesian_fusion}. (f) Gradient adaptive fusion~\cite{awad2019adaptive_rgb_nir}. (g) DarkVisionNet~\cite{Jin_2023_DarkVision}. (h) VGG based fusion~\cite{li2018infrared_vgg}. (i) HSV chanenl fusion (our baseline)~\cite{fredembach2008colouring_hsv}, (j) Our fusion method. }
  \label{fig:image_fusion_compari_2}
\end{figure}

Figure~\ref{fig:image_fusion_compari} and~\ref{fig:image_fusion_compari_2} illustrate different image fusion methods applied to our real RGB-NIR dataset. We conducted comparative studies on downstream vision applications using our image fusion model alongside other color fusion techniques.

\paragraph{Bayesian Fusion \cite{zhao2020bayesian_fusion}}
This method employs Bayesian estimation to fuse two different mono-channel images. We implemented per-channel fusion for the R, G, and B channels with NIR using the Bayesian Fusion algorithm. The fusion process involves an iterative optimization procedure that incorporates total variation (TV) regularization to preserve edge details and reduce noise. Specifically, the fusion is achieved by solving the following optimization problem:

\begin{equation}
\min_{C} \|k_1 * C - D_1\|^2 + \|k_2 * C - D_2\|^2 + \|X - C\|^2 + \lambda \, \text{TV}(C)
\end{equation}

where $ C $ is the fused image channel, $ D_1 $ and $ D_2 $ are the degraded observations from the RGB and NIR channels respectively, $ k_1 $ and $ k_2 $ are convolution kernels, $ X $ is the current estimate of the fused image, and $ \lambda $ is a regularization parameter controlling the strength of the TV term. The optimization is performed using proximal gradient methods, where the TV regularization is handled by the proximal operator:
\begin{equation}
C^{(t+1)} = \text{prox}_{\lambda \, \text{TV}} \left( \frac{2 C Y + \rho H}{2 C + 2 D + \rho} \right),
\end{equation}
where $ Y $ represents the difference between the NIR and RGB images, $ \rho $ is a scaling factor, and $ H $ is updated through the proximal TV operation. The iterative process continues for a predefined number of iterations to obtain the final fused channel. The fused RGB image is then constructed by stacking the individually fused R, G, and B channels. We prepared the comparison using the source code provided by the authors, ensuring an accurate implementation of the Bayesian Fusion method for our dataset.

\paragraph{DarkVision \cite{Jin_2023_DarkVision}}
DarkVision employs a Multi-Layer Perceptron (MLP) network to enhance and denoise dark RGB images with guidance from high-quality NIR images. We utilized the pretrained version of this network for our comparisons, focusing on its ability to improve image quality without incorporating explicit mathematical formulations.

\paragraph{Infrared-VGG Fusion \cite{li2018infrared_vgg}}
This method is based on the VGG architecture and fuses infrared and visible images using a pretrained VGG module as the encoder. We followed the installation instructions provided by the authors and utilized their publicly available code to set up the comparison environment, enabling a direct evaluation of the fusion performance without delving into the underlying equations.

\paragraph{YCbCr Channel Fusion \cite{herrera2019color}}
The YCbCr Channel Fusion method leverages the superior detail preservation of NIR images under various environmental conditions by fusing the luminance (Y) channel in the YCbCr color space. The fusion process is carried out as follows:
First, the RGB image is converted to the YCbCr color space, and its grayscale version is obtained.
\begin{equation}
\begin{bmatrix}
    l_{RGB} \\ C_1 \\ C_2
\end{bmatrix}
    = M_{\text{RGB}\rightarrow\text{YCbCr}}I_\text{RGB}\begin{bmatrix}
        I_R \\ I_G \\ I_B
    \end{bmatrix}.
\end{equation}
The fusion weight is computed by normalizing the difference between the NIR and grayscale images:
\begin{equation}
l_{\text{V}} = \frac{I_{\text{NIR}} - l_{\text{RGB}}}{I_\text{max}},
\end{equation}
where $ I_{\text{NIR}} $ is the intensity of the NIR image and $ I_{\text{Gray}} $ is the intensity of the grayscale RGB image. $I_\text{max}$ is the maximum intensity value for normalization scaling the difference to the range [0, 1].
The fused luminance channel is then calculated as:
\begin{equation}
l_{\text{fused}} = l_{\text{RGB}} \cdot l_{\text{V}} + I_{\text{NIR}} \cdot (1 - l_{\text{V}}),
\end{equation}
where $ l_{\text{RGB}} $ represents the luminance component of the YCbCr image.
To enhance the chromatic channels, a scaling factor $ m $ is determined by:
\begin{equation}
m = \frac{l_{\text{RGB}} - l_{\text{fused}}}{l_{\text{RGB}}}, \quad \text{where} \quad m = 0 \quad \text{if} \quad l_{\text{RGB}} = 0.
\end{equation}
This ensures numerical stability by avoiding division by zero.
The chromatic channels are then adjusted using the scaling factor and then the fused YCbCr image is reconstructed by stacking the fused luminance and adjusted chromatic channels:
\begin{align}
    C_{1,\text{fused}} = C_{1} \cdot (1 + m)\text{, } C_{2,\text{fused}} = C_{2} \cdot (1 + m) \\
    I_{\text{fused}} = M_{\text{YCbCr}\rightarrow\text{RGB}}[l_{\text{fused}}, C_{1,\text{fused}}, C_{2,\text{fused}}]^\intercal.
\end{align}

This fused YCbCr image is then converted back to the RGB color space to obtain the final fused image. We prepared the comparison based on the equations described in the paper, implementing the fusion process as outlined in the authors' source code.

\paragraph{HSV Channel Fusion \cite{fredembach2008colouring_hsv}}
This method transforms the images into the HSV color space and fuses the V channel by addition:
\begin{equation}
V_\text{fused} = 0.5 V_\text{RGB} + 0.5 I_\text{NIR}.
\end{equation}
The resulting HSV image is then converted back to RGB. We prepared the comparison by blending the V channel and NIR image at a 0.5:0.5 ratio.

\paragraph{Adaptive RGB-NIR Fusion \cite{awad2019adaptive_rgb_nir}}
The Adaptive RGB-NIR Fusion method enhances the edges of the RGB image by leveraging the pixel gradients from the NIR image. The fusion process consists of several key steps: local contrast computation, fusion map generation, high-pass filtering, and image enhancement. 

The first step is a local contrast computation.
The method computes the local contrast for both the luminance component of the YCbCr-transformed RGB image and the NIR image. The YCbCr transformation is defined as:
\begin{equation}
\begin{bmatrix}
    l_{RGB} \\ C_1 \\ C_2
\end{bmatrix}
    = M_{\text{RGB}\rightarrow\text{YCbCr}}I_\text{RGB}\begin{bmatrix}
        I_R \\ I_G \\ I_B
    \end{bmatrix},
\end{equation}
where $ l_\text{RGB} $ is the luminance (Y channel), and $ C_1, C_2 $ represent the chrominance channels (Cb and Cr, respectively).  
For an intensity image $ I $ (either luminance $ l_\text{RGB} $ or NIR intensity $ I_{\text{NIR}} $), the local contrast $ \text{LocalContrast}_I $ is defined as a combination of intensity variation and amplitude variation within a local window:
\begin{equation}
\text{LocalContrast}_I = \alpha \cdot (I_{\text{max}} - I_{\text{min}}) + (1 - \alpha) \cdot \text{MaxAmplitude}(I),
\end{equation}
where:
\begin{itemize}
    \item $ I_{\text{max}} $ and $ I_{\text{min}} $ are the maximum and minimum intensity values within the local window.
    \item $ \text{MaxAmplitude}(I) $ is the maximum amplitude obtained from the gradient magnitude computed using Sobel filters:
    \begin{equation}
    \text{MaxAmplitude}(I) = \max\left(\sqrt{\left(\frac{\partial I}{\partial x}\right)^2 + \left(\frac{\partial I}{\partial y}\right)^2}\right)
    \end{equation}
    \item $ \alpha $ is a weighting factor (set to 0.5 in our implementation).
\end{itemize}

The next step is generating the fusion map.
Using the local contrasts $ \text{LocalContrast}_{l_{RGB}} $ and $ \text{LocalContrast}_{\text{NIR}} $, the fusion map $ \text{FusionMap} $ is generated to determine the regions where the NIR image can provide enhanced details:
\begin{equation}
\text{FusionMap} = \frac{\max(0, \text{LocalContrast}_{\text{NIR}} - \text{LocalContrast}_{l_\text{RGB}})}{\max(\text{LocalContrast}_{\text{NIR}}, \epsilon)},
\end{equation}
where $ \epsilon $ is a small constant (e.g., $ 1 \times 10^{-6} $) to prevent division by zero. This fusion map emphasizes areas where the NIR image has higher local contrast compared to the RGB image.

\textbf{A high-pass filter is then applied} to the NIR image to extract high-frequency details:
\begin{equation}
\text{HPF}(I_{\text{NIR}}) = I_{\text{NIR}} - \text{GaussianBlur}(I_{\text{NIR}}, \sigma),
\end{equation}
where $ \text{GaussianBlur} $ applies a Gaussian filter with a specified kernel size (e.g., $ 19 \times 19 $) to smooth the NIR image, and $ \sigma $ is the standard deviation of the Gaussian kernel.

The final enhancement is performed by adding the product of the fusion map and the high-pass filtered NIR image to each channel of the YCbCr-converted RGB image. The fused YCbCr image is then converted back to the RGB color space.
The final fused RGB image $ I_{\text{fused}} $ is computed as:
\begin{align}
I_\text{HDF} = \text{FusionMap} \cdot \text{HPF}(I_{\text{NIR}})
\\
I_{\text{fused}} = M_{\text{YCbCr}\rightarrow\text{RGB}}
\begin{bmatrix}
 l_{RGB} + I_\text{HDF} \\  
 C_1 + I_\text{HDF} \\  
 C_2 + I_\text{HDF}
\end{bmatrix}.
\end{align}
Finally, the fused RGB image $ I_{\text{fused}} $ is clipped to the valid intensity range $ [0, 255] $ to ensure proper image representation.

\subsection{RGB-NIR Image Fusion for Object Detection}
We evaluated the performance of our color fusion approach on object detection tasks~\cite{yolov8_ultralytics}. 
\subsubsection{Evaluation}

To evaluate the performance of our object detection model, we employed standard metrics commonly used in single-class multi-object detection tasks. 
Specifically, we utilized precision, recall, F1-score, mean Average Precision (mAP), and average Intersection over Union (IoU) as our primary evaluation metrics.
Given that the synthetic dataset was generated with precise object index information, we were able to construct accurate ground-truth annotations for object detection. 
For the real dataset, we prepared the data by directly creating pseudo labels to serve as ground-truth annotations.

\subsubsection{Additional Results}

Table~\ref{tab:fusion_comparison_detection} presents a comprehensive evaluation of various image fusion methods using YOLO~\cite{yolov8_ultralytics} for object detection across multiple performance metrics. 
Notably, our proposed RGB-NIR fusion technique achieves the highest Detection mAP (0.828), surpassing the second-best Bayesian fusion method (0.773) by a significant margin. 
This superior performance is consistently reflected across all evaluated metrics, including Average IOU (0.509), F1 score (0.688), Precision (0.734), and Recall (0.687), indicating a robust enhancement in both localization and classification capabilities. 
In comparison, traditional methods such as HSV baseline~\cite{fredembach2008colouring_hsv} and YCrCb~\cite{herrera2019color} demonstrate lower performance, with Detection mAP values of 0.744 and 0.745, respectively. 
Furthermore, specialized approaches like DarkVision~\cite{Jin_2023_DarkVision} exhibit considerably lower metrics, highlighting the effectiveness of our fusion strategy. 
The consistent outperformance across diverse metrics underscores the efficacy of our RGB-NIR fusion method in enhancing object detection accuracy, precision, and recall, thereby establishing it as a superior choice for image fusion in object detection tasks.

\begin{table}[h]
\small
\centering
\resizebox{0.8\linewidth}{!}{%
\begin{tabular}{|c|c|c|c|c|c|}
\hline
\textbf{Methods} & \textbf{Detection mAP~$\uparrow$}& \textbf{Average IOU~$\uparrow$}& \textbf{F1 score~$\uparrow$}& \textbf{Precision~$\uparrow$}&\textbf{Recall~$\uparrow$}\\ 
\hline
RGB & 0.756 & 0.494 & 0.623 & 0.642 &0.631 
\\ 
NIR & 0.703 & 0.445 & 0.558 & 0.617 &0.551 
\\
YCrCb~\cite{herrera2019color} & 0.745 & 0.479 & 0.596 & 0.651 &0.590 
\\ 
Bayesian\cite{zhao2020bayesian_fusion} & 0.773 & 0.485 & 0.641 & 0.661 &0.647 
\\
DarkVision\cite{Jin_2023_DarkVision} & 0.571 & 0.351 & 0.465 & 0.504 &0.466 
\\
Adaptive\cite{awad2019adaptive_rgb_nir} & 0.762 & 0.473 & 0.630 & 0.665 &0.632 
\\
VGG-NIR\cite{li2018infrared_vgg} &0.726 & 0.449 & 0.573 & 0.588 &0.590 
\\
HSV(our baseline)~\cite{fredembach2008colouring_hsv} & 0.744 &0.464 & 0.612 & 0.649 &0.608 
\\
\Xhline{2\arrayrulewidth}\Xhline{0.2pt}
\textbf{Ours} & 0.828 & 0.509 & 0.688 & 0.734 &0.687 
\\
\hline
\end{tabular}
}
\caption{\textbf{Comparison of image fusion methods for  YOLO~\cite{yolov8_ultralytics}.} Our RGB-NIR image fusion method outperforms other image-fusion methods for object detection. }
\label{tab:fusion_comparison_detection}
\end{table}

\subsection{RGB-NIR Image Fusion for Structure-from-Motion}

\paragraph{Experiments}

In this study, we evaluated the results of our image fusion approach within the COLMAP framework, a robust Structure-from-Motion (SfM) pipeline, to reconstruct 3D geometry from RGB, Near-Infrared (NIR), and fused images. 
SfM is a crucial process that estimates both intrinsic and extrinsic camera parameters from multiple images with unknown camera settings, ultimately generating a coherent 3D model of the scene.
This reconstruction process begins by selecting a subset of frames to define a world coordinate system, followed by aligning subsequent frames through feature matching, ensuring precise alignment across all images.
Effective feature extraction is essential for successful SfM, especially under challenging lighting conditions where regions may suffer from underexposure or overexposure, resulting in unreliable features.
Single-channel images, such as those in the NIR spectrum, often lack sufficient color variation, complicating feature extraction.
To overcome these limitations, we propose a fusion approach that integrates RGB images, which offer rich color details, with NIR images, which improve feature detection in low-contrast areas. 
This fusion is designed to enhance feature quality, enabling more accurate and robust 3D reconstructions within the COLMAP environment.

For our experiments, we used a well-illuminated nighttime video dataset, extracting 200 consecutive frames.
Each frame contained stereo RGB and stereo NIR images, yielding a total of 400 RGB images, 400 NIR images, and 400 fused images produced using our image fusion method. 
To ensure consistency across different reconstruction scenarios, we uniformly applied the Simple Radial model for feature extraction across all image sets.
Prior to feature extraction, all images underwent stereo calibration to determine intrinsic parameters, which were subsequently used to undistort the images.
This preprocessing step mitigated lens distortion and ensured geometric consistency, with the intrinsic parameters provided as presets for the Simple Radial model.

Feature matching was performed using exhaustive matching with a block size of 50, enabling comprehensive correspondence across image pairs. 
We conducted three reconstruction experiments, each evaluating one of the image modalities: RGB, NIR, and fused images. 
By comparing the outcomes, we assessed the impact of image fusion on feature detection and overall reconstruction accuracy under challenging lighting conditions.

\begin{figure}[h]
  \centering
  \includegraphics[width=\linewidth]{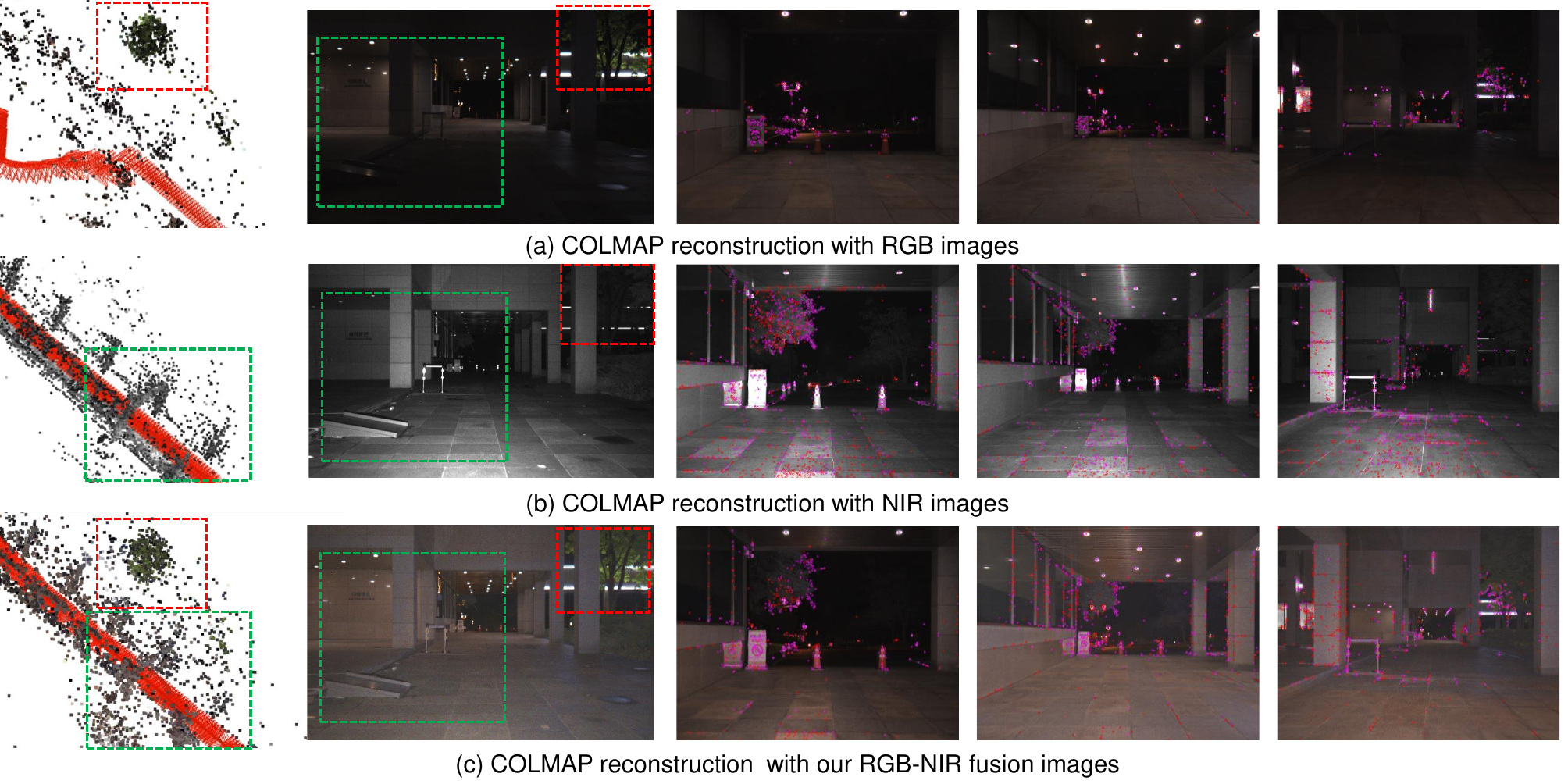}
  \caption{\textbf{COLMAP reconstruction examples for RGB, NIR and our fused images}. (a) RGB images. (b) NIR images. (c) Our fused images. }
  \label{fig:colmap_detail}
\end{figure}

\paragraph{Result}
Figure~\ref{fig:colmap_detail} presents a qualitative comparison of COLMAP feature extraction applied to RGB, NIR, and fusion images. 
Red dots represent the extracted features, while pink dots indicate matched keypoints used for feature correspondence. (c) demonstrates that the fusion images effectively integrate features from both RGB (a) and NIR (b), compensating for dark regions in one spectral domain with information from the other. 
This highlights the capability of the fusion method to leverage complementary spectral information for robust feature extraction and matching.

\subsection{Sparse Depth Reconstruction}

Sparse depth reconstruction aims to transform a sparse depth map into a full-resolution dense depth map using an RGB image as guidance \cite{Kam_2022_costdcnet, Tang_2024_bpnet}. 
Many depth estimation methods, such as LiDAR and RGB-D cameras, provide depth information at resolutions that are significantly lower than those of the accompanying RGB images.
High-end LiDAR sensors typically have a maximum resolution of 2048$\times$128, and stereo depth estimation methods are often considered to have limitations in reliability.
Excluding low-confidence regions from these methods' results leads to further sparsity and significant reductions in depth data.
Furthermore, when stereo depth maps are computed and warped onto different camera image planes, the detailed pixel-level depth information is often lost.
To address these challenges, methods such as CostDCNet \cite{Kam_2022_costdcnet} and BPNet \cite{Tang_2024_bpnet} use RGB images as guidance to reconstruct dense depth maps from sparse data.
However, certain regions of the RGB image may lack sufficient detail for depth estimation, and significant lighting variations can further degrade the reconstruction quality.
Our proposed RGB-NIR data and RGB-NIR image fusion methodology generates images with excellent photometric consistency even under challenging lighting conditions.
As a result, this approach enhances the performance of sparse depth reconstruction by enabling more reliable guidance images and improving depth estimation accuracy.

\subsection{RGB-NIR Fusion for Depth Estimation}

\subsubsection{Implementation of RGB-NIR Stereo Depth Estimations for Comparative Analysis}

This study evaluates our proposed feature-fusion method against several state-of-the-art approaches for stereo depth estimation. 
The methods compared include single-spectral stereo disparity estimation (RGB and NIR) using RAFT-Stereo~\cite{lipson2021raft}, RAFT-Stereo integrated with our image fusion technique, and other image fusion approaches~\cite{herrera2019color,fredembach2008colouring_hsv,Jin_2023_DarkVision,awad2019adaptive_rgb_nir,li2018infrared_vgg,zhao2020bayesian_fusion}. 
Additionally, comparisons were made with RGB-NIR stereo depth estimation~\cite{Zhi_2018_material_cross_spectral} and fine-tuned versions of cross-spectral stereo depth estimation models~\cite{Guo_2023_cross_stereo_est_and_data,tian2023dps_polar_stereo}.

For comparisons using RAFT-Stereo, we tested various input images while maintaining consistent weights in the RAFT-Stereo model to ensure fair pairwise comparisons. 
For multi-spectral models, we fine-tuned each method using our augmented synthetic dataset and real-world dataset to achieve optimal performance.

The CS-Stereo model~\cite{Zhi_2018_material_cross_spectral} processes left RGB and right NIR images by performing color conversion to generate RGB and NIR stereo images. 
These four images are then used for depth estimation. Instead of relying on synthetic images generated via conversion, our approach directly utilizes RGB-NIR stereo pairs as inputs and fine-tunes the depth network accordingly.

The CSPD model~\cite{Guo_2023_cross_stereo_est_and_data}, originally designed for stereo depth estimation with RGB and thermal stereo pairs, was adapted for this study. 
We replaced thermal stereo images with NIR stereo pairs and retrained the model using our dataset to facilitate direct comparison.

Lastly, the DPSNet model~\cite{tian2023dps_polar_stereo}, which extends RAFT-Stereo by incorporating polarimetry stereo images alongside RGB stereo inputs, was adapted for NIR stereo.
We replaced polarimetry stereo images with NIR stereo pairs and conducted supervised training to fine-tune the model for our application.

\subsubsection{Evaluation Metrics}

We evaluated our stereo depth estimation approach using labeled ground-truth depth data, employing a comprehensive set of metrics to assess performance across both synthetic and real-world datasets.

For the synthetic dataset, ground truth disparity maps were provided. To evaluate the accuracy of the predicted disparity maps, we utilized two primary error metrics: the Mean Absolute Error (MAE) and the Root Mean Square Error (RMSE).
We computed the absolute error (Mean Absolute Error, MAE) and the root mean square error (RMSE) between the predicted disparity map and the ground truth disparity map as primary evaluation metrics. 
Additionally, we calculated the proportion of pixels with Euclidean distance errors below thresholds of 1, 3, and 5 pixels to assess the spatial precision of the predictions. 
These evaluations resulted in five distinct metrics: MAE, RMSE, and the percentage of pixels with errors under 1, 3, and 5 pixels ($e<1\text{px},e<3\text{px},e<5\text{px}$).

For the real-world dataset, which included sparse LiDAR depth labels, we evaluated the predicted depth values by comparing them to ground truth depth measurements at LiDAR-indicated points.
The primary metrics applied in this evaluation were the MAE and RMSE of the depth values, expressed in meters, providing a robust measurement of accuracy in real-world scenarios.
Furthermore, we computed the ratio
\begin{equation}
\text{ratio} = \max\left(\frac{z_\text{pred}(u,v)}{z_\text{gt}(u,v)}, \frac{z_\text{gt}(u,v)}{z_\text{pred}(u,v)}\right)
\end{equation}
for all pixels $(u,v)$ and defined three thresholds: $\delta_1$ (ratio $\leq 1.25$), $\delta_2$ (ratio $\leq 1.25^2$), and $\delta_3$ (ratio $\leq 1.25^3$). 
These thresholds provided additional evaluation metrics, resulting in a total of five metrics: MAE based on depth, RMSE, $\delta_1$, $\delta_2$, and $\delta_3$.

\subsubsection{Additional Quantitative Results}

We present the stereo depth estimation results for both our RGB-NIR real dataset and the augmented synthetic dataset. 
Building upon the summary results presented in the main paper, we have conducted a more comprehensive analysis by categorizing the data based on different environmental conditions and expanding the evaluation metrics to include Mean Absolute Error (MAE), Root Mean Squared Error (RMSE), and threshold-based accuracy metrics ($\delta_1$, $\delta_2$, $\delta_3$).
This detailed approach facilitates a deeper understanding of the algorithm's performance across diverse scenarios.

Specifically, performance in daytime conditions is illustrated in Table~\ref{tab:depth_real_day}, which highlights the accuracy and reliability of depth estimation in well-lit environments. 
In contrast, Table~\ref{tab:depth_real_night} presents the results for night-time conditions, demonstrating the algorithm's robustness in low-light settings. 

\begin{table}[H]
    \centering

    \begin{tabular}{|l|c|c|c|c|c|}
        \hline
        Methodology     & MAE (m) & RMSE (m) & $\delta_1$ & $\delta_2$ & $\delta_3$ \\
        \hline
        RGB & 2.6322    & 4.7878    & 0.7392    & 0.8647    & 0.9353   \\
        NIR & 2.7229    & 5.017    & 0.7245    & 0.8432    & 0.9284   \\
        YCrCb~\cite{herrera2019color} & 2.6491    & 5.0697    & 0.7283    & 0.8398    & 0.9323   \\
        Bayesian\cite{zhao2020bayesian_fusion} & 2.6449    & 4.8911    & 0.7367    & 0.8647    & 0.9377   \\
        DarkVision\cite{Jin_2023_DarkVision} & 2.8401    & 5.1645    & 0.7264    & 0.8562    & 0.9277   \\
        Adaptive\cite{awad2019adaptive_rgb_nir} & 2.6662    & 4.8777    & 0.7408    & 0.8652    & 0.9289   \\
        VGG-NIR\cite{li2018infrared_vgg}  & 2.6348    & 4.7804    & 0.7394    & 0.8648    & 0.9352   \\
        HSV(our baseline)~\cite{fredembach2008colouring_hsv} & 2.6384    & 4.8087    & 0.7361    & 0.8655    & 0.937   \\   \Xhline{2\arrayrulewidth}\Xhline{0.2pt}
        CS-Stereo~\cite{Zhi_2018_material_cross_spectral}& 4.8067    & 6.5301    & 0.1842    & 0.4313    & 0.6679   \\
        CSPD~\cite{Guo_2023_cross_stereo_est_and_data}& 4.9555    & 9.3971    & \textbf{0.7782}    & 0.8251    & 0.8547   \\
        DPSNet~\cite{tian2023dps_polar_stereo}& 2.5224    & 4.6475    & 0.7584    & \textbf{0.8692}    & \textbf{0.9434}  
        \\\Xhline{2\arrayrulewidth}\Xhline{0.2pt}
               Our RGB-NIR image fusion
& 2.5498    & 4.6787    & 0.771    & 0.8586    & 0.9240   \\
        Our RGB-NIR feature fusion& \textbf{2.4938}    & \textbf{4.3216}    & \underline{0.7765}    & \underline{0.8665}    & \underline{0.9414}   \\
        \hline
    \end{tabular}
        \caption{\textbf{Metrics for Stereo Depth Estimation.} Evaluation conducted on a real-world dataset captured in a daytime environment. Results are grouped as follows: (1) the first group represents metrics computed using our RGB-NIR dataset with 3-channel stereo input processed by the RAFT-Stereo~\cite{lipson2021raft} model; (2) the second group includes results from other multi-spectral stereo depth estimation approaches, fine-tuned on our dataset; and (3) the final group showcases methods from Sections 4.1 and 4.2 of our main paper.}
        \label{tab:depth_real_day}
\end{table}

\begin{table}[h]
    \centering

    \begin{tabular}{|l|c|c|c|c|c|}
        \hline
        Methodology     & MAE (m) & RMSE (m) & $\delta_1$ & $\delta_2$ & $\delta_3$ \\
        \hline
        RGB & 3.7596    & 8.4858    & 0.5368    & 0.8548    & 0.9604   \\
        NIR& 3.3059    & 7.6952    & \textbf{0.5978}    & 0.8832    & 0.9673   \\
          YCrCb~\cite{herrera2019color}& 3.0223    & 6.1631    & 0.5127    & 0.8875    & 0.9745   \\
      
         Bayesian\cite{zhao2020bayesian_fusion}& 3.3822    & 7.2234    & 0.5409    & 0.8747    & 0.9672   \\
          DarkVision\cite{Jin_2023_DarkVision}& 3.7542    & 8.3078    & 0.5408    & 0.8548    & 0.957   \\
         Adaptive\cite{awad2019adaptive_rgb_nir}& 3.3019    & 7.5157    & 0.5576    & 0.7805    & 0.8483   \\
        VGG-NIR\cite{li2018infrared_vgg}& 3.2415    & 6.809    & 0.5422    & 0.8823    & 0.9685   \\
       
         HSV(our baseline)~\cite{fredembach2008colouring_hsv}& 3.6597    & 7.2938    & 0.4821    & 0.7238    & 0.8175   \\ \Xhline{2\arrayrulewidth}\Xhline{0.2pt}
        CS-Stereo~\cite{Zhi_2018_material_cross_spectral}& 5.7455    & 7.6971    & 0.1491    & 0.3468    & 0.553   \\
        CSPD~\cite{Guo_2023_cross_stereo_est_and_data}& 4.0065    & 10.4303    & 0.5426    & 0.7777    & 0.8628   \\
        DPSNet~\cite{tian2023dps_polar_stereo}& 3.8131    & 9.157    & 0.5362    & 0.858    & 0.9606   \\\Xhline{2\arrayrulewidth}\Xhline{0.2pt}
                Our RGB-NIR image fusion& 3.2169    & 7.1046    & 0.5474    & \textbf{0.8927}    & \textbf{0.9779}   \\
         Our RGB-NIR feature fusion& \textbf{2.9038}    & \textbf{5.5321}    & \underline{0.5665}    & \underline{0.8905}    & \underline{0.9768}   \\
        \hline
    \end{tabular}
        \caption{\textbf{Metrics for Stereo Depth Estimation.} Evaluation conducted on a real-world dataset captured in a nighttime environment. Results are grouped as follows: (1) the first group represents metrics computed using our RGB-NIR dataset with 3-channel stereo input processed by the RAFT-Stereo~\cite{lipson2021raft} model; (2) the second group includes results from other multi-spectral stereo depth estimation approaches, fine-tuned on our dataset; and (3) the final group showcases methods from Sections 4.1 and 4.2 of our main paper.}
         \label{tab:depth_real_night}
\end{table}

\subsubsection{Additional Qualitative Samples}

\begin{figure}[h]
    \centering
    \includegraphics[width=\linewidth]{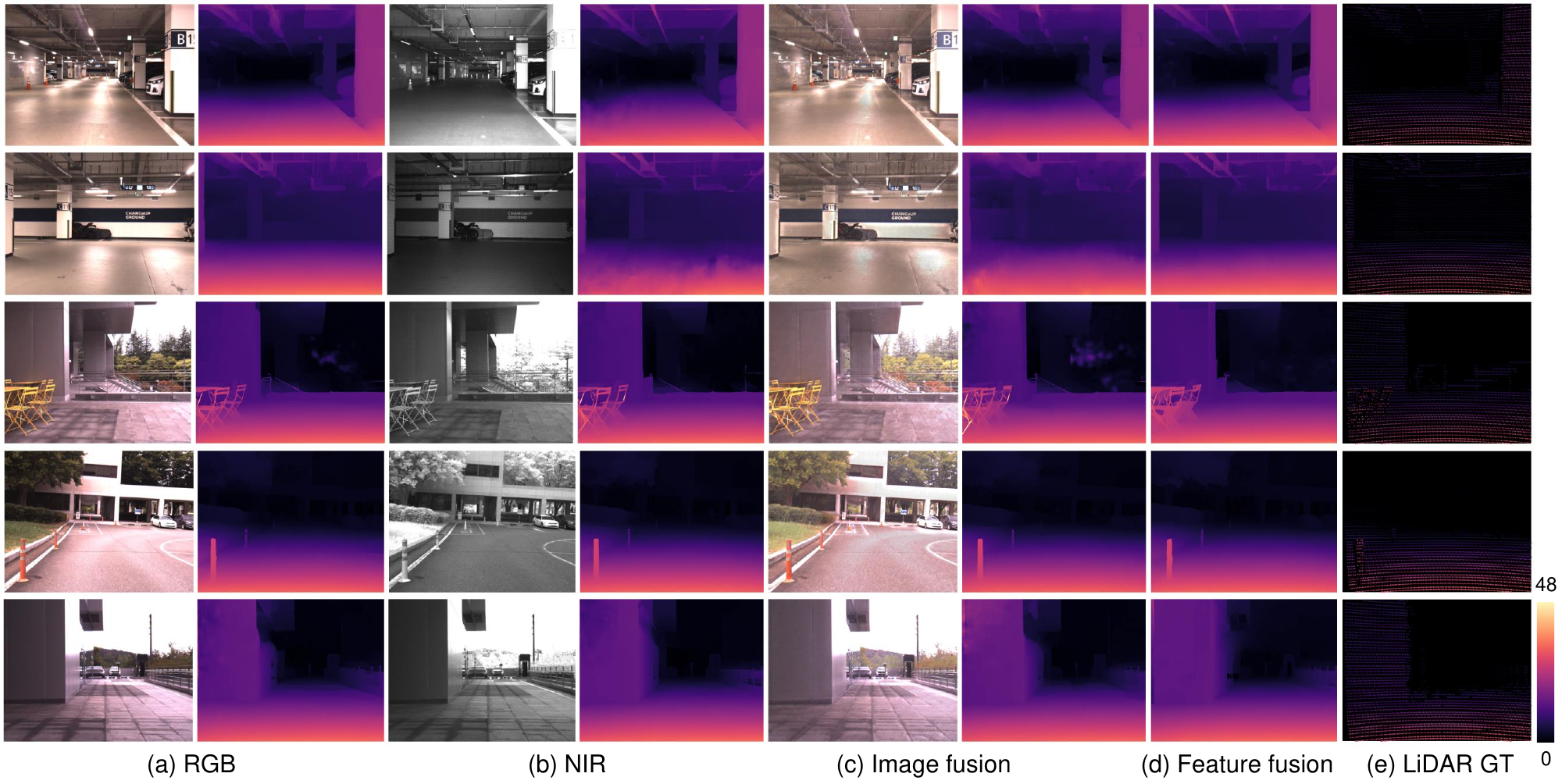}
    \caption{\textbf{Depth estimation samples of our feature fusion model}. (a) RGB images and outputs of ~\cite{lipson2021raft} with them. (b) NIR images and outputs with them. (c) Fused images by our image fusion method and outputs with them. (d) Stereo depth estimation with our feature fusion based method. (e) Ground-truth sparse LiDAR.}
    \label{fig:depth_quality_1}
\end{figure}

\begin{figure}[h]
    \centering
    \includegraphics[width=\linewidth]{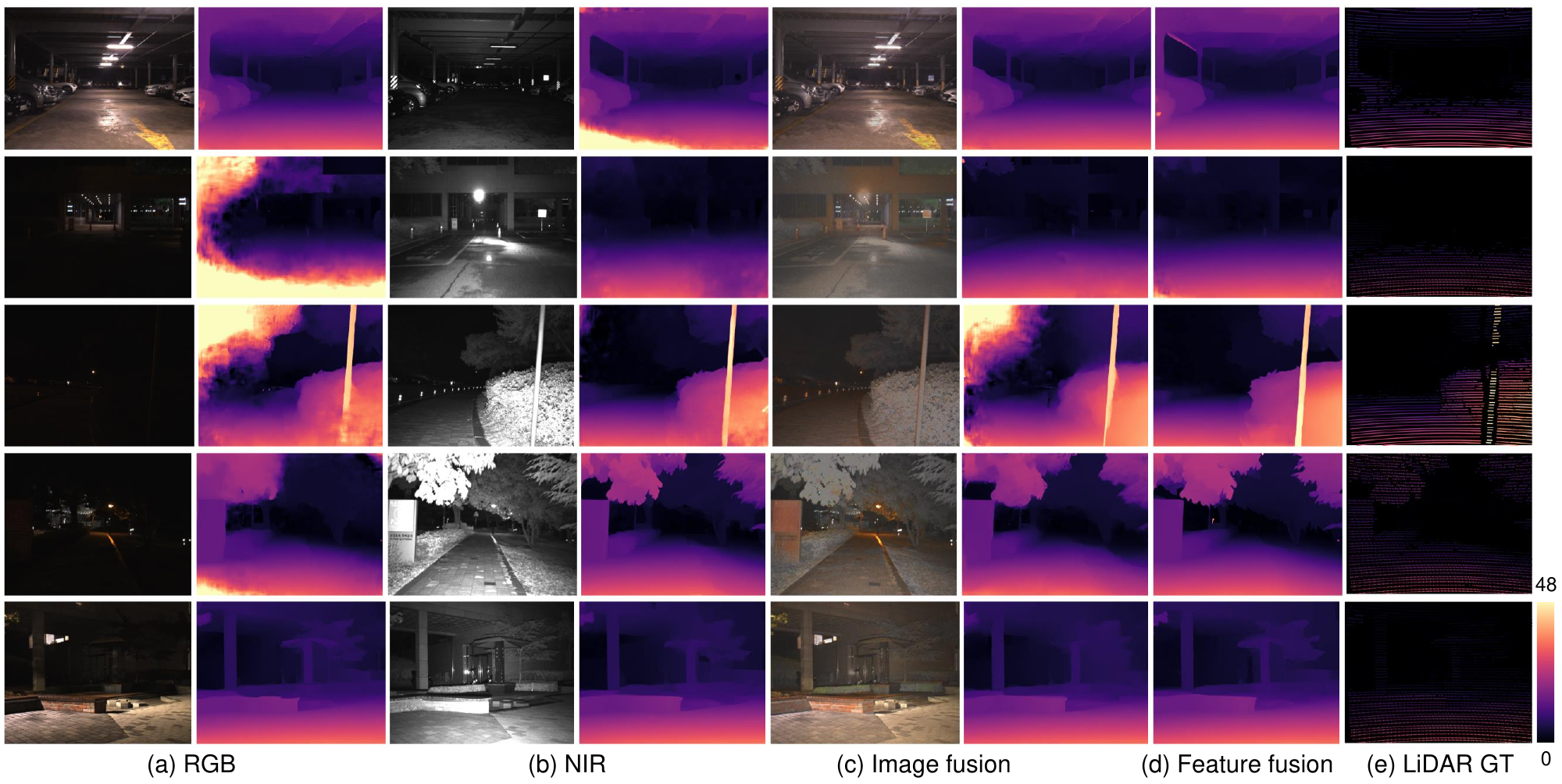}
    \caption{\textbf{Depth estimation samples on challenging lighting conditions of our feature fusion model}. (a) RGB images and outputs of ~\cite{lipson2021raft} with them. (b) NIR images and outputs with them. (c) Fused images by our image fusion method and outputs with them. (d) Stereo depth estimation with our feature fusion based method. (e) Ground-truth sparse LiDAR.}
    \label{fig:depth_quality_2}
\end{figure}

Figure~\ref{fig:depth_quality_1} presents a qualitative comparison of our proposed stereo depth estimation network with RAFT-Stereo~\cite{lipson2021raft}, evaluated on RGB images (a), NIR images (b) and our image fusion method (c). 
Our method demonstrates superior performance, particularly in regions with overexposure and high specular reflectance. 
Additionally, Figure~\ref{fig:depth_quality_2} showcases qualitative results under challenging lighting conditions. 
For low-light regions, our feature fusion approach effectively compensates for missing information by leveraging complementary spectral data, resulting in more accurate and consistent depth estimation.

\subsection{Additional Ablation Study}

\subsubsection{Pretrained Feature Encoder Weights}

\begin{figure}[h]
  \centering
  \includegraphics[width=\textwidth]{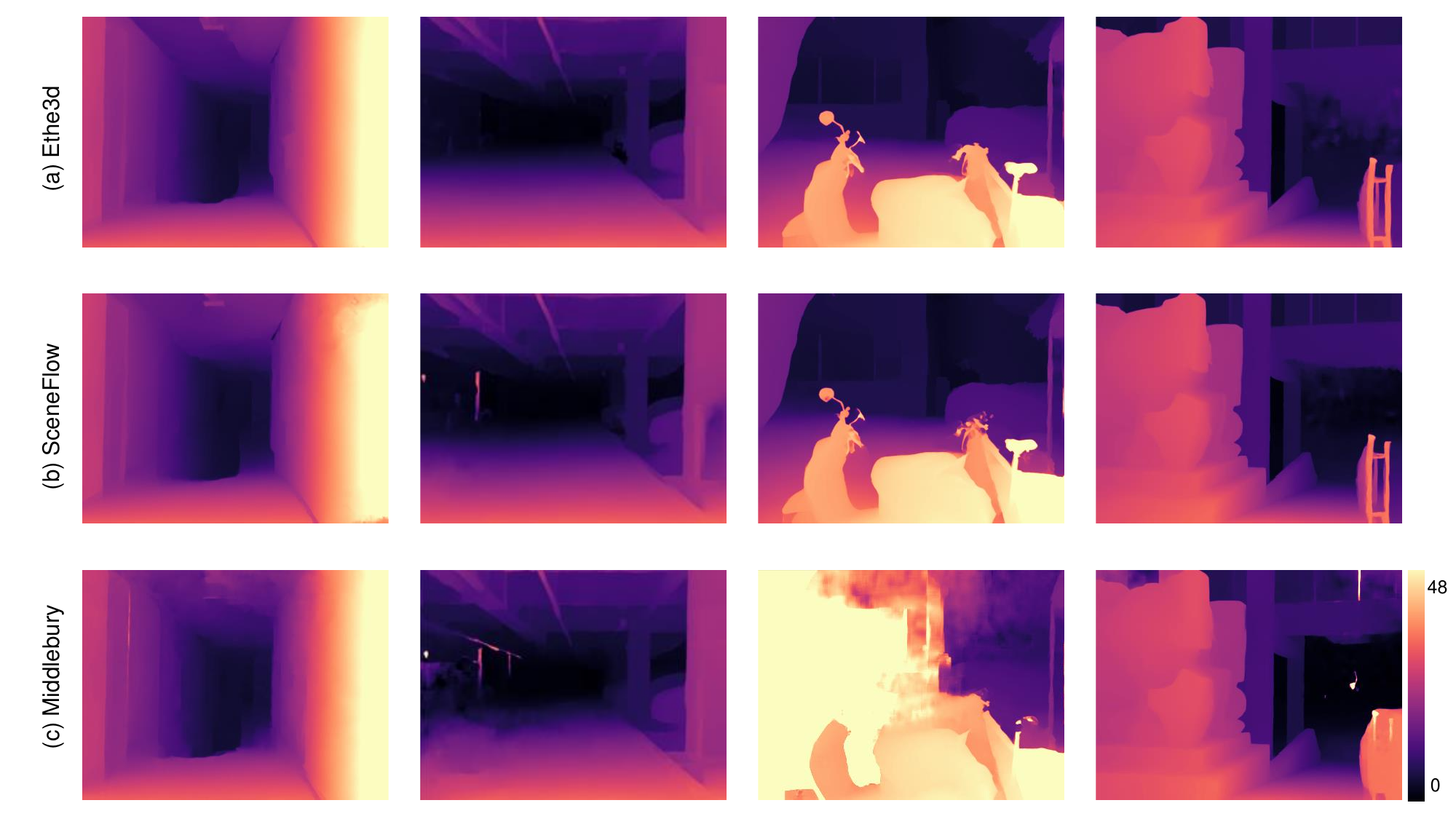}
  \caption{\textbf{Ablation study on pretrained encoder}. (a) Pretrained with~\cite{schops2019eth3d}. (b) Pretrained with~\cite{mayer2016sceneflow}. (c) Pretrained with~\cite{scharstein2014middlebury}. }
  \label{fig:ablation_encoder}
\end{figure}

Table~\ref{tab:ablation_encoder} and Figure~\ref{fig:ablation_encoder} show an ablation study of the pretrained feature encoder. 
We quantitatively and qualitatively compared the performance of our feature fusion depth estimation using three encoder weights provided by~\cite{lipson2021raft}, trained on Eth3D~\cite{schops2019eth3d}, SceneFlow~\cite{mayer2016sceneflow}, and Middlebury~\cite{scharstein2014middlebury}. 
The encoder based on Eth3D was found to be the most suitable among the three.

\begin{table}[h]
    \centering

    \begin{tabular}{|l|c|c|c|c|c|}
        \hline
        Pretrained     & MAE (m) & RMSE (m) & $\delta_1$ & $\delta_2$ & $\delta_3$ \\
        \hline
        SceneFlow~\cite{mayer2016sceneflow} & 2.6191    & 6.7860    & 0.6473    & 0.8768    & 0.9545   \\
        Middlebury~\cite{scharstein2014middlebury} & 5.0356    & 8.8270    & 0.2838    & 0.4143    & 0.4594   \\
                Eth3d~\cite{schops2019eth3d} & \textbf{2.5886}    & \textbf{6.7470}    & \textbf{0.6526}    & \textbf{0.8775}    & \textbf{0.9556}    \\
        \hline
    \end{tabular}
        \caption{\textbf{Ablation study on different pretrained feature encoder}.}
         \label{tab:ablation_encoder}
\end{table}

\subsubsection{Ablation on Feature Fusion Implementation}

\begin{figure}[h]
  \centering
  \includegraphics[width=\textwidth]{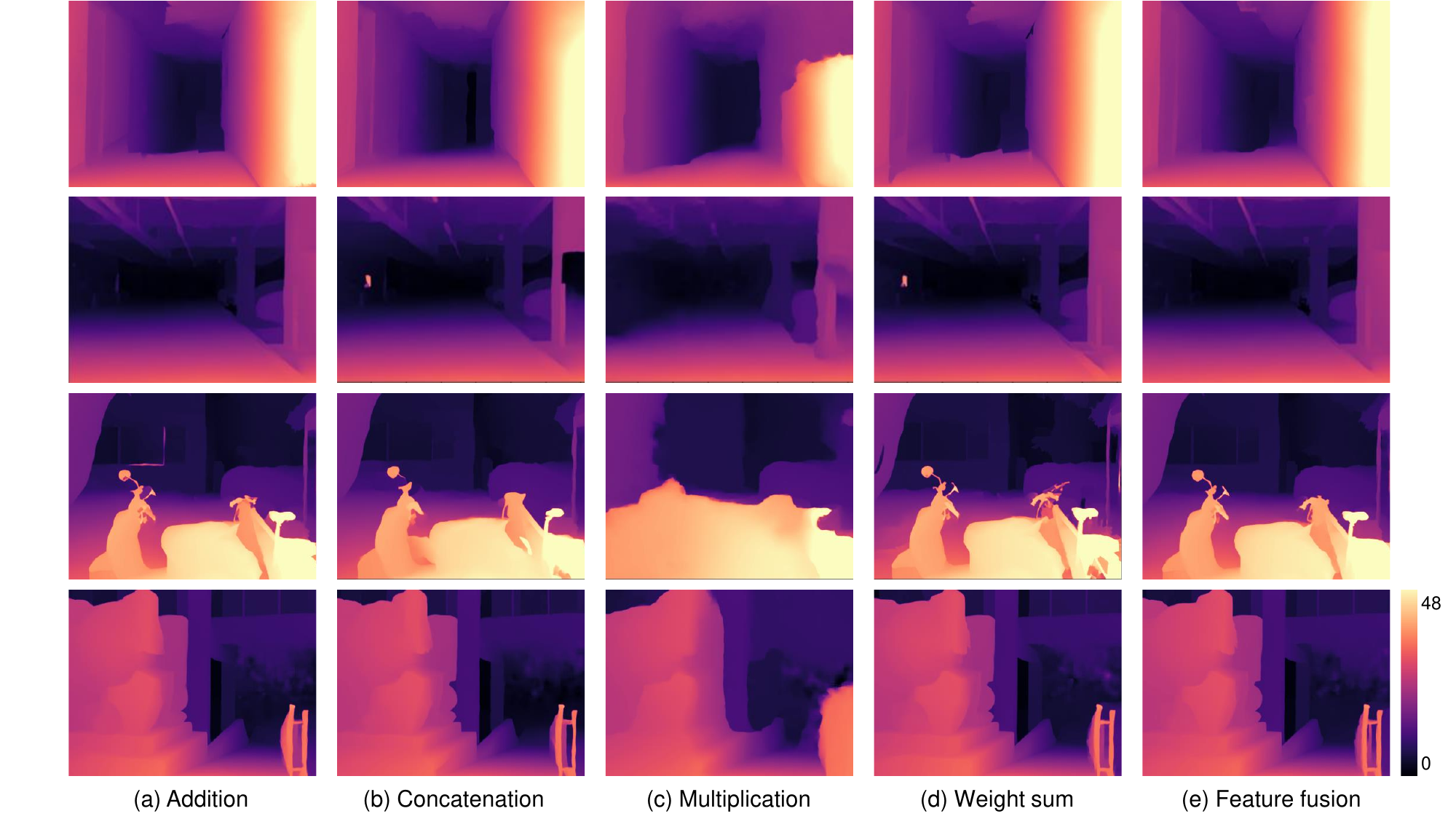}
  \caption{\textbf{Ablation study on feature fusion method}. (a) Feature addition. (b) Feature concatenation. (c) Feature multiplication. (d) Feature weight summation. (e) Attentional feature fusion~\cite{Dai_2021_aff} }
  \label{fig:ablation_fusion}
\end{figure}

Table~\ref{tab:ablation_feature_fusion} and Figure~\ref{fig:ablation_fusion} show an ablation study of different feature fusion methods. 
We explored five methods: simple feature addition, concatenation with a convolution layer to down-sample concatenated feature channels, pointwise feature multiplication, weighted sum with RGB features at 0.25 and NIR features at 0.75, and our implementation of the attentional feature fusion method inspired by~\cite{Dai_2021_aff}. 
Although feature multiplication showed the best quantitative results when evaluated using sampled sparse LiDAR information, it performed poorly in qualitative analysis. 
The attentional feature fusion method demonstrated the best adaptation to spatially varying lighting conditions.

\begin{figure}[h]
  \centering
  \includegraphics[width=\textwidth]{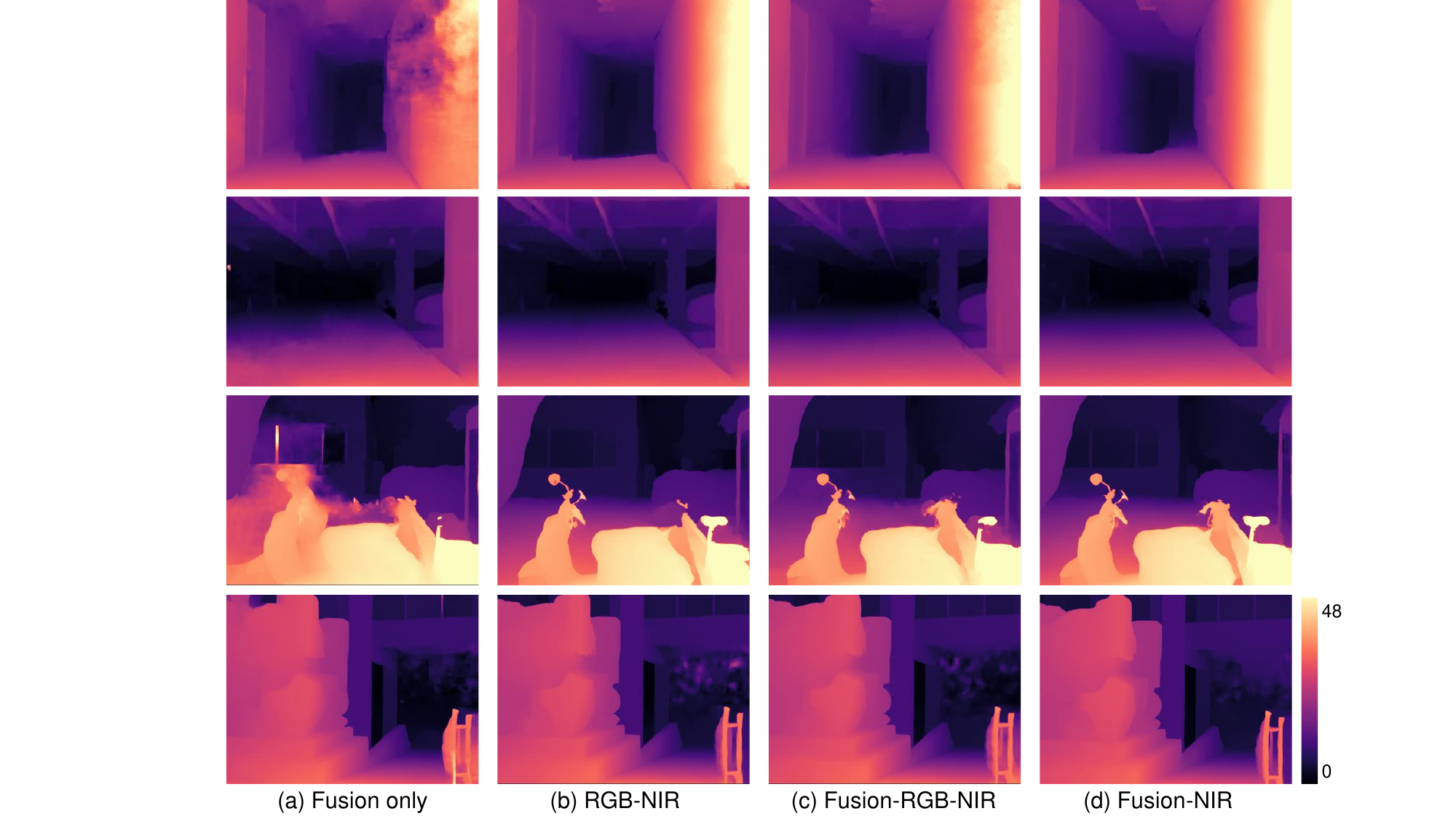}
  \caption{\textbf{Ablation study on alternative correlation volume searching}. (a) Fusion correlation volume only. (b) RGB and NIR correlation volumes. (c) Fusion, RGB and NIR correlation volumes. (d) Fusion and NIR correlation volumes. }
  \label{fig:ablation_alter}
\end{figure}

Table~\ref{tab:ablation_alternation} and Figure~\ref{fig:ablation_alter} show an ablation study on different correlation volume alternating methods. 
We compared using only the Fusion volume, alternating between RGB and NIR volumes, alternating among Fusion, RGB, and NIR volumes, and alternating between Fusion and NIR volumes. 
In a quantitative evaluation using sparse LiDAR ground truth, the Fusion-RGB-NIR alternating method achieved the best performance. However, the most reliable qualitative results were obtained with the Fusion-NIR alternating method.

\begin{table}[h]
    \centering

    \begin{tabular}{|l|c|c|c|c|c|}
        \hline
        Pretrained     & MAE (m) & RMSE (m) & $\delta_1$ & $\delta_2$ & $\delta_3$ \\
        \hline
Feature addition & 2.611    & 6.5365    & 0.6526    & 0.8772    & 0.9552   \\
Concatenation & 2.5719    & 6.004    & 0.6455    & \textbf{0.8782}    & 0.954   \\
Multiplication & \textbf{2.4524}    & \textbf{5.8743}    & \textbf{0.6634}    & 0.8768    & 0.9517   \\
Weight sum & 2.6251    & 6.9073    & 0.6491    & 0.8686    & 0.9535   \\
Attentional feature fusion & \underline{2.5886}    & 6.7470    & 0.6422    & \underline{0.8775}    & \textbf{0.9556}   \\
        \hline
    \end{tabular}
        \caption{\textbf{Ablation study on different implementations of feature fusion}.}
         \label{tab:ablation_feature_fusion}
\end{table}

\begin{table}[h]
    \centering

    \begin{tabular}{|l|c|c|c|c|c|}
        \hline
        \textbf{Correlation volumes for disparity estimation}    & \textbf{MAE (m)} & \textbf{RMSE (m)} & $\delta_1$ & $\delta_2$ & $\delta_3$ \\
        \hline
    Fusion correlation volumes only & 3.3979    & 7.4405    & 0.5073    & 0.7409    & 0.8073   \\
    Alternating RGB-NIR correlation volumes & 2.6508    & 8.5712    & \textbf{0.6575}    & \textbf{0.8841}    & 0.9555   \\
    Alternating Fusion-RGB-NIR correlation volumes & \textbf{2.5238}    &  7.4262   & 0.6336    & 0.8763    & 0.9550   \\
    Alternating Fusion-NIR correlation volumes & \underline{2.5886}    & \textbf{6.7470}    & \underline{0.6422}    & \underline{0.8775}    & \textbf{0.9556}   \\
        \hline
    \end{tabular}
        \caption{\textbf{Ablation study on different implementations of correlation volume alternations}.}
         \label{tab:ablation_alternation}
\end{table}


{\small
\bibliographystyle{ieee_fullname}
\bibliography{references}
}

\end{CJK}